\documentclass[journal]{IEEEtran}
\usepackage{amsmath,amsfonts}
\usepackage{algorithmic}
\usepackage{algorithm}
\usepackage{array}
\usepackage[caption=false,font=normalsize,labelfont=sf,textfont=sf]{subfig}
\usepackage{textcomp}
\usepackage{stfloats}
\usepackage{url}
\usepackage{verbatim}
\usepackage{graphicx}
\usepackage{cite}
\usepackage[table,svgnames]{xcolor} 
\usepackage{caption} 
\usepackage{array} 
\usepackage{booktabs} 

\usepackage[column=0]{cellspace} 
\usepackage{booktabs}
\usepackage{graphicx}
\usepackage{amsmath}
\usepackage{mathabx}
\usepackage{subcaption}
\usepackage{float}

\begin{document}

\title{RS5M and GeoRSCLIP: A Large Scale Vision-Language Dataset and A Large Vision-Language Model for Remote Sensing}


\author{
Zilun Zhang, 
Tiancheng Zhao,
Yulong Guo,
Jianwei Yin

\thanks{
\textit{Corresponding author: Tiancheng Zhao}. Zilun Zhang, Yulong Guo, and Jianwei Yin are with the College of Computer Science and Technology, Zhejiang University, Hangzhou, China; Tiancheng Zhao is with the Binjiang Research Institute of Zhejiang University (e-mail: zilun.zhang@mail.zju.edu.cn; tianchez@zju-bj.com; gyl\_cs@zju.edu.cn; zjuyjw@zju.edu.cn).

}
}

\markboth{Journal of \LaTeX\ Class Files,~Vol.~14, No.~8, August~2021}%
{Shell \MakeLowercase{\textit{et al.}}: A Sample Article Using IEEEtran.cls for IEEE Journals}


\maketitle

\begin{abstract}
  Pre-trained Vision-Language Models (VLMs) utilizing extensive image-text paired data have demonstrated unprecedented image-text association capabilities, achieving remarkable results across various downstream tasks. A critical challenge is how to make use of existing large-scale pre-trained VLMs, which are trained on common objects, to perform the domain-specific transfer for accomplishing domain-related downstream tasks. In this paper, we propose a new framework that includes the Domain pre-trained Vision-Language Model (DVLM), bridging the gap between the General Vision-Language Model (GVLM) and domain-specific downstream tasks. Moreover, we present an image-text paired dataset in the field of remote sensing (RS), RS5M, which has 5 million RS images with English descriptions. The dataset is obtained from filtering publicly available image-text paired datasets and captioning label-only RS datasets with pre-trained VLM. These constitute the first large-scale RS image-text paired dataset. Additionally, we fine-tuned the CLIP model and tried several Parameter-Efficient Fine-Tuning methods on RS5M to implement the DVLM. Experimental results show that our proposed dataset is highly effective for various tasks, and our model GeoRSCLIP improves upon the baseline or previous state-of-the-art model by $3\%\sim20\%$ in Zero-shot Classification (ZSC) tasks, $3\%\sim6\%$ in Remote Sensing Cross-Modal Text–Image Retrieval (RSCTIR) and $4\%\sim5\%$ in Semantic Localization (SeLo) tasks. Dataset and models have been released in: \url{https://github.com/om-ai-lab/RS5M}.
\end{abstract}


\section{Introduction}
Remote sensing (RS) images have been playing an important role in environmental monitoring \cite{rsen}, urban planning \cite{rsup}, and natural disaster management \cite{rsnda}, etc. However, the rapid growth of RS images has introduced new challenges in efficiently and effectively processing, analyzing, and understanding the information contained within RS data. Over the past decade, supervised deep learning models have become powerful tools for tackling these challenges, demonstrating great success in RS tasks such as scene classification, object detection, semantic segmentation, and change detection. Despite these advances, the performance of deep learning models in RS applications is often constrained by small-scale labeled datasets. The interpretation of RS images typically requires domain expertise, leading to an expensive cost of RS image labeling, causing a bottleneck in further improvement in RS downstream tasks. As a natural supervision for the RS image, the paired text has incredible potential to help learn better data representation and serve as a proxy for various RS image modalities, such as SAR, hyperspectral, and imagery acquired from different satellites. 

The rapid development of deep learning models has led to significant progress in both CV and NLP domains, and researchers have begun to explore the potential of combining visual and textual modalities to develop more powerful and versatile models capable of understanding multimodal content. Pre-trained Vision-Language Models (VLMs) (\cite{clip}, \cite{align}, \cite{visualbert}, \cite{vilt}, \cite{uniter}, \cite{albef}, \cite{oscar}, \cite{coca}, \cite{flamingo}, \cite{florence}, \cite{declip}, \cite{blip}, \cite{blip2}, \cite{beit3}, \cite{kosmos}) have been a promising approach to leverage the strengths of natural language's tokenized information and the abundant visual information in images to serve as the General Vision-Language Model. A notable example is CLIP \cite{clip}, which utilizes the contrastive loss function to connect two modalities, leading to unprecedented generalizability in many downstream tasks and domain transfer. Another important application for VLMs is generative models such as DALLE \cite{dalle} and stable-diffusion \cite{stable-diffusion} for AI-generated Content. However, due to the nature of training with common object data, VLMs usually underperform in specialized domains such as remote sensing \cite{clip} and medical imaging\cite{DAMMFM} because of the mismatch between domains.

\begin{figure*}
    \centering
    \includegraphics[width=\textwidth]{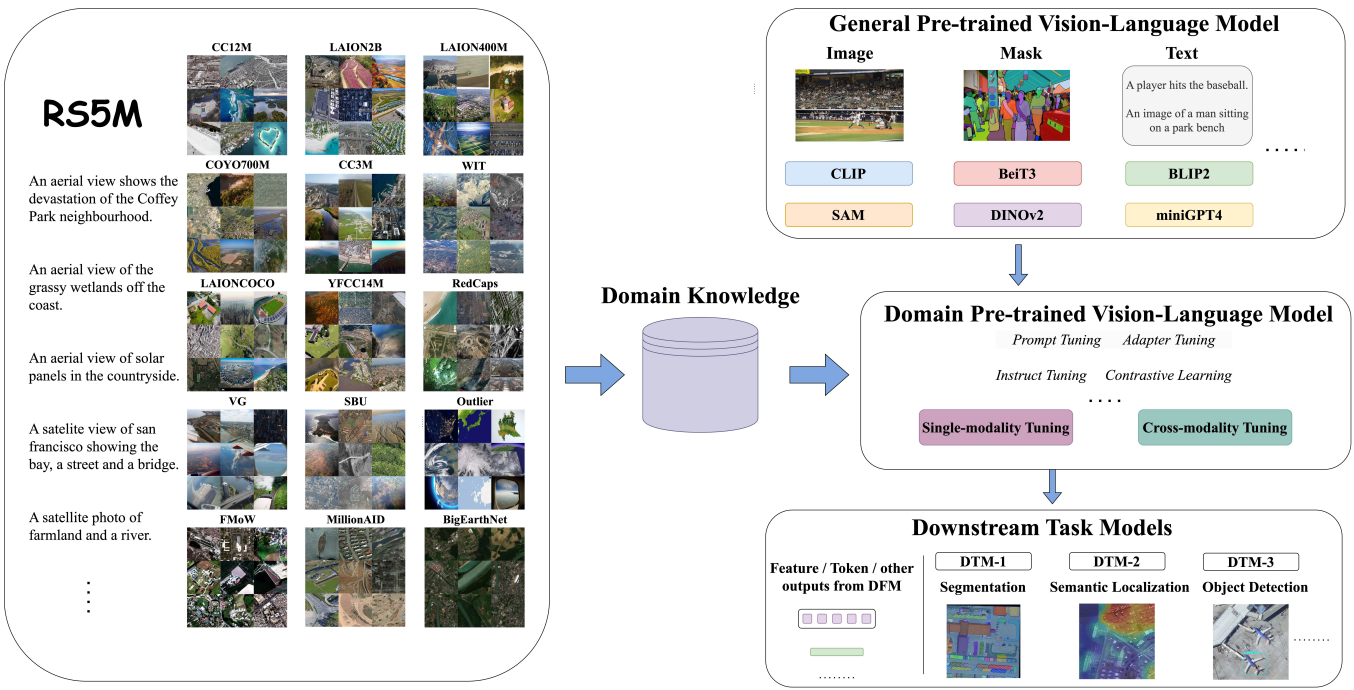}
    \caption{Illustration of our proposed Framework. The Domain Vision-Language Model (DVLM) plays a central role in accepting the general knowledge from the General Vision-Language Model (GVLM) and is injected with massive domain-specific knowledge from external data. With the proper learning paradigm, DVLM is able to transfer the general knowledge with domain-specific prior to the Downstream Task Model (DTM) for domain-specific tasks. A demo for our proposed RS5M dataset is on the left.}
    \label{fig:dfm}
\end{figure*}

To make use of the power of the GVLM in the RS domain, it is important to design a DVLM capable of leveraging the generalizability of the GVLM, incorporating external domain prior knowledge, and transferring this knowledge to a domain-specific Downstream Task Model (DTM) through a suitable learning paradigm to solve downstream tasks, as depicted in Figure \ref{fig:dfm}. Alfassy et. al proposed FETA \cite{alfassy2022feta}, a specializing Vision-Language model for expert task applications, directly tuning the Vision-Language model using LoRA for retrieval tasks in public car manuals and sales catalogue brochures, but the $GVLM \xrightarrow{} DVLM \xrightarrow{} DTM$ structure and the importance of DVLM were not widely discussed. The amount of training data to develop a DVLM may not be as much as GVLM (400M for CLIP \cite{clip}, 1B for ALIGN \cite{align}, 88M for DeCLIP \cite{declip}, etc.), but still is the foundation for the success of DVLMs. 

In terms of RS, textual information such as geospatial metadata, land cover annotations, expert descriptions, and image captions provides natural supervision for RS images, offering richer context than class-level labels alone. He et al. improve the CLIP zero-shot image recognition top-1 accuracy by 17.86\% on the EuroSAT dataset when using the synthetic data from GLIDE to fine-tune the classifier (supervised by cross-entropy) \cite{wortsman2022robust}, presenting a promising potential on CLIP model with auxiliary in-domain data \cite{synthetic}. Several studies have proposed RS image-text paired datasets, including \cite{UCMSydeney} \cite{UCMSydeney} \cite{RSICD} \cite{RSITMD} \cite{rsvg}. However, these datasets contain too few samples to effectively transfer or fine-tune large-scale pre-trained VLMs. Concurrently, there are large-scale RS datasets \cite{millionaid} \cite{ben} \cite{fmow} containing millions of RS images but with only class-level labels. In overall, large-scale image-text paired dataset is rare in the field of RS, therefore gathering extensive in-domain data is crucial. 

The contributions of this paper can be summarized as follow:  We introduce the first large-scale remote sensing image-text paired dataset, \textbf{RS5M}, which is entirely based on filtering large-scale image-text paired datasets and captioning RS datasets with the pre-trained model. Extensive denoising methods are applied. RS5M is nearly 1000 times larger than the existing largest RS image-text paired datasets.  We propose the concept of the \textbf{Domian Vision-Language Model} (DVLM) to better utilize the \textbf{General Vision-Language Model} (GVLM) and domain-specific data. In the RS field, we implement the DVLM with several Parameter-Efficient Tuning methods with Vision-Language Models for the RS-related vision-language tasks. Through extensive experiments, we demonstrate that our framework, in combination with our proposed RS5M dataset, can \textbf{successfully transfer pre-trained VLMs to the RS domain and perform better on related downstream tasks}. \textbf{Our proposed model GeoRSCLIP, trained with RS5M, improving upon the baseline/state-of-the-art by $3\%\sim20\%$ in Zero-shot Classification tasks, $3\%\sim6\%$ in Remote Sensing Cross-Modal Text–Image Retrieval and $4\%\sim5\%$ in Semantic Localization tasks}.

\section{Related Work} \label{relatedwork}
Detailed introduction on related works can be found in Appendix A.1 \ref{appendix:related_work}. We will introduce RS datasets, pre-trained VLMs, VLM for RS, pre-trained models for RS, and PEFT for LLMs and VLMs. 

Commonly used RS image-text paired datasets include \textbf{UCM Captions} \cite{UCMSydeney}, \textbf{Sydney Captions} \cite{UCMSydeney}, \textbf{RSICD} \cite{RSICD}, \textbf{RSITMD} \cite{RSITMD}, and \textbf{RSVGD} \cite{rsvg}. These datasets' image sizes span from 224 $\times$ 224 pixels up to 800 $\times$ 800 pixels, while spatial resolution varies from 0.5m to 30m. Among them, RSVGD holds the largest collection with 38,320 RS image-text pairs, albeit with some image duplication. In addition, there are larger-scale image datasets like \textbf{BigEarthNet} \cite{bigearthnet}, \textbf{Functional Map of the World} (FMoW) \cite{fmow}, and \textbf{MillionAID} \cite{millionaid}, consisting of 590,326, 1,047,691, and 1 million RS images respectively. These images contain class-level labels.

Large-scale pre-trained VLMs can be categorized based on their pre-training task objectives, such as contrastive vision-text alignment, image-text matching, masked language modeling, etc. \cite{du2022survey}. \cite{clip}, \cite{align}, \cite{filip}, and \cite{declip} align textual and visual information in a shared semantic space using contrastive learning task. \cite{uniter}, \cite{albef}, and \cite{blip} employ image-text matching task objectives. Models such as \cite{oscar}, \cite{flip}, and \cite{beit3} utilize Masked Language Modeling objectives. Most pre-trained VLMs combine multiple pre-training task objectives and use them to mine fine-grained relationships between modalities. For instance, \cite{albef} employs contrastive loss and image-text matching loss, \cite{coca} utilizes contrastive loss and captioning loss, and \cite{flip} uses contrastive loss and loss from MAE \cite{mae}. The success of VLMs is closely linked to the vast amount of paired data. In terms of RS, Zhang et al.\cite{airs} provided a comprehensive overview of recent advancements in applying artificial intelligence techniques to remote sensing data analysis. Wen et al. \cite{wen2023visionlanguage} survey the current progress and discuss the future trends of VLMs in the field of RS. Lobry et al. introduced the RSVQA task \cite{rsvqa}, a system where images can be queried to obtain specific information about their content. Hu et al. presented RSIEval \cite{hu2023rsgpt}, a benchmark consisting of human-annotated captions and visual question-answer pairs, enabling a thorough assessment of VLMs in remote sensing. Yuan et al. \cite{RSITMD} introduced an asymmetric multimodal feature matching network for cross-modal RS Vision-Language Retrieval tasks. They also proposed Semantic Localization task \cite{selo}, a weak visual grounding task enabling semantic-level retrieval with caption-level annotation, and GaLR \cite{galr}, a method that combined local and global features of RS images. Basso introduced CLIP-RS \cite{clip-rs}, and Arutiunian et al. fine-tuned CLIP with RSICD, achieving significant improvements in top-1 accuracy for zero-shot classification \footnote{https://huggingface.co/blog/fine-tune-clip-rsicd}. Wang et al. pre-trained CNN and ViT-based backbones \cite{rsp} in Million-AID, examining on various downstream tasks. They also proposed a 100M ViT with rotated varied-size window attention, achieving competitive results for downstream tasks such as classification, object detection, and segmentation. However, their dataset and models are single-modality and therefore cannot utilize the supervision from text, suggesting potential improvements with VLMs.

Large Language Models (LLMs) such as Bert \cite{bert} and GPT \cite{gpt} trained on vast text corpora, have achieved state-of-the-art results across numerous NLP tasks. However, their millions or billions of parameters make full fine-tuning for each downstream task unrealistic. Adapters \cite{houlsby2019parameterefficient} offer an alternative solution for LLM fine-tuning, as they freeze the pre-trained LLM's weights while training only the adapter's parameters, which have significantly less number of parameters. This approach speeds up adaptation while maintaining comparable performance to full fine-tuning. \cite{pfeiffer2021adapterfusion}, \cite{li2021prefixtuning}, \cite{hu2021lora}, \cite{unipelt} further improve the methods. \cite{clipadapter}, \cite{tipadapter}, \cite{vladapter} are introduced for tuning VLMs on visual classification, VQA and image captioning tasks. Prompt-based learning methods such as \cite{coop} and \cite{cocoop} learn prompt tokens for input in the text encoder to assist zero-shot classification.

\section{Dataset Construction}

\begin{figure*}
    \centering
    \includegraphics[width=0.9\textwidth]{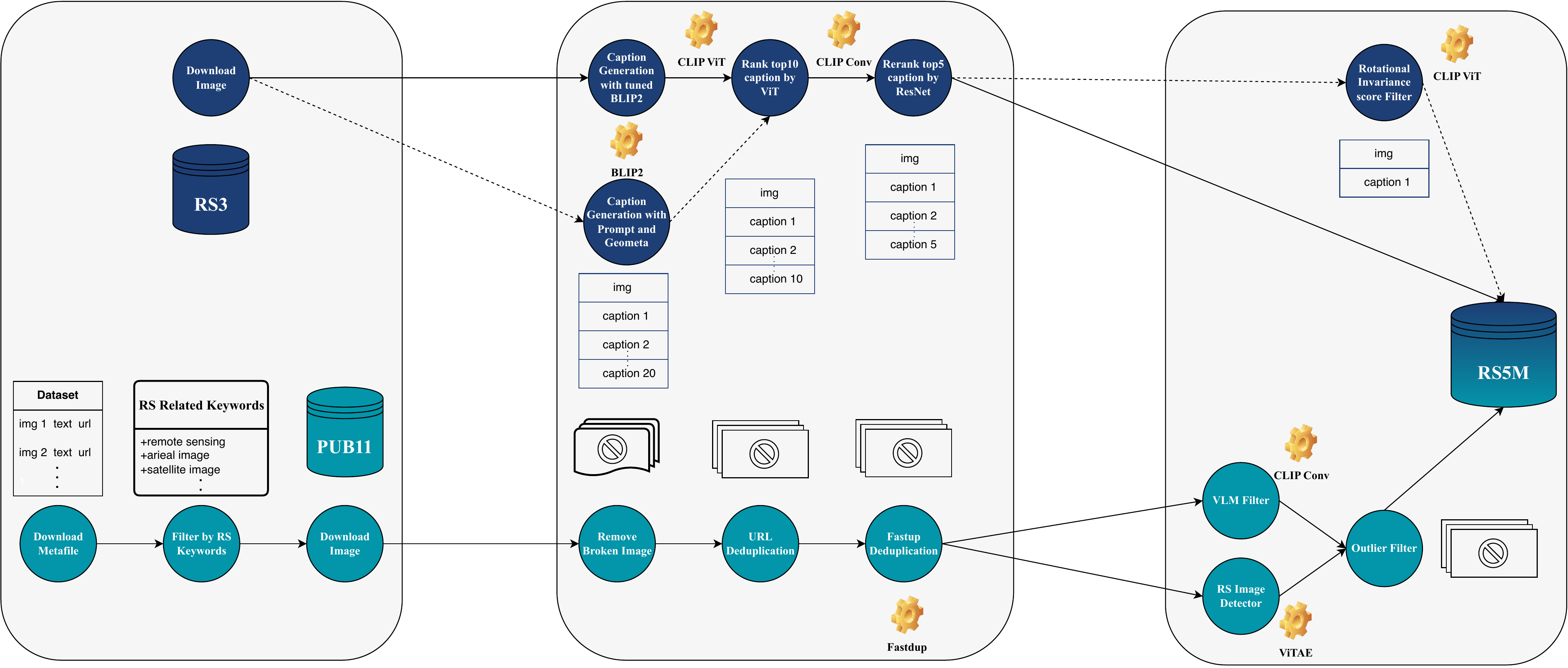}
    \caption{Overview of the collection process for RS5M. Circles represent different steps, gears stand for the model utilized, rectangles represent the images, and dash lines connect to the optional step.}
    \label{fig:collection_process}
    
\end{figure*}

We constructed the RS5M through two sources (see Figure \ref{fig:collection_process}) \footnote{For better reading experience, we moved the construction details into Appendix B}. First, we gather 11 publicly available image-text paired datasets (PUB11) and filter them using RS-related keywords. We then utilize the URLs and other tools to deduplicate images. Next, we use a pre-trained VLM and an RS image detector to remove non-RS images. Second, we utilize BLIP2 \cite{blip2} to generate captions for 3 large-scale RS datasets (RS3) that only have class-level labels. We conduct a series of quality assurance methods including a self-supervised one to acquire descriptive and suitable captions for RS images. Finally, we merge the results from both sources. More details can be found in Appendix \ref{appendix:detail_filter_pub} and Appendix \ref{appendix:metacap}. License information is listed in Appendix \ref{appendix:license}.

\subsection{Filter Large-Scale Image-Text Paird Datasets}
\label{pub11}

We have chosen 11 public large-scale English image-text paired datasets to build the PUB11 subset, including LAION2B-en \cite{laion5b}, LAION400M \cite{laion400m}, LAIONCOCO, COYO700M \cite{coyo700m}, CC3M \cite{cc3m}, CC12M \cite{cc12m}, YFCC15M \cite{yfcc100m}, WIT \cite{wit}, Redcaps \cite{redcaps}, SBU \cite{sbu}, and Visual Genome \cite{vg}. A brief introduction on them can be found in Appendix \ref{appendix:pub11}. We collected \textbf{3 million image-text pairs} in this procedure. The aerial view images are predominant, but there are still some satellite images in the collection. Table~\ref{appendix:pubdataset_stat} in the Appendix lists the statistics for each dataset including the number of images that remained in each dataset after filtering. We put most of the processing details in Appendix \ref{appendix:detail_filter_pub}. 

We establish a set of keywords closely related to RS, which consists of two groups: RS-related nouns and RS-related applications \& companies names (Appendix \ref{keywords}). To identify image-text pairs with text containing the keyword patterns, we utilize regular expressions. After downloading all relevant images from the internet, we utilize fastdup \footnote{https://visual-layer.readme.io} for invalid image checking and deduplication. We first filter out corrupted images, and apply deduplication based on URLs. Then, fastdup is used to cluster duplicate images. We keep one image and discard the rest for each cluster of duplicate images. 

After checking for invalid images and performing deduplication, we proceeded to clean the dataset using VLM and the RS image Detector. First, we develop a set of handcrafted RS-related text prompt templates with length $n$, $t_{j \in \{1, n\}}$ (refer to Appendix \ref{rs_pt} for details). For each image $x_i$, we select a CNN based CLIP-ConvNext-XXL model \cite{openclip} to compute the cosine similarity $s_i$ between the average text feature $f_{t} = \frac{\sum_{j=1}^{n} f_{text}(t_j)}{n}$ of the prompt templates and the image feature $f_{image}(x_i)$, i.e., $s_i =\frac{f_t \cdot f_{image}(x_i)}{|f_t| \cdot |f_{image}(x_i)| }$, since we will jointly use a ViT-based model later. Then, we construct a classification dataset comprising two classes: RS images ($c_{RS}$) and non-RS images ($c_{nRS}$). Details on this classification dataset can be found in Appendix \ref{bi_cls_dataset}. Next, we fine-tune a classifier, which is integrated with the ViTAE pre-trained model \cite{wang2022advancing}, to serve as an RS image detector. We denote the probability of an image $x_i$ is an RS image to be $c_i = P(c_{RS}| x_i )$. Lastly, we filter the images in RS5M based on the joint score $(s_i, c_i)$. We keep images with $s_i \geq m$ and $c_i \geq n$, where $m$ and $n$ represent some thresholds. In practice, we set $m$ and $n$ to specific values to only keep image-text pairs that have the top 90\% $s_i$ score and top 80\% $c_i$ score among all image-text pairs. The PUB11 subset we constructed included both the satellite view and aerial view images. There is an analysis of outliers and misfiltered images for PUB11 in Appendix \ref{appendix:outlier_analysis}. We have 3,007,809 image-text pairs in total.

\subsection{Caption Remote Sensing Image Datasets}
\label{rs3}

Despite the domain difference exists in RS images and images with common objects, captioning RS images with VLMs pre-trained on images with common objects has proven to be effective, as demonstrated in \cite{blip} and Appendix \ref{appendix:RS3}, Figure \ref{fig:caption_compare}. We employ the tuned BLIP2 model (tuning details can be found in Appendix \ref{appendix:tunedblip2})  \cite{blip2} with the OPT 6.7B checkpoint in half-precision from Huggingface for caption generation. We choose nucleus sampling as it generates more diverse captions (refer to Appendix \ref{sampleing_method}). The selected datasets include BigEarthNet \cite{bigearthnet}, FMoW \cite{fmow}, and MillionAID \cite{millionaid}, which are detailed in section \ref{relatedwork}. We use only the training set for FMoW (727,144 images) and BigEarthNet (344,385 images), as some downstream tasks evaluate the test set. For the MillionAID dataset, we select the test set (990,848 images). We have 2,062,377 images in total for the RS3 subset.

We follow the work of Schuhmann et al. \footnote{https://laion.ai/blog/laion-coco/} in the LAIONCOCO dataset and refine their approach. We generate 20 candidate captions per image and rank the top 10 results using CLIP ViT-H/14. Then, we re-rank these top 10 results using CLIP Resnet50x64 to obtain the top 5 captions. Moreover, we enhanced the dataset by integrating metainformation (geo-meta information, class labels, UTM, UTC, etc.) into readable sentences as a part of the image caption. More can be found in Appendix \ref{appendix:metacap}. This structured meta-caption, combined with the model-generated caption, offers a more comprehensive view. Appendix \ref{appendix:RS3}, Figure \ref{fig:blip2_captioning_result} highlights several examples of our captioning results (machine-generated part only). By sampling 2,000 captions and evaluating them through human assessment, we found the top captions provide a satisfactory degree of description for the RS images from these datasets (see \ref{appendix:ratingexp} for experiment details). In the examples provided, objects such as airports, rivers, farmland, bridges, streets, bays, and roundabouts are all present in the images. 

Rotation-invariant features are crucial in the field of remote sensing, as targets on the ground captured by satellites or drones typically maintain their shape, size, and color, such as rivers, forests, and cultivated lands. However, changes in the shooting angle may result in rotations for targets. Therefore, we aim to generate captions that accurately describe the image content, regardless of the shooting angle. To achieve this, we design a rotation-invariant criterion for obtaining high-quality captions. Our criterion is that for an image $x$, suppose we have $k$ captions for the image from the previous steps, denoted by $t^j$, where $j \in \{1\dots k\}$, and we augment the image by rotating it at 12 different angles with an increment of 30 degrees (denote as  ${x_n}$, $n \in \{1, \dots, 12\}$). Our goal is to find a $j$ so that $t^j$ minimizes the variance of cosine similarity between image features for images in different angles and text features. In other words, regardless of the image rotation angle, the matching score between the caption and the rotated images should be only negligibly influenced. That is, $\underset{j}{\arg\min} \{s^j\}_{j \in \{1 \dots k\}}, s^j =\textbf{Var}(\frac{f_{text}{(t^j)} \cdot f_{image}{(x_n)}}{|f_{text}{(t^j)}| \cdot |f_{image}{(x_n)}| })$.  The selection results can also be observed in Figure \ref{fig:blip2_captioning_result} and a visualization of this process is displayed in Appendix \ref{appendix:rotational_invariance}, Figure \ref{fig:rotation}. Captions chosen through this method tend to be more general and broader, effectively excluding hallucinated captions. However, detailed descriptions of image content may sometimes be omitted.

\section{Dataset Description}

Figure \ref{fig:dataset_description} left shows the frequency statistics of keywords (can be found in Appendix \ref{appendix:pubdataset_stat}) appearing in the image captions. The phrase "aerial view" is predominant in the captions, resulting in a significant number of aerial view remote sensing images in the RS5M dataset. The middle Figure presents a word cloud of words extracted from the RS5M captions. All special characters and numbers have been removed, as well as the majority of prepositions. Frequently occurring words in the captions include "satellite", "field", "building", "road", and "farm". The right figure shows the distribution of caption length in log scale. The distribution is long-tailed, and the average caption length is 49 words (maximum 3,248). The showcase of image-text pairs from PUB11 and RS3 and the statistics for image size can be found in Appendix \ref{appendix:visualization}.

\begin{figure}[htbp]
    \centering
    \includegraphics[width=0.15\textwidth]{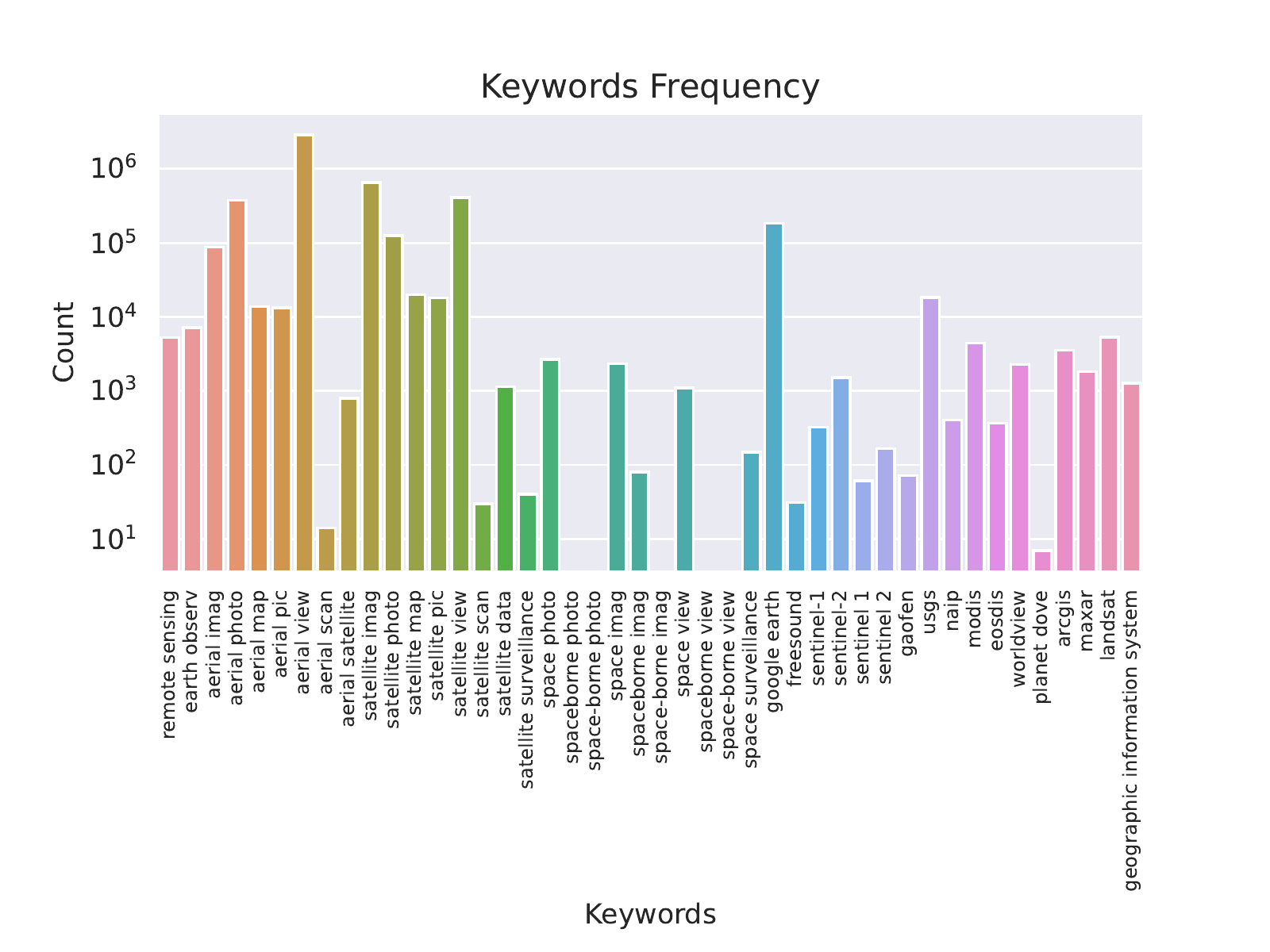}
    \includegraphics[width=0.15\textwidth]{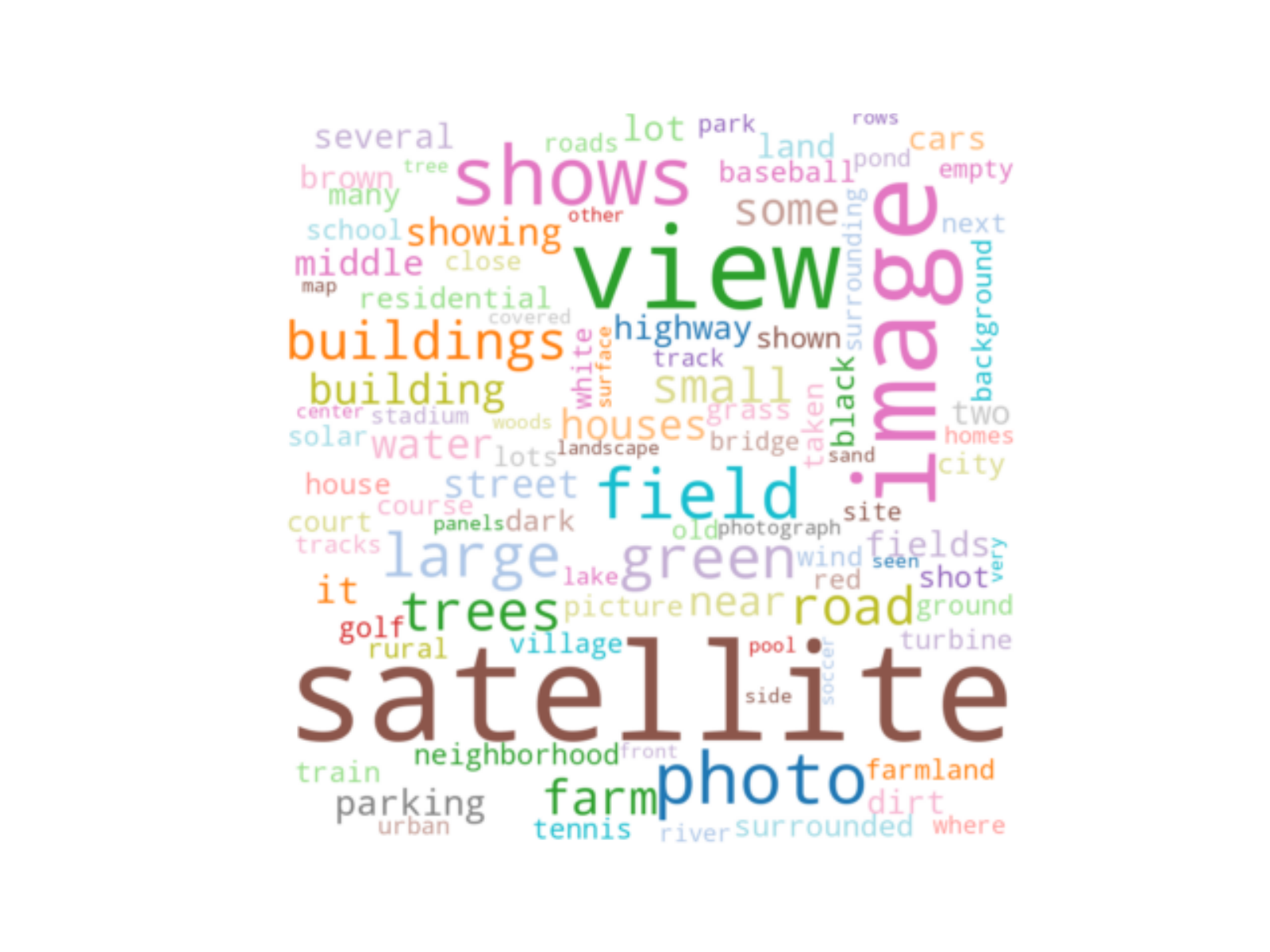}
    \includegraphics[width=0.15\textwidth]{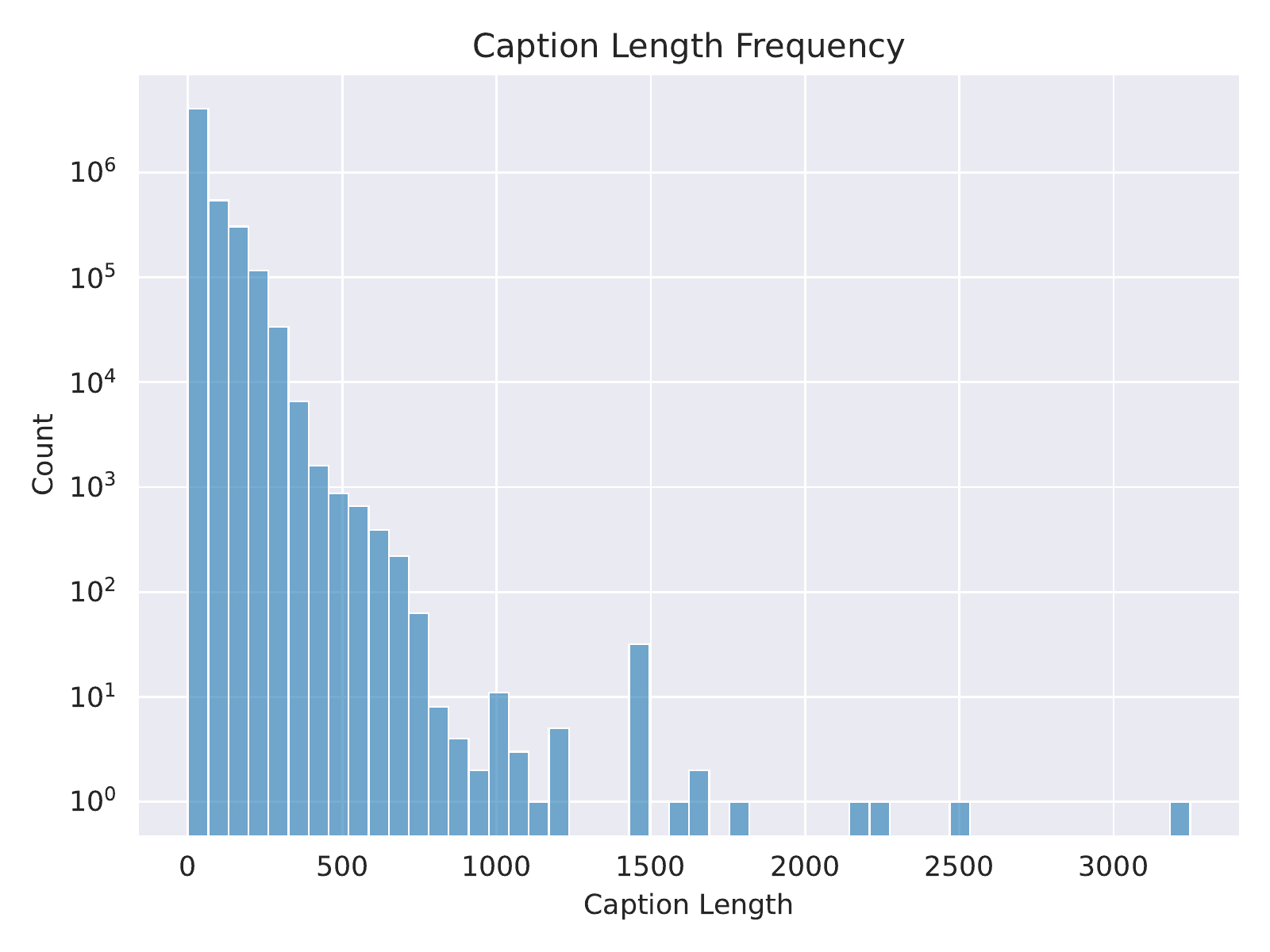}
    \caption{PUB11 Visualization}
    \label{fig:dataset_description}
\end{figure}

\begin{figure}[htbp]
    \centering
    \includegraphics[width=0.15\textwidth]{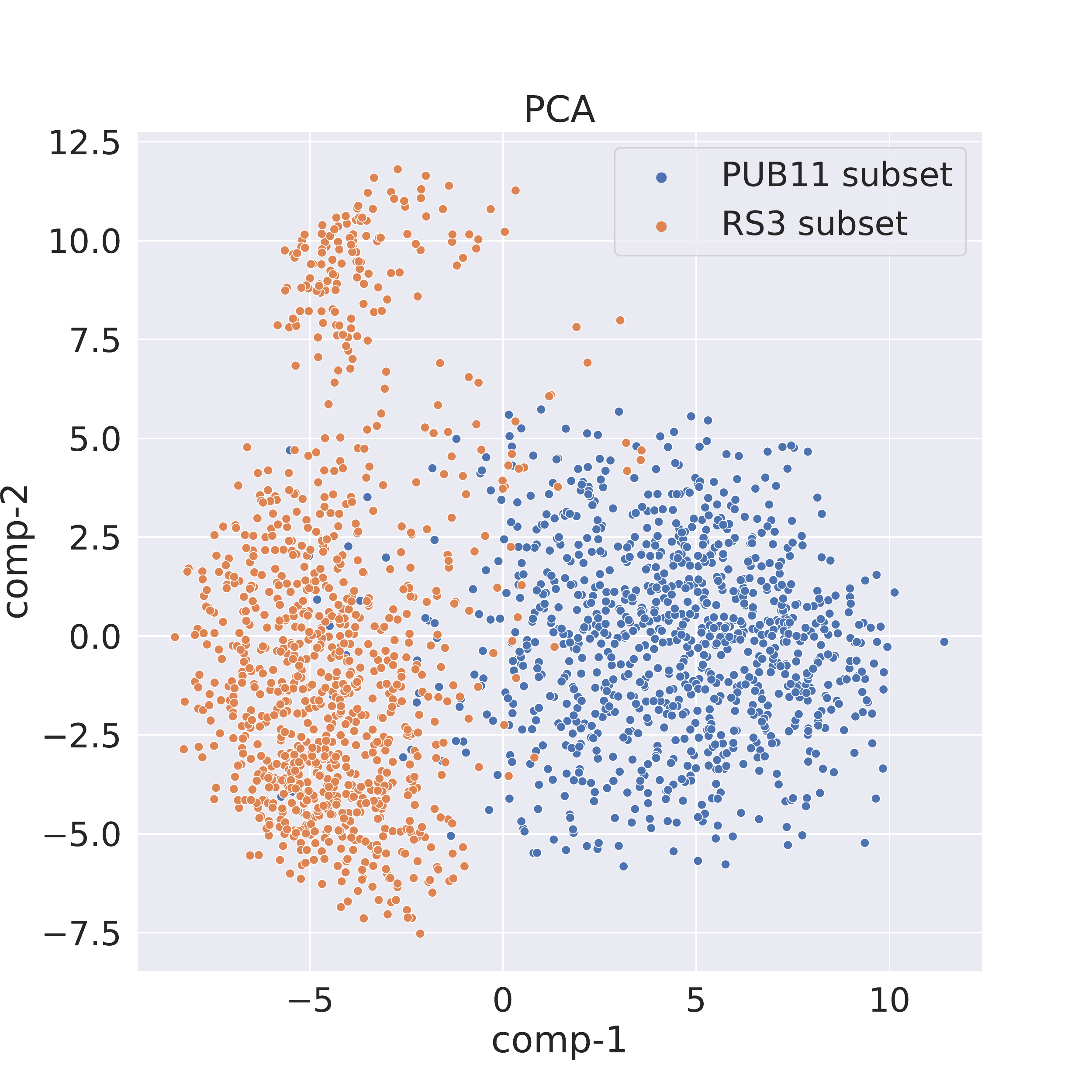}
    \includegraphics[width=0.15\textwidth]{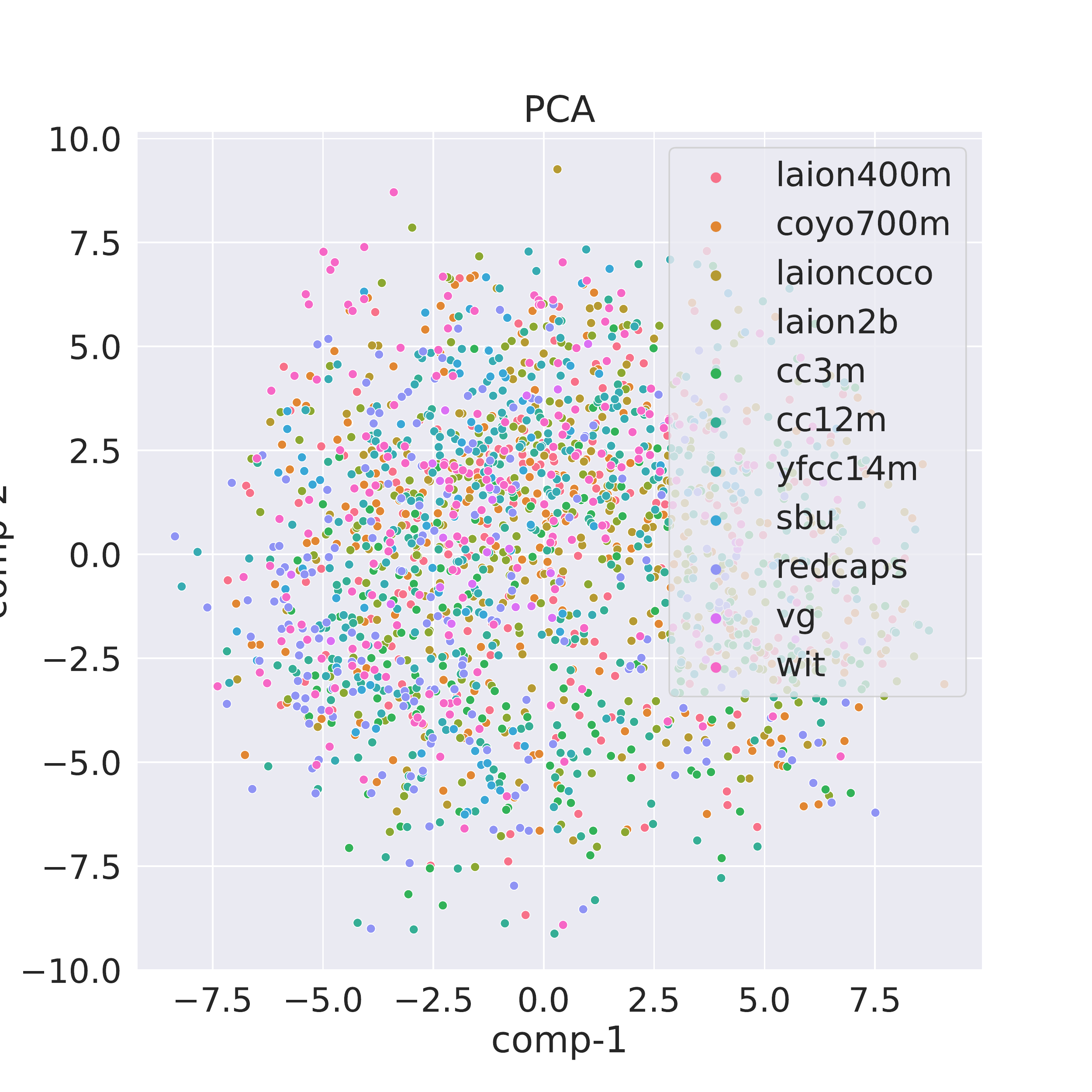}
    \includegraphics[width=0.15\textwidth]{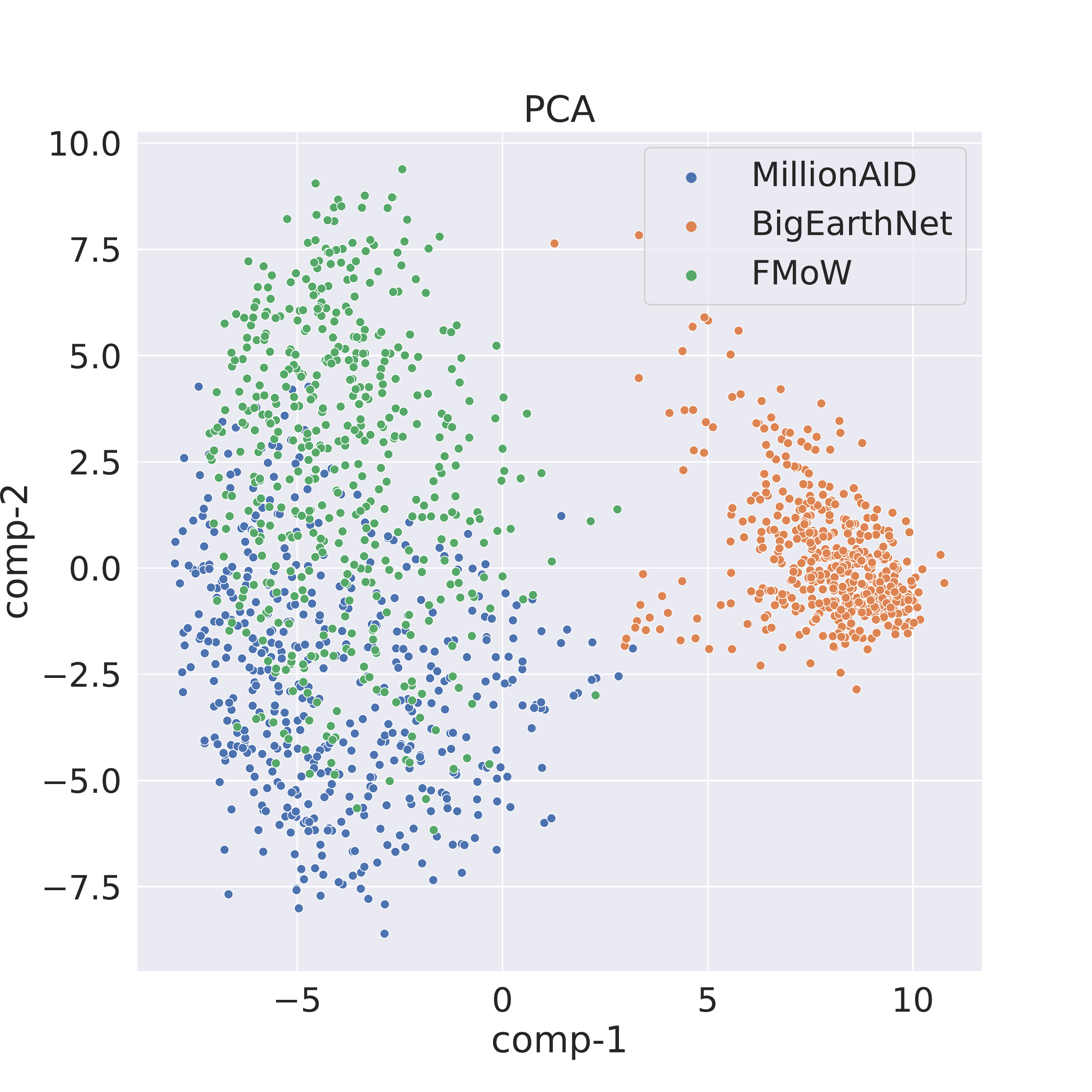}
    \caption{PCA. \textbf{Left}: PUB11 and RS3. \textbf{Middle}: 11 public datasets. \textbf{Right}: 3 RS datasets.}
    \label{fig:pca}
\end{figure}

 We then use CLIP's visual encoder (ConvNext-XXL Visual Encoder from OpenCLIP's implementation) to extract image features from PUB11 and RS3, visualizing the results using PCA. We sampled 1,000 images equally from PUB11 and RS3. Figure \ref{fig:pca} left demonstrates the discriminative domain differences between PUB11 and RS3, possibly due to the massive amount of aerial images in PUB11 and satellite images in RS3. Figure \ref{fig:pca} middle displays the PCA visualization for 2,200 samples from the 11 datasets in PUB11. Interestingly, no significant domain differences are observed among the RS images from them, as the data points are intermingled. Figure \ref{fig:pca} right reveals a clear separation between BigEarthNet and the other two datasets (500 examples for each), which may be attributed to the lower resolution (120 $\times$ 120) of all BigEarthNet images compared to the higher resolutions of the other two datasets.
 
\section{Geographical Analysis and Potential Negative Societal Implications}
\begin{figure}
    \centering
    \includegraphics[width=0.45\textwidth]{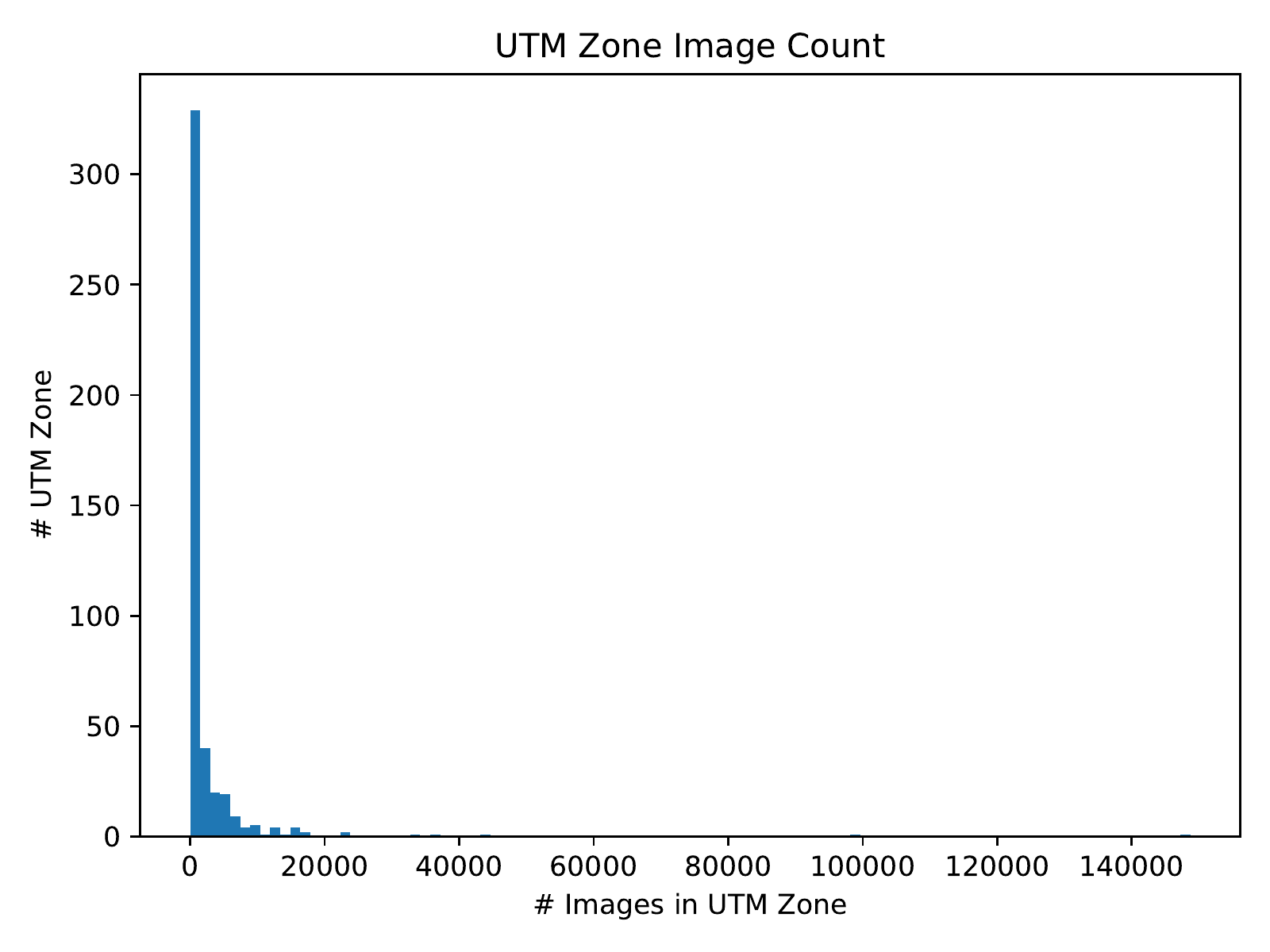}
    \caption{The distribution of images per UTM zone}
    \label{fig:geolocation_dist_hist}
\end{figure}

\begin{figure}
    \centering
    \includegraphics[width=0.45\textwidth]{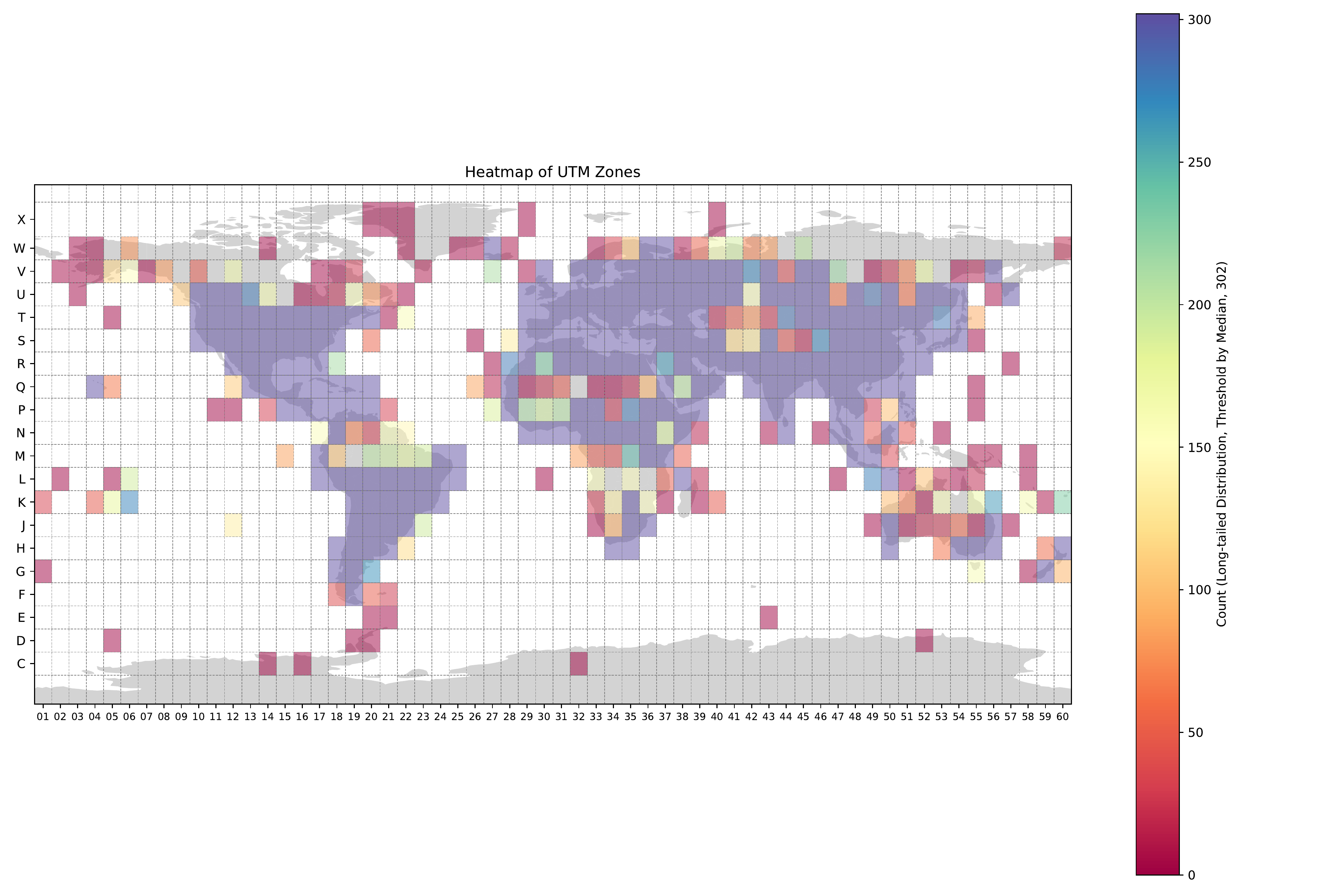}
    \caption{The distribution of geolocation for images in RS5M}
    \label{fig:geolocation_dist}
\end{figure}

In our dataset, there are two potential concerns. First is the data overrepresentation and underrepresentation problems in some parts of the world. We analyzed the geolocation information of images in our dataset (\textbf{based on 1,079,370 images with geo-information from Fmow, BEN, and YFCC}). Our analysis reveals a long-tailed distribution for the "number of images per UTM zone" statistics, as shown in Figure \ref{fig:geolocation_dist_hist}. In Figure \ref{fig:geolocation_dist}, image density (number of images per UTM zone) is sparse in Middle Africa (zones 29Q - 36Q) and Southern Africa (rectangle zone from 33M to 37K). This might be attributed to the presence of the Sahara Desert and the South African Plateau, which are less inhabited regions. Southern Indonesia and Australia (specifically the desert regions spanning a rectangle zone from 49L to 56H) exhibit low image density. However, an exception is observed in Southern Australia, characterized by its flat terrain and heightened human activity. Northern South America (rectangle zone from 19N to 23M) and Central Asia (rectangle zone from 40T to 44S) display a reduced distribution of images. The former is peculiar, as one would expect higher human activity in this region. Northern regions of Canada and Russia have a low image density, which is understandable given their proximity to the Arctic Circle. High image density is observed in North America, Europe, and most parts of Asia and South America. The low image density areas are overlapped with many underdevelopment areas and inhabitable areas, and this could bring bias into the model trained with RS5M. Second, the RS3 subset may contain wrong captions or misleading information, which could lead to mistakes that might have real-world consequences. 

Upon analysis, we discovered that the \textbf{captions from the PUB11 dataset} contain a significant amount of location information. As a result, we executed a NER (Named Entity Recognition) extraction on the PUB11 subset. The complete file was uploaded to the huggingface repo \footnote{huggingface.co/datasets/Zilun/RS5M}.

We hypothesize that the location information in the captions is closely related to the image's content and its shooting location. While this might introduce some noise, given that most PUB11 images originate from the internet and the paired text's purpose is to supplement the image, we believe most of the location data is useful.

We specifically extracted entities labeled as "GPE" (geopolitical entities). However, most of these entities are country or city names, not UTM zones or latitude/longitude details. While city names can be readily converted to UTM zones, captions containing only country names provide us with coarse spatial information. Nonetheless, this is a valuable addition to our analysis of RS5M's geographic distribution.

Out of the dataset, 880,354 images have captions with location information. We took the NER tool from the NLTK implementation. We also tried Stanford NER models, but the estimated processing time was 900 hours. In the future, we plan to develop an algorithm to convert extracted GPEs to UTM zones if applicable.

Compared with datasets converted from RS detection/segmentation datasets, our RS5M includes real-world geo-information such as geographical location, season information, etc., which could be helpful for GeoAI-related tasks such as geo-localization and GeoQA. 

\section{Experiment}

\subsection{Experiment Setup}
To verify the effectiveness of RS5M dataset, we conducted some experiments on training models using RS5M. We selected the CLIP ViT-B\/32, CLIP ViT-B\/16, CLIP ViT-L\/16, CLIP ViT-H\/14 model as the GVLM (mentioned in the introduction section). We fine-tuned the models with the RS5M dataset, and employed 4 different Parameter-Efficient Fine-Tuning (PEFT) methods as DVLM candidates on the CLIP ViT-B\/32: Pfeiffer adapter \cite{pfeiffer2021adapterfusion}, LoRA adapter \cite{hu2021lora}, Prefix-tuning adapter \cite{li2021prefixtuning}, and UniPELT adapter \cite{unipelt} (a vanilla adapter, a low-rank-approximated adapter, a prompt-based adapter, and a composite adapter). Since the downstream tasks in this paper only require image and text features, no DTM is needed. Then, for the RS3 subset, we randomly chose the rank 1 caption or rotationally invariant caption. 

We evaluated the domain generalizability of DVLM tuned by the RS5M dataset from 3 vision-language tasks: zero-shot classification (ZSC), remote sensing cross-modal text–image retrieval (RSCTIR), and semantic localization (SeLo). 

\begin{itemize}
    \item \textbf{Zero-shot Classification}. Thanks to the CLIP \cite{clip} model's strong image-text association capability, this task can be converted from any image classification dataset if the category names are provided. The model classifies the most relevant category for a given image. It's termed "zero-shot" because the test categories are unseen during training. The evaluation metric is accuracy.
    \item \textbf{Remote Sensing Cross-Modal Text–Image Retrieval} uses text/image to retrieve paired image/text. Pre-trained VLM mostly uses MSCOCO and flikcr30k datasets to evaluate the model. In the field of RS, UTMCaption, SydeneyCaption, RSICD and RSITMD datasets are frequently used to evaluate the model's VLR capability. Common metrics are recall@1/5/10 and mean recall.
     \item \textbf{Semantic Localization}, proposed in SeLo paper \cite{selo}. The SeLo task is defined as using cross-modal information such as text to locate semantically similar regions in large-scale RS scenes. SeLo achieves semantic-level retrieval with only caption-level annotation. It can be considered as a weak detection task without the need to label the bounding box in the training set. The metrics are $R_{su}$, $R_{as}$, $R_{da}$, and $R_{mi}$. The detailed mathematical definition will not be introduced here. $R_{su}$ aims to calculate the attention ratio of the ground-truth area to the non-GT area. $R_{as}$ attempts to quantify the shift distance of the attention from the GT center. $R_{da}$ evaluates the discreteness of the generated attention from probability divergence distance and candidate attention number. $R_{mi}$ is the comprehensive indicator above all. All of them are ranging from 0\~1, the higher the better, except $R_{as}$.
\end{itemize}

For ZSC, the complete AID\cite{aid}, RESISC45\cite{RESISC45}, and EuroSAT\cite{eurosat} datasets were selected. The RSICD and RSITMD datasets were chosen for VLR tasks. We adopted the test split given by Yuan et al. \cite{RSITMD} to align the setting of the previous works. Lastly, the AIR-SLT dataset was used for the SeLo task. We used top-1 accuracy to assess the ZSC task, recall@1/5/10/mean\_recall for evaluating VLR task, and $R_{su}$, $R_{as}$, $R_{da}$, $R_{mi}$ for the SeLo task.

We utilized the OpenCLIP implementation for GVLM. For DVLM, full fine-tuning and adapters with default parameters from the AdapterHub \footnote{https://docs.adapterhub.ml/classes/models/clip.html}) are adopted. The weights for CLIP were frozen except when using adapters. The mixed-precision (AMP) mode was used, and the AdamW optimizer \cite{adamw} was employed. Modality interaction was only through the InfoNCE loss \cite{infonce}. Learning rates were set to $1e^{-6}$ with weight decay set to $0.5$. A cosine learning rate scheduler was used, and the batch size was set to 700 for a single RTX A100 40GB. The training lasted for 20 epochs, and 5\% RS5M data (evenly drawn from each subset) were used as the validation set and the rest of them became the training set.

\subsection{Main Experiment Result}

Tables \ref{table:zsc_selo}  and \ref{table:i2t_t2i} report the baseline and fine-tuned results for VLR, ZSC, and SeLo tasks. The "CLIP-Baseline" method refers to the untuned CLIP model, which is used as GVLM-only approach. \textbf{ViT-B-32 model from OpenAI's CLIP is used unless we specify}. In Table \ref{table:zsc_selo}, the "SeLo-v1" \cite{selo} and "SeLo-v2" \cite{selov2} approaches were supervisedly trained using the RSITMD dataset's training set. In Table \ref{table:i2t_t2i}, the methods "VSE++" \cite{vse++}, "AFMFN"\cite{RSITMD}, "KCR" \cite{kcr}, "GaLR" \cite{galr}, "LW-MCR" \cite{LW-MCR}, "HVSA" \cite{hvsa}, "FAAMI" \cite{Zheng_2023}, "PIR" \cite{pir}, "RemoteCLIP" \cite{liu2023remoteclip}, "Multilanguage" \cite{multilingual}, "PE-RSITR"\cite{yuan2023parameterefficient}, and "MTGFE"\cite{MTGFE} are some competitive approaches. 

We used full fine-tuning or different adapters (LoRA, Pfeiffer, Prefixtuning, UniPELT) to implement the DVLM, which can be seen in the "Paradigm" column. \textbf{We named any model tuned with RS5M in this paper to be "GeoRSCLIP"}. 


\begin{table*}
\caption{Results for ZSC task and SeLo task. "GVLM" means only GVLM is used, "Supervised" represents the method was trained supervisedly on the labeled dataset, and "GVLM + $\bigtimes$" stands for the DVLM is applied on top of the GVLM, and the DVLM is implemented by the adapter tuned in RS5M, or full fine-tuning.}
\label{table:zsc_selo}
\centering
\small
\begin{tabular}{cccccccccc}
\toprule
\multicolumn{3}{c}{} & \multicolumn{3}{c}{\textbf{Zero-shot Classification}} & \multicolumn{4}{c}{\textbf{Semantic Localization}} \\
\midrule
\multicolumn{3}{c}{\textbf{}} & AID & RESISC45 & EuroSAT & \multicolumn{4}{c}{AIR-SLT} \\
\midrule
\textbf{\textbf{Method}} & \textbf{\textbf{Paradigm}} & \textbf{\textbf{Tuned on}} & \multicolumn{3}{c}{Top-1 Accuracy} & {$R_{su\uparrow}$} & {$R_{as \downarrow}$} & {$R_{da\uparrow}$} & {$R_{mi\uparrow}$} \\
\midrule
SeLo-v1 \cite{selo} & Supervised & RSITMD & - & - & - & 0.6920 & 0.3323 & 0.6667 & 0.6772\\
SeLo-v2 \cite{selov2} & Supervised & RSITMD & - & - & - & 0.7199 & 0.2925 & 0.6658 & 0.7021\\
CLIP-Baseline \cite{clip} & GVLM & - & 66.22  & 60.90 & 47.21 &0.7188 &	0.3006	& 0.6992 &	0.7071  \\
\textbf{GeoRSCLIP} & GVLM + Pfeiffer & RS5M  & 72.76 & 68.07 & 61.41 & 0.7402 & \textbf{0.2541}  & 0.6948 & 0.7308\\
\textbf{GeoRSCLIP}& GVLM + Prefix\-tuning & RS5M  & 69.17 & 63.11 & 61.59 & 0.7440  & 0.2551 & 0.7013 & 0.7336 \\
\textbf{GeoRSCLIP}& GVLM + LoRA & RS5M & 72.33 & 66.82 & 60.10 & 0.7461 & 0.2642  & 0.6636 & 0.7218\\
\textbf{GeoRSCLIP}& GVLM + UniPELT & RS5M  & 71.51 & 65.37 & 60.46 & 0.7489  & 0.2550 & 0.7021 & 0.7358\\
\textbf{GeoRSCLIP} & GVLM + FT & RS5M & 73.72 & 71.89 & 61.49 & 0.7546	& 0.2610	 & 0.7180		& 0.7400\\
\textbf{GeoRSCLIP} (ViT-H-14) & GVLM + FT & RS5M & \textbf{76.33} & \textbf{73.83} & \textbf{67.47}  & \textbf{0.7595} &	0.2566	 & \textbf{0.7418} & \textbf{0.7494} \\
\bottomrule
\end{tabular}
\end{table*}

Table \ref{table:zsc_selo} demonstrates that the CLIP-based methods, especially when fine-tuned (CLIP-FT), exhibit superior performance in ZSC across the AID, RESISC45, and EuroSAT datasets. The highest top-1 accuracy is achieved by the CLIP-FT (ViT-H-14) model, indicating the effectiveness of fine-tuning larger models on specialized datasets like RS5M. Models enhanced with DVLM (like CLIP-Pfeiffer, CLIP-Prefix-tuning, CLIP-LoRA, and CLIP-UniPELT) also show notable improvements over the baseline, signifying the benefits of combining GVLM with DVLM. In SeLo, the highest scores for $R_{su}$, $R_{da}$, and $R_{mi}$ are again observed in the CLIP-FT (ViT-H-14) variant, suggesting its robustness in localizing semantic elements. Interestingly, the lower $R_{as}$ score in CLIP-Pfeiffer highlights its performance in semantic localization. Across the SeLo metrics, the DVLM-enhanced CLIP models generally outperform the baseline and purely supervised models, demonstrating the value of fine-tuning with task-specific datasets. The RS5M dataset's role as a tuning dataset for DVLM implementations illustrates its potential to enhance model performance for both ZSC and SeLo tasks. The results suggest that larger, fine-tuned models can more effectively leverage the rich information present in specialized datasets like RS5M.

\begin{table*}
\caption{Results for image-to-text and text-to-image retrieval tasks. $ \dagger $ represents the results tested on RSICD test dataset, and $\S$ stands for the results tested on the RSITMD test dataset. Recall@1/5/10 and mean recall are computed. We used full fine-tuning (FT) or different adapters (LoRA, Pfeiffer, Prefixtuning, UniPELT) to implement the DVLM, which can be seen in the "Paradigm" column. We named any model related with RS5M in this paper to be "GeoRSCLIP". The GeoRSCLIP-FT model is a GeoRSCLIP model fine-tuned on the training set of the RSICD, the RSITMD or the RET-2 (combineation of the two, with data deduplication to avoid data leakage following \cite{liu2023remoteclip}), which shows the superior performance on each retrieval task.}



\label{table:i2t_t2i}
\centering
\footnotesize
\setlength{\tabcolsep}{7pt}
\begin{tabular}{cccccccccc}
\toprule
\multicolumn{3}{c}{} & \multicolumn{3}{c}{\textbf{Image-to-Text Retrieval}} & \multicolumn{3}{c}{\textbf{Text-to-Image Retrieval}} & \multicolumn{1}{c}{} \\
\cmidrule(rl){4-6} \cmidrule(rl){7-9}
\textbf{Method}  & \textbf{Paradigm} & \textbf{Tuned on} & {R@1} & {R@5} & {R@10} & {R@1} & {R@5} & {R@10} & {mR} \\
\midrule
LW-MCR \cite{LW-MCR} $ \dagger $ & Supervised & RSICD&3.29  & 12.52  & 19.93  &  4.66  & 17.51   & 30.02  &  14.66 \\ 
VSE++ \cite{vse++} $ \dagger $ & Supervised & RSICD &  3.38  & 9.51  & 17.46  & 2.82  & 11.32  & 18.10  & 10.43  \\
AFMFN \cite{RSITMD} $ \dagger $& Supervised & RSICD &  5.39  & 15.08  & 23.40  & 4.90  & 18.28  & 31.44  & 16.42  \\
KCR \cite{kcr} $ \dagger $ & Supervised & RSICD &  5.84  & 22.31  & 36.12  & 4.76  & 18.59  & 27.20  & 19.14  \\
GaLR \cite{galr} $ \dagger $ & Supervised & RSICD &  6.59  & 19.85  & 31.04  & 4.69  & 19.48  & 32.13  & 18.96  \\
SWAN $ \dagger $ & Supervised & RSICD & 7.41  & 20.13  & 30.86  & 5.56  & 22.26  & 37.41  & 20.61  \\
HVSA \cite{hvsa} $ \dagger $ & Supervised & RSICD&7.47 &20.62 &32.11 &5.51 &21.13 &34.13 & 20.16  \\ 
PIR \cite{pir} $ \dagger $ & Supervised & RSICD& 9.88   & 27.26   & 39.16  & 6.97   & 24.56   & 38.92  & 24.46 \\ 
FAAMI \cite{Zheng_2023} $ \dagger $ & Supervised & RSICD& 10.44  & 22.66  & 30.89  & 8.11  &  25.59  & 41.37  & 23.18  \\ 
Multilanguage \cite{multilingual} $ \dagger $ &Supervised & RSICD &10.70 & 29.64& 41.53& 9.14&28.96 &44.59 & 27.42\\
PE-RSITR \cite{yuan2023parameterefficient} $ \dagger $ &GVLM + FT& RSICD & 14.13 & 31.51 & 44.78 & 11.63 & 33.92 & 50.73 & 31.12\\
MTGFE \cite{MTGFE} $ \dagger $ &Supervised& RSICD & 15.28& 37.05 & 51.60 & 8.67& 27.56& 43.92 & 30.68\\
RemoteCLIP \cite{liu2023remoteclip} $ \dagger $ & GVLM + FT & RET-3 + DET-10 + SEG-4 & \textbf{17.02} & 37.97& 51.51& \textbf{13.71} & 37.11 & 54.25 & \textbf{35.26} \\
CLIP-Baseline \cite{clip} $ \dagger $ & GVLM & - &	5.31  & 14.18  & 23.70  & 5.78  &	17.73  & 27.76  & 15.74   \\
\textbf{GeoRSCLIP} $ \dagger $ &GVLM + UniPELT & RS5M & 8.97  &23.79  &  35.04  &7.52  & 23.35  & 35.66  & 22.39 \\
\textbf{GeoRSCLIP} $ \dagger $ &GVLM + Prefix\-tuning & RS5M & 9.06  & 20.04  & 29.73  & 5.95  & 20.37   & 32.99   & 19.69 \\
\textbf{GeoRSCLIP} $ \dagger $ &GVLM + LoRA& RS5M & 9.52   & 21.13  & 31.75  &  6.73  & 23.24  & 35.59  & 21.33  \\
\textbf{GeoRSCLIP} $ \dagger $  &GVLM + Pfeiffer & RS5M &  10.61  & 24.79  & 36.51  & 8.12  & 24.81   & 38.79  & 23.94 \\

\textbf{GeoRSCLIP} $ \dagger $ & GVLM + FT & RS5M & 11.53   &	28.55  	& 39.16   & 	9.52   & 	27.37  	& 40.99   &	26.18  \\

\textbf{GeoRSCLIP-FT} $ \dagger $ & GVLM + FT & RS5M + RSICD &\textbf{22.14} & 40.53 & 51.78&\textbf{15.26}&40.46& 57.79 & \textbf{38.00} \\
\textbf{GeoRSCLIP-FT} $ \dagger $ & GVLM + FT & RS5M + RET-2 & \textbf{21.13}& 41.72& 55.63&\textbf{15.59} & 41.19 &57.99 & \textbf{38.87} \\
\midrule
LW-MCR \cite{LW-MCR} $ \S $ & Supervised & RSITMD&10.18  & 28.98  & 39.82  & 7.79  & 30.18  & 49.78  & 27.79 \\ 
VSE++ \cite{vse++} $ \S $& Supervised & RSITMD & 10.38  & 27.65  & 39.60  & 7.79  & 24.87  & 38.67  & 24.83  \\
AFMFN \cite{RSITMD} $ \S $& Supervised & RSITMD & 11.06  & 29.20  & 38.72  & 9.96  & 34.03  & 52.96  & 29.32  \\
HVSA \cite{hvsa} $ \S $ & Supervised & RSITMD&13.20  &32.08  &45.58  &11.43  &39.20  &57.45  & 33.15  \\ 
SWAN $\S $ & Supervised & RSITMD & 13.35  & 32.15  & 46.90  & 11.24  & 40.40  & 60.60  & 34.11  \\
GaLR \cite{galr} $ \S $& Supervised & RSITMD & 14.82  & 31.64  & 42.48  & 11.15  & 36.68  & 51.68  & 31.41  \\
FAAMI \cite{Zheng_2023} $ \S $ & Supervised & RSITMD & 16.15  & 35.62  & 48.89  & 12.96   & 42.39  & 59.96  & 35.99  \\ 
MTGFE \cite{MTGFE} $ \S $ &Supervised& RSITMD & 17.92 & 40.93 & 53.32& 16.59&48.50 & 67.43 & 40.78\\
PIR \cite{pir} $ \S $ & Supervised & RSITMD & 18.14  & 41.15  &52.88  & 12.17  & 41.68  & 63.41   & 38.24 \\ 
Multilanguage \cite{multilingual} $ \S $ &Supervised& RSITMD & 19.69& 40.26& 54.42& 17.61& 49.73&66.59 & 41.38\\
PE-RSITR \cite{yuan2023parameterefficient} $ \S $ & GVLM + FT & RSITMD & 23.67 & 44.07 & 60.36& 20.10&50.63 &67.97 & 44.47\\
RemoteCLIP \cite{liu2023remoteclip} $ \S $ & GVLM + FT & RET-3 + DET-10 + SEG-4 & \textbf{27.88} & 50.66 & 65.71 & \textbf{22.17} & 56.46 & 73.41 & \textbf{49.38}\\
CLIP-Baseline \cite{clip} $ \S $ & GVLM & - & 9.51 	& 23.01 &  	32.74 	& 8.81 	& 27.88  & 	43.19  & 	24.19  \\
\textbf{GeoRSCLIP} $ \S $ &GVLM + Prefix\-tuning & RS5M & 15.93  &31.86  &41.81 & 11.42  & 34.20  &50.88  &31.02 \\
\textbf{GeoRSCLIP} $ \S $ &GVLM + UniPELT& RS5M & 16.15  & 31.86   & 42.04  & 11.68  &  36.15  & 52.70   &31.76  \\
\textbf{GeoRSCLIP} $ \S $ &GVLM + LoRA& RS5M &  16.37  & 32.96  & 42.70  & 12.92   &34.87  & 50.88   &31.78  \\
\textbf{GeoRSCLIP} $ \S $  &GVLM + Pfeiffer & RS5M & 16.81  &34.73  & 44.69   &  12.26  &  38.10  & 53.94  & 33.42 \\
\textbf{GeoRSCLIP} $ \S $ & GVLM + FT & RS5M & 19.03  & 	34.51  & 	46.46 	 & 14.16 	&  42.39  & 	57.52   & 	35.68 \\
\textbf{GeoRSCLIP-FT} $ \S $ & GVLM + FT & RS5M + RSITMD & \textbf{30.09} & 51.55 & 63.27 & \textbf{23.54} &57.52&74.6 & \textbf{50.10}  \\
\textbf{GeoRSCLIP-FT} $ \S $ & GVLM + FT & RS5M + RET-2 &\textbf{32.30} &53.32&67.921&\textbf{25.04}&57.88&74.38 & \textbf{51.81}  \\
\bottomrule
\end{tabular}
\end{table*}

In Table \ref{table:i2t_t2i}, we demonstrate the retrieval performance across both datasets (RSICD and RSITMD) and both tasks (image-to-text and text-to-image retrieval). This again indicates the effectiveness of GeoRSCLIP models. Moreover, the results show that models tuned with the RS5M dataset generally exhibit better performance compared to most of the baseline and other supervised methods. This suggests that the RS5M dataset can provide valuable domain-specific information that enhances model performance in RSCTIR tasks. 

In our trials leveraging PEFT approaches, a noteworthy observation is the methodology's inherent advantage in requiring less extensive hyperparameter tuning, which significantly reduces the computational overhead. However, it is essential to acknowledge that they do not consistently achieve the same level of effectiveness as observed in comprehensive full fine-tuning models. Despite this, the performance of PEFT methods remains competitive, particularly when juxtaposed with certain supervised learning techniques, indicating a balance between efficiency and efficacy. 

Moreover, the fact that RS5M does not include any part of the RSITMD and RSICD training sets. Not like other approaches that trained on the training set of RSITMD or RSICD, RS5M does not necessarily follow the data distribution of RSITMD and RSICD, that is, models trained on RS5M and tested on RSITMD or RSICD are being evaluated on data that is entirely new to them (trained on RSITMD and RSICD training sets might have an advantage due to their exposure to similar data during training). 
Even the training dataset that doesn't overlap with common training sets like RSITMD and RSICD, the tuned model can still obtain competitive results on the evaluation set. This scenario can be seen as a robust test of a large model's ability to generalize and perform well on unseen data, which is crucial in real-world applications. 

Finally, when the training set of the RSICD, the RSITMD dataset, or the RET-2 dataset (a combination of the previous two, with data deduplication to avoid data leakage following \cite{liu2023remoteclip}) is used to further tune the GeoRSCLIP model for 1-3 epochs, the superior performance of GeoRSCLIP on each retrieval task is observed, achieving the state-of-the-art in this task, showing that GeoRSCLIP is a strong fine-tuner.

\subsection{Ablation Study}
CLIP’s ViT-B-32 model is used to train the GeoRSCLIP (unless we specify). 
\subsubsection{Per Subset Analysis}
\label{per_subset_analysis}
The RS5M dataset comprises two components: PUB11 and RS3. Given that PUB11 predominantly contains aerial images and RS3 contains only satellite images, we decided to analyze them separately. This approach allows us to ascertain the individual contributions of each subset, particularly in understanding the potential impact of training the model with a large number of aerial images on satellite-image-based downstream tasks. Besides, we developed another subset of RS5M, denoted as Geometa, with 1,036,734 geo-information riched image-text pairs, consisting of data from FMoW, BigEartherNet, and YFCC14M. The geo-information includes the target class label, location of the image, date of taken, etc.

\begin{table}[ht]
\caption{Results for ZSC task, RSCTIR task, and SeLo task with models trained by subsets of RS5M.}
\label{table:rs5msubset}
\footnotesize
\setlength{\tabcolsep}{1.75pt}
\begin{tabular}{cccccccc}
\toprule
& \multicolumn{3}{c}{\textbf{Zero-shot Classification}} & \multicolumn{4}{c}{\textbf{Semantic Localization}} \\
\midrule
\textbf{Subset} & AID & RESISC45 & EuroSAT & $R_{su\uparrow}$ & $R_{as\downarrow}$ & $R_{da\uparrow}$ & $R_{mi\uparrow}$ \\
\midrule
 RS5M & \textbf{73.72} &  \textbf{71.89} & \textbf{61.49} & \textbf{0.7546} & 0.2610 &\textbf{0.7180} &\textbf{0.7400} \\
 PUB11 &72.41&68.35&60.94	&0.7447& 0.2895	&	0.6440	& 0.7076\\
 RS3 &73.01&71.41&51.70&0.7492&\textbf{0.2534}&0.7158&0.7399 \\
 Geometa &64.76 &57.43 &55.15 &0.7068 & 0.3157 & 0.6848 & 0.6934\\
\midrule
\midrule
RSITMD & \multicolumn{3}{c}{\textbf{Image-to-Text Retrieval}} & \multicolumn{3}{c}{\textbf{Text-to-Image Retrieval}} & \\
\midrule
\textbf{Subset} &  {R@1} & {R@5} & {R@10} & {R@1} & {R@5} & {R@10} & {mR} \\
 \midrule
  RS5M & \textbf{19.03} & 	34.51 & 	46.46	 & \textbf{14.16}	&  \textbf{42.39} & 	\textbf{57.52}  & 	\textbf{35.68}\\
PUB11 &13.27&	31.86	&42.04	&12.70	&34.38	&49.82& 30.68\\
RS3 &17.26 &	\textbf{34.96}&	\textbf{48.45}	&13.94 &	40.58 &	55.93	&35.18\\
Geometa &9.29	&22.57	&33.63	&6.99	&23.81 &	37.52	&22.30\\
\midrule
RSICD & \multicolumn{3}{c}{\textbf{Image-to-Text Retrieval}} & \multicolumn{3}{c}{\textbf{Text-to-Image Retrieval}} & \\
\midrule
\textbf{Subset} &  {R@1} & {R@5} & {R@10} & {R@1} & {R@5} & {R@10} & {mR} \\
 \midrule
  RS5M & 11.53  &	\textbf{28.55} 	& \textbf{39.16}  & 	9.52  & 	\textbf{27.37} 	& \textbf{40.99}  &	\textbf{26.18} \\
PUB11 &7.69&	19.03	&29.92 &	7.45&	22.58&	34.40& 	20.18\\
RS3 &\textbf{12.17}	&26.53&	37.51	&\textbf{9.59}	&27.10 &	40.51 &	25.57\\
Geometa &6.68 &	17.38 &	25.16 &	4.54 &	16.21	&26.09	&16.01\\
\bottomrule
\end{tabular}
\end{table}

Analyzing the Table \ref{table:rs5msubset}, it is evident that models trained on the RS5M dataset outperform those trained on the subsets PUB11, RS3, or Geometa across various tasks such as Zero-shot Classification, Semantic Localization, and Image-to-Text/Text-to-Image Retrieval in most cases. Specifically, in Zero-shot Classification, RS5M achieves the highest scores in AID, RESISC45, and EuroSAT, and in Semantic Localization, it leads in most of the metrics including $R_{su\uparrow}$, $R_{da\uparrow}$, and $R_{mi\uparrow}$. This trend extends to the retrieval tasks, where RS5M consistently shows superior performance in both Image-to-Text and Text-to-Image retrieval across all Recall metrics.

The RS5M's robust performance can be attributed to its ability to amalgamate the strengths of the subsets. PUB11's advantage in zero-shot classification and RS3's capability in retrieval tasks are combined in RS5M, making it a more versatile and comprehensive dataset. This integration not only leverages the individual advantages of PUB11 and RS3 but also mitigates their respective limitations. In contrast, the Geometa subset underperforms, possibly due to the limited variety in its prompt templates, which is a significant drawback for CLIP-based models. The dataset's vast size (around 1 million pairs) is not effectively utilized due to the limited diversity in language patterns, as it randomly selects from a pool of only a few hundred prompts. This lack of linguistic diversity fails to adequately challenge and train the CLIP models solely, leading to poorer performance.

Furthermore, the observations highlight that the use of generated captions (MillionAID) and web data (PUB11) effectively addresses the diversity issue inherent in captions based solely on geographic information (Geometa subset). This approach introduces a richer linguistic context, crucial for the success of CLIP-based models, thus enhancing their performance and generalizability. The RS5M dataset's comprehensive nature and its combination of various data types, including web data and generated captions, provide it with a clear advantage over the individual subsets, particularly in the context of tasks requiring a nuanced understanding of complex visual-textual relationships.

\subsubsection{Influence of RS5M Scale}

To verify if the current scale of data in RS5M is necessary, we explored the impact of dataset size on model performance using the RS5M dataset. To assess this, we trained models on randomly selected subsets of the RS5M dataset, specifically using the full dataset, as well as subsets representing 1/2, 1/4, and 1/8 of the total data. The effectiveness of each model was evaluated based on 8 factors, which included AID zero-shot accuracy, RESISC45 zero-shot accuracy, EuroSAT zero-shot accuracy, image-to-text recall@1 and text-to-image recall@1 for both RSITMD and RSICD, and the SeLo's $R_{mi}$ indicator.

\begin{figure}
    \centering
    \includegraphics[width=0.48\textwidth]{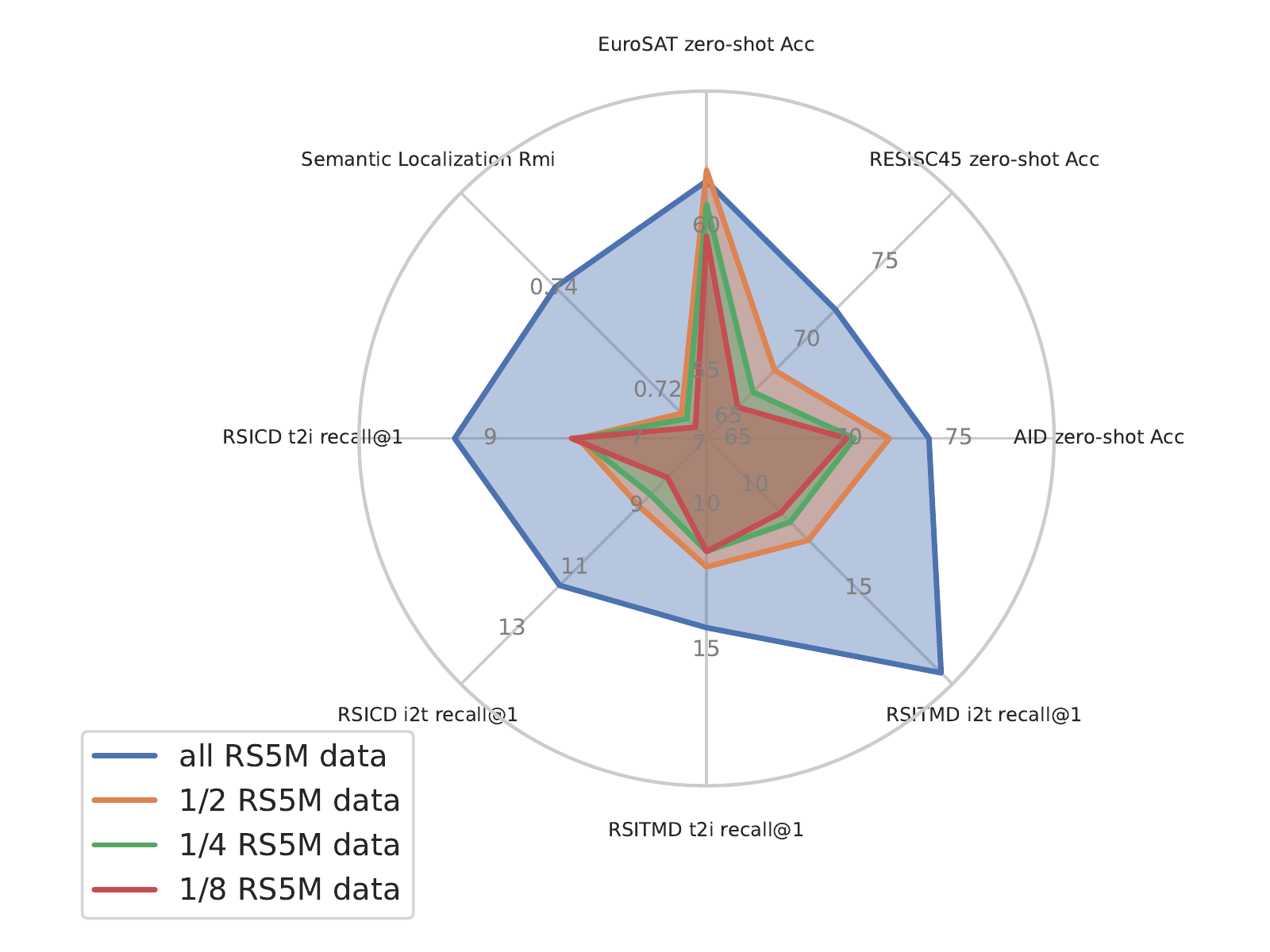}
    \caption{Influence of Dataset Scale}
    \label{fig:data_amount}
\end{figure}

To visualize and compare the performances across these varying dataset sizes, we employed a radar chart, as shown in Figure \ref{fig:data_amount}. The radar chart distinctly revealed a consistent trend: as the amount of training data increased, there was a corresponding and noticeable enhancement in model performance across all selected factors. This trend was observed uniformly in the zero-shot accuracy metrics for AID, RESISC45, and EuroSAT, as well as in the recall metrics for both RSITMD and RSICD tasks. Furthermore, the SeLo's $R_{mi}$ indicator, a comprehensive metric that indicates the overall performance in the SeLo task, also demonstrated significant improvement with the increment in data size. This experimental outcome robustly underscores the critical role of dataset volume in training models with better performance.

\subsubsection{Influence of Image Normalization}

Image normalization is pivotal in model tuning, particularly when bridging a significant domain gap between the pre-trained and fine-tuned datasets. Given the distinct nature of remote sensing imagery compared to common object domains, we investigated the impact of varying normalization parameters derived from different pre-training datasets on the evaluation outcomes. We computed the normalization parameters (mean and standard deviation) for the RS5M dataset using a moving average approach. Additionally, we take the statistical parameters from WIT (the dataset used for pre-training CLIP) and ImageNet into account. The values of these parameters are outlined in Table \ref{table:norm_param}.

\begin{table}[ht]
\caption{Different Normalization Parameters for WIT (CLIP), ImageNet, and RS5M}
\label{table:norm_param}
\centering
\begin{tabular}{|c|c|c|}
\toprule
Dataset Name & Mean & Standard Deviation \\
\midrule
WIT (CLIP) &[0.481, 0.458, 0.408]	& [0.269, 0.261, 0.276] \\
\midrule
ImageNet & [0.485, 0.456, 0.406]	& [0.229, 0.224, 0.225] \\
\midrule
RS5M  & [0.445, 0.469, 0.441]	& [0.208, 0.193, 0.213] \\
\bottomrule
\end{tabular}
\end{table}

\begin{table}[ht]
\caption{Results for ZSC task, RSCTIR task, and SeLo task with models trained using RS5M with different image normalizations.}
\label{table:imagenorm}
\footnotesize
\setlength{\tabcolsep}{1.75pt}
\begin{tabular}{cccccccc}
\toprule
& \multicolumn{3}{c}{\textbf{Zero-shot Classification}} & \multicolumn{4}{c}{\textbf{Semantic Localization}} \\
\midrule
\textbf{Image Norm} & AID & RESISC45 & EuroSAT & $R_{su\uparrow}$ & $R_{as\downarrow}$ & $R_{da\uparrow}$ & $R_{mi\uparrow}$ \\
\midrule
 WIT (CLIP) & 73.72 & 71.89 & \textbf{61.49} & 0.7546 & 0.2610 &0.7180 &0.7400 \\
 ImageNet &75.48&72.14&57.66	&\textbf{0.7583} & 0.2596	&\textbf{0.7304}		&0.7451\\
 RS5M &\textbf{75.71}&\textbf{72.61}&58.19 &0.7548&\textbf{0.2512}&0.7269&\textbf{0.7457} \\
\midrule
\midrule
RSITMD & \multicolumn{3}{c}{\textbf{Image-to-Text Retrieval}} & \multicolumn{3}{c}{\textbf{Text-to-Image Retrieval}} & \\
\midrule
\textbf{Subset} &  {R@1} & {R@5} & {R@10} & {R@1} & {R@5} & {R@10} & {mR} \\
 \midrule
  WIT (CLIP) & 19.03 & 	34.51 & 	46.46	 & 14.16	&  \textbf{42.39} & 	\textbf{57.52}  & 	35.68\\
 ImageNet &19.03	&35.84	&\textbf{47.35}	&14.42	&42.26	&57.04	&35.99	\\
 RS5M &\textbf{19.47}	&\textbf{36.95}	&47.12	&\textbf{14.60}	&41.95	&57.04	&\textbf{36.19} \\
\midrule
RSICD & \multicolumn{3}{c}{\textbf{Image-to-Text Retrieval}} & \multicolumn{3}{c}{\textbf{Text-to-Image Retrieval}} & \\
\midrule
\textbf{Subset} &  {R@1} & {R@5} & {R@10} & {R@1} & {R@5} & {R@10} & {mR} \\
 \midrule
  WIT (CLIP) & 11.53  &	\textbf{28.55} 	& 39.16  & 	9.52  & 	27.37 	& 40.99  &	26.18 \\
 ImageNet &\textbf{12.35}	&\textbf{28.73}	&40.26	&9.57	&26.72	&41.08	&26.45\\
 RS5M &11.62	&28.18	&\textbf{40.53}&	\textbf{9.57}&	\textbf{27.43}&	\textbf{41.61} & \textbf{26.49} \\
\bottomrule
\end{tabular}
\end{table}


As illustrated in Table \ref{table:imagenorm}, the result of using RS5M's normalization parameters shows a slight advantage among others. However, it is important to note that the differences in results attributed to varying normalization parameters are extremely marginal. The overall impact of using these three sets of parameters is subtle.

\subsubsection{Influence of Noisy Level in PUB11} 
\begin{figure}
    \centering
    \includegraphics[width=0.48\textwidth]{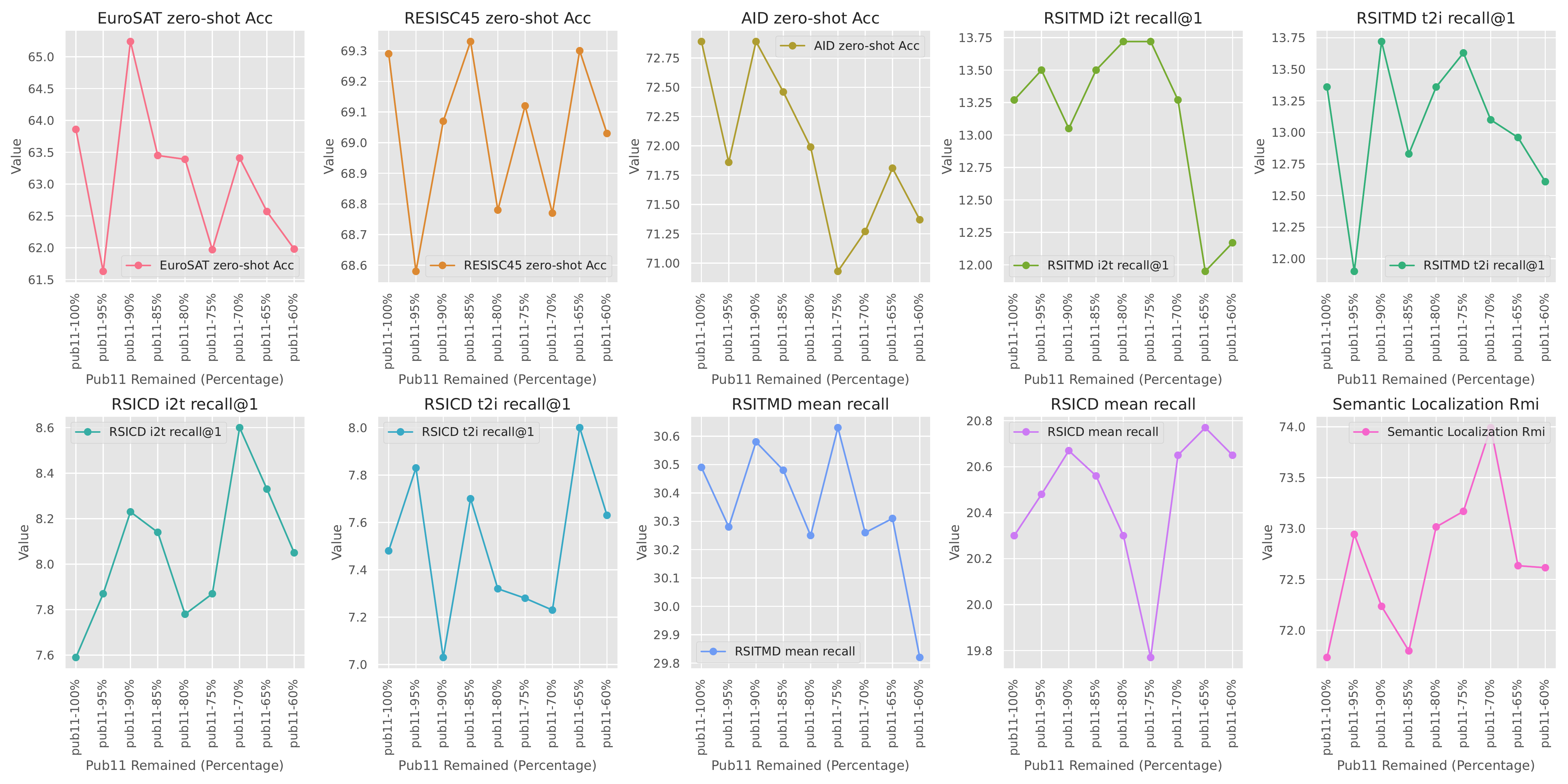}
    \caption{Influence of Noisy Level in PUB11}
    \label{fig:noise_level}
\end{figure}

As shown in Table \ref{appendix:pubdataset_stat} and Figure \ref{fig:num_image_heatmap}, approximately 1 million image-text pairs were filtered out by the VLM Filter and RS image detector using top 90\% $s_i$ and top 80\% $c_i$ as thresholding parameters. These parameters play a critical role in regulating the noise level of the PUB11 subset. Theoretically, reducing the values of $s_i$ and $c_i$ (i.e., retaining only image-text pairs which have the top 60\% $s_i$ and $c_i$ values) should result in lower noise levels. To empirically find the impact of noise levels in PUB11, we adjusted $s_i$ and $c_i$ values to generate more PUB11 subsets with varying noise levels. Subsequently, we fine-tuned the ViT-B-32 CLIP model using these subsets and evaluated on ZSC, VLR, and SeLo tasks. As Figure \ref{fig:noise_level} illustrates, a decrease in noise level generally led to an enhancement in the performance. However, the model's performance could potentially deteriorate if the PUB11 subset size is excessively reduced. The tuning points for most metrics are within the range of $s_i$ and $c_i$ from top 90\% to 80\%.

\subsubsection{Influence of Freezing Encoder}

To have a more comprehensive insight into the contributions of the textual and visual components within the CLIP framework during fine-tuning with the RS5M dataset, we conducted an experiment where either the text encoder or the image encoder was selectively frozen during the training process. This approach also help us evaluate the intrinsic quality of both the textual and visual content present in the RS5M dataset.

\begin{table}[ht]
\caption{Results for ZSC task, RSCTIR task, and SeLo task with tuned CLIP model which has frozen text encoder or image encoder.}
\label{table:freeze_encoder}
\footnotesize
\setlength{\tabcolsep}{1.0pt}
\begin{tabular}{cccccccc}
\toprule
& \multicolumn{3}{c}{\textbf{Zero-shot Classification}} & \multicolumn{4}{c}{\textbf{Semantic Localization}} \\
\midrule
\textbf{Freezed Enc} & AID & RESISC45 & EuroSAT & $R_{su\uparrow}$ & $R_{as\downarrow}$ & $R_{da\uparrow}$ & $R_{mi\uparrow}$ \\
\midrule
 Non & 73.72 & \textbf{71.89} & \textbf{61.49} & 0.7546 & 0.2610 &0.7180 &0.7400 \\
 Text & \textbf{74.06} & 71.58 & 60.37 & \textbf{0.7559} & \textbf{0.2570} &0.7144 & \textbf{0.7410}	 \\
 Image & 70.13 & 66.14 &50.81 &0.7349	&0.2670	&\textbf{0.7218}	&0.7310	  \\
\midrule
\midrule
RSITMD & \multicolumn{3}{c}{\textbf{Image-to-Text Retrieval}} & \multicolumn{3}{c}{\textbf{Text-to-Image Retrieval}} & \\
\midrule
\textbf{Subset} &  {R@1} & {R@5} & {R@10} & {R@1} & {R@5} & {R@10} & {mR} \\
 \midrule
 Non & \textbf{19.03} & 	\textbf{34.51} & 	\textbf{46.46}	 & \textbf{14.16}	&  \textbf{42.39} & 	\textbf{57.52}  & 	\textbf{35.68}\\
 Text &17.04	&33.19	&45.58	&13.05	&39.20	&55.22	&33.88\\
 Image  &15.71	&32.30	&46.02	&11.11	&32.52	&50.09	&31.29	\\
\midrule
RSICD & \multicolumn{3}{c}{\textbf{Image-to-Text Retrieval}} & \multicolumn{3}{c}{\textbf{Text-to-Image Retrieval}} & \\
\midrule
\textbf{Subset} &  {R@1} & {R@5} & {R@10} & {R@1} & {R@5} & {R@10} & {mR} \\
 \midrule
 Non & \textbf{11.53}  &	\textbf{28.55} 	& \textbf{39.16}  & 	\textbf{9.52}  & 	\textbf{27.37} 	& \textbf{40.99}  &	\textbf{26.18} \\
 Text &10.70	&25.98	&36.23	&9.09	&25.60	&39.41	&24.50\\
 Image  &9.06	&22.60	&33.94	&6.53	&20.46	&33.74	&21.06 \\
\bottomrule
\end{tabular}
\end{table}

As shown in Table \ref{table:freeze_encoder}, we displayed results for freezing either the image encoder or text encoder. All models are trained using the RS5M dataset. 

The model with a non-frozen encoder performs best in RESISC45 and EuroSAT, indicating the importance of cooperative learning between text and image encoders in these tasks. In the SeLo task, the non-frozen and text-frozen models are close, with the text-frozen model showing a marginally better performance.
The image-frozen model lags in most metrics implying the important role of the image encoder in this context. Across both RSITMD and RSICD tasks, the non-frozen encoder model consistently outperforms the other variants in all recall metrics. This indicates the criticality of dynamic interaction between the text and image encoders for effective retrieval performance. The performance drop is more pronounced in the image-frozen model compared to the text-frozen one, highlighting the image encoder's significant role in retrieval tasks.

\subsubsection{Influence of Model size}

In this section, we delve into the relationship between model size and performance in downstream tasks. We chose to compare models using CLIP with ViT-B-32, ViT-B-16, ViT-L-14, and ViT-H-14 encoders. All models were tuned on the RS5M dataset. We use OpenAI's implementation for ViT-B-32, ViT-B-16, and ViT-L-14; OpenCLIP's implementation for ViT-H-14.

\begin{figure}
    \centering
    \includegraphics[width=0.47\textwidth]{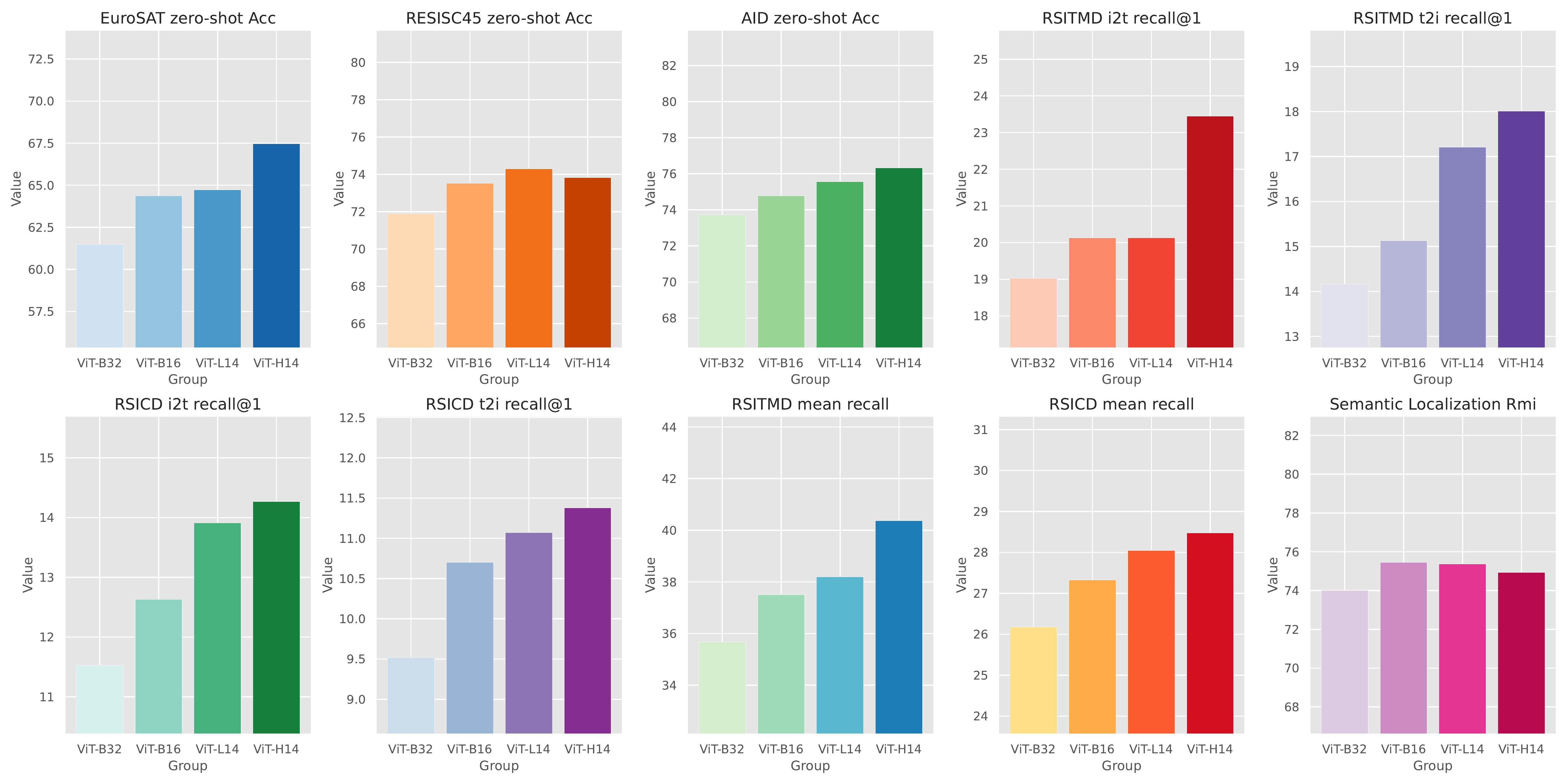}
    \caption{Influence of Model Size}
    \label{fig:model_size}
\end{figure}

As shown in Figure \ref{fig:model_size}, AID zero-shot accuracy, RESISC45 zero-shot accuracy, EuroSAT zero-shot accuracy, image-to-text recall@1, image-to-text mean recall, text-to-image recall@1, and text-to-image mean recall for both RSITMD and RSICD, and the SeLo's $R_{mi}$ indicator are reported. 
The overarching trend observed is that an increase in model size correlates with improved performance across these metrics (when fine-tuning on the RS5M dataset). This underscores the efficacy of scaling up the model for better task performance. However, an exception is noted in the case of SeLo's $R_{mi}$ indicator, which does not show significant variation. This suggests that scaling up the model size does not necessarily contribute to improvements in the semantic localization task, highlighting a potential task-specific ceiling in performance gains from increased model complexity.

\subsubsection{Influence of Batch size}

\begin{figure}
    \centering
    \includegraphics[width=0.47\textwidth]{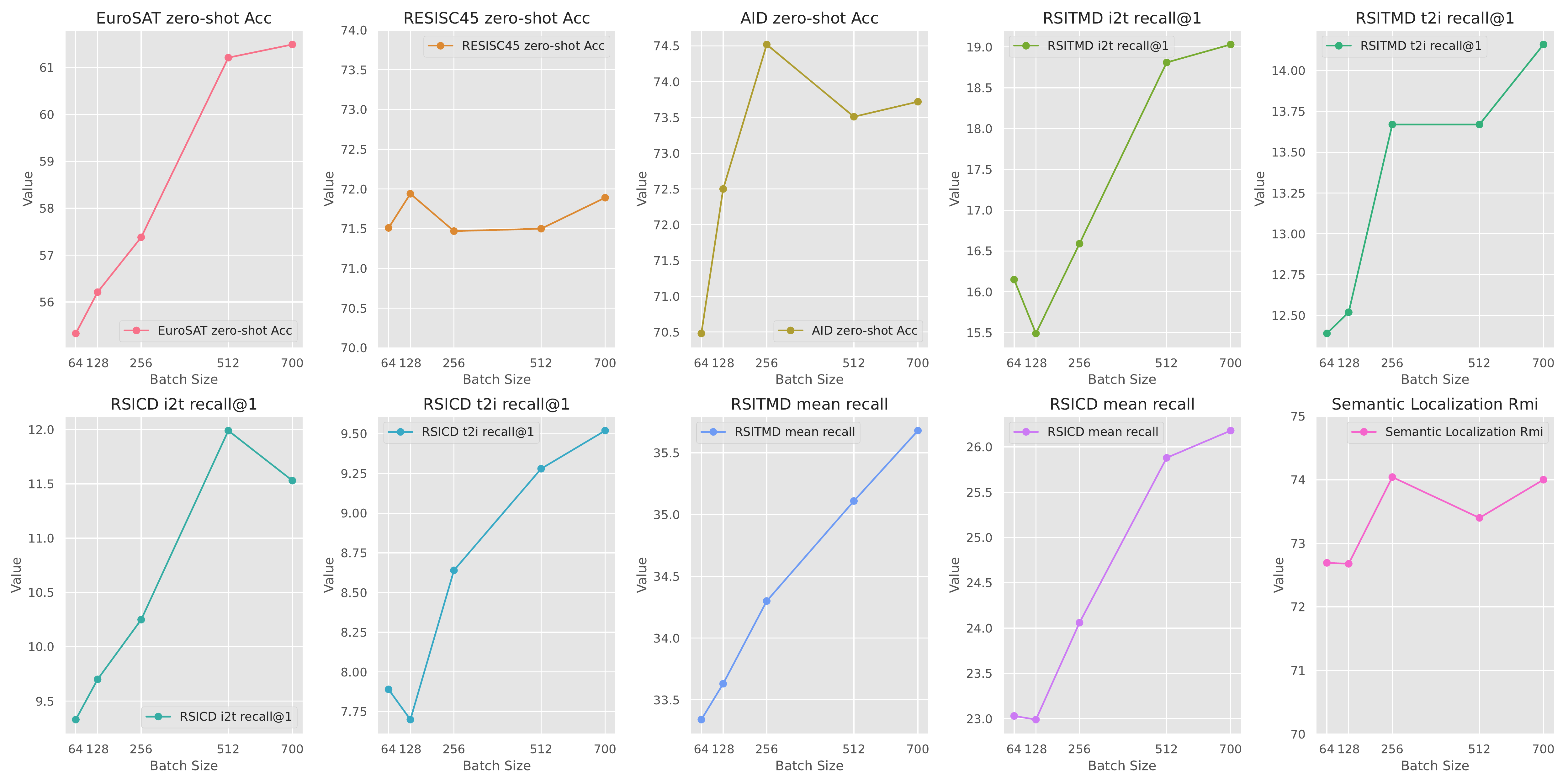}
    \caption{Influence of Batch Size}
    \label{fig:batch_size}
\end{figure}

We explored the effects of varying batch sizes on the performance of the ViT-B-32 model, fine-tuned with the RS5M dataset. The batch sizes selected for this study were 64, 128, 256, 512, and 700, providing a comprehensive range to assess the influence of batch size on the model.
The results from this experiment (shown in Figure \ref{fig:batch_size}) demonstrated a general trend: as the batch size increased, there was a corresponding improvement in the model's performance. This enhancement was consistently observed across the various metrics we used in previous subsections. The increase in performance with larger batch sizes could be attributed to the model's ability to learn more generalized features from a larger set of examples per training iteration. 




\section{Conclusion, Limitation, and Future Work}
We introduced a novel framework ($GVLM \xrightarrow{} DVLM \xrightarrow{} DTM$) and constructed the first large-scale RS image-text paired dataset, RS5M. We tried 4 PEFT approaches and fine-tuned the CLIP model with RS5M (GeoRSCLIP), and this framework has proven effective in tasks such as ZSC, RSCITR, and SeLo. However, either PEFT methods or full fine-tuning do not account for the interaction between the image and text modalities. This calls for the creation of more complex DVLMs in future work. Moreover, while VLM models were utilized to rank generated captions, we see potential in adopting more sophisticated selection criteria, such as decomposing captions into phrases and mapping them to image content, enabling a fine-grained alignment between an image and its caption. Another consideration pertains to our reliance on several CLIP models in our processing pipeline, which may propagate inherent biases within CLIP. Finally, we believe it is crucial to extend the exploration of advanced DVLMs' performance to other RS-related downstream tasks. Examples of these tasks include change detection, object detection, scene classification, semantic segmentation, RSVQA, and geo-localization for UAVs and satellite images. These explorations could offer valuable insights in the RS research domain.


\bibliographystyle{ieeetr}
\bibliography{ref}

\begin{thebibliography}{100}

\bibitem{rsen}
M.~Bauer, ``Remote sensing of environment: History, philosophy, approach and contributions, 1969 –2019,'' {\em Remote Sensing of Environment}, vol.~237, p.~111522, 02 2020.

\bibitem{rsup}
Y.~Xiao and Q.~Zhan, ``A review of remote sensing applications in urban planning and management in china,'' pp.~1 -- 5, 06 2009.

\bibitem{rsnda}
C.~Westen, ``Remote sensing for natural disaster management,'' 01 2000.

\bibitem{clip}
A.~Radford, J.~W. Kim, C.~Hallacy, A.~Ramesh, G.~Goh, S.~Agarwal, G.~Sastry, A.~Askell, P.~Mishkin, J.~Clark, G.~Krueger, and I.~Sutskever, ``Learning transferable visual models from natural language supervision,'' 2021.

\bibitem{align}
C.~Jia, Y.~Yang, Y.~Xia, Y.-T. Chen, Z.~Parekh, H.~Pham, Q.~V. Le, Y.~Sung, Z.~Li, and T.~Duerig, ``Scaling up visual and vision-language representation learning with noisy text supervision,'' 2021.

\bibitem{visualbert}
L.~H. Li, M.~Yatskar, D.~Yin, C.-J. Hsieh, and K.-W. Chang, ``Visualbert: A simple and performant baseline for vision and language,'' 2019.

\bibitem{vilt}
W.~Kim, B.~Son, and I.~Kim, ``Vilt: Vision-and-language transformer without convolution or region supervision,'' 2021.

\bibitem{uniter}
Y.-C. Chen, L.~Li, L.~Yu, A.~E. Kholy, F.~Ahmed, Z.~Gan, Y.~Cheng, and J.~Liu, ``Uniter: Universal image-text representation learning,'' 2020.

\bibitem{albef}
J.~Li, R.~R. Selvaraju, A.~D. Gotmare, S.~Joty, C.~Xiong, and S.~Hoi, ``Align before fuse: Vision and language representation learning with momentum distillation,'' 2021.

\bibitem{oscar}
X.~Li, X.~Yin, C.~Li, P.~Zhang, X.~Hu, L.~Zhang, L.~Wang, H.~Hu, L.~Dong, F.~Wei, Y.~Choi, and J.~Gao, ``Oscar: Object-semantics aligned pre-training for vision-language tasks,'' 2020.

\bibitem{coca}
J.~Yu, Z.~Wang, V.~Vasudevan, L.~Yeung, M.~Seyedhosseini, and Y.~Wu, ``Coca: Contrastive captioners are image-text foundation models,'' 2022.

\bibitem{flamingo}
J.-B. Alayrac, J.~Donahue, P.~Luc, A.~Miech, I.~Barr, Y.~Hasson, K.~Lenc, A.~Mensch, K.~Millican, M.~Reynolds, R.~Ring, E.~Rutherford, S.~Cabi, T.~Han, Z.~Gong, S.~Samangooei, M.~Monteiro, J.~Menick, S.~Borgeaud, A.~Brock, A.~Nematzadeh, S.~Sharifzadeh, M.~Binkowski, R.~Barreira, O.~Vinyals, A.~Zisserman, and K.~Simonyan, ``Flamingo: a visual language model for few-shot learning,'' 2022.

\bibitem{florence}
L.~Yuan, D.~Chen, Y.-L. Chen, N.~Codella, X.~Dai, J.~Gao, H.~Hu, X.~Huang, B.~Li, C.~Li, C.~Liu, M.~Liu, Z.~Liu, Y.~Lu, Y.~Shi, L.~Wang, J.~Wang, B.~Xiao, Z.~Xiao, J.~Yang, M.~Zeng, L.~Zhou, and P.~Zhang, ``Florence: A new foundation model for computer vision,'' 2021.

\bibitem{declip}
Y.~Li, F.~Liang, L.~Zhao, Y.~Cui, W.~Ouyang, J.~Shao, F.~Yu, and J.~Yan, ``Supervision exists everywhere: A data efficient contrastive language-image pre-training paradigm,'' 2022.

\bibitem{blip}
J.~Li, D.~Li, C.~Xiong, and S.~Hoi, ``Blip: Bootstrapping language-image pre-training for unified vision-language understanding and generation,'' 2022.

\bibitem{blip2}
J.~Li, D.~Li, S.~Savarese, and S.~Hoi, ``Blip-2: Bootstrapping language-image pre-training with frozen image encoders and large language models,'' 2023.

\bibitem{beit3}
W.~Wang, H.~Bao, L.~Dong, J.~Bjorck, Z.~Peng, Q.~Liu, K.~Aggarwal, O.~K. Mohammed, S.~Singhal, S.~Som, and F.~Wei, ``Image as a foreign language: Beit pretraining for all vision and vision-language tasks,'' 2022.

\bibitem{kosmos}
S.~Huang, L.~Dong, W.~Wang, Y.~Hao, S.~Singhal, S.~Ma, T.~Lv, L.~Cui, O.~K. Mohammed, B.~Patra, Q.~Liu, K.~Aggarwal, Z.~Chi, J.~Bjorck, V.~Chaudhary, S.~Som, X.~Song, and F.~Wei, ``Language is not all you need: Aligning perception with language models,'' 2023.

\bibitem{dalle}
A.~Ramesh, M.~Pavlov, G.~Goh, S.~Gray, C.~Voss, A.~Radford, M.~Chen, and I.~Sutskever, ``Zero-shot text-to-image generation,'' 2021.

\bibitem{stable-diffusion}
R.~Rombach, A.~Blattmann, D.~Lorenz, P.~Esser, and B.~Ommer, ``High-resolution image synthesis with latent diffusion models,'' in {\em Proceedings of the IEEE Conference on Computer Vision and Pattern Recognition (CVPR)}, 2022.

\bibitem{DAMMFM}
A.~O. Wiehe, ``Domain adaptation for multi-modal foundation models,'' 2022.

\bibitem{alfassy2022feta}
A.~Alfassy, A.~Arbelle, O.~Halimi, S.~Harary, R.~Herzig, E.~Schwartz, R.~Panda, M.~Dolfi, C.~Auer, K.~Saenko, P.~J. Staar, R.~Feris, and L.~Karlinsky, ``Feta: Towards specializing foundation models for expert task applications,'' 2022.

\bibitem{wortsman2022robust}
M.~Wortsman, G.~Ilharco, J.~W. Kim, M.~Li, S.~Kornblith, R.~Roelofs, R.~Gontijo-Lopes, H.~Hajishirzi, A.~Farhadi, H.~Namkoong, and L.~Schmidt, ``Robust fine-tuning of zero-shot models,'' 2022.

\bibitem{synthetic}
R.~He, S.~Sun, X.~Yu, C.~Xue, W.~Zhang, P.~Torr, S.~Bai, and X.~Qi, ``Is synthetic data from generative models ready for image recognition?,'' 2023.

\bibitem{UCMSydeney}
B.~Qu, X.~Li, D.~Tao, and X.~Lu, ``Deep semantic understanding of high resolution remote sensing image,'' in {\em 2016 International Conference on Computer, Information and Telecommunication Systems (CITS)}, pp.~1--5, 2016.

\bibitem{RSICD}
X.~Lu, B.~Wang, X.~Zheng, and X.~Li, ``Exploring models and data for remote sensing image caption generation,'' {\em IEEE Transactions on Geoscience and Remote Sensing}, vol.~56, no.~4, pp.~2183--2195, 2017.

\bibitem{RSITMD}
Z.~Yuan, W.~Zhang, K.~Fu, X.~Li, C.~Deng, H.~Wang, and X.~Sun, ``Exploring a fine-grained multiscale method for cross-modal remote sensing image retrieval,'' {\em IEEE Transactions on Geoscience and Remote Sensing}, vol.~60, pp.~1--19, 2022.

\bibitem{rsvg}
Y.~Zhan, Z.~Xiong, and Y.~Yuan, ``Rsvg: Exploring data and models for visual grounding on remote sensing data,'' 2022.

\bibitem{millionaid}
Y.~Long, G.-S. Xia, S.~Li, W.~Yang, M.~Y. Yang, X.~X. Zhu, L.~Zhang, and D.~Li, ``On creating benchmark dataset for aerial image interpretation: Reviews, guidances and million-aid,'' {\em IEEE Journal of Selected Topics in Applied Earth Observations and Remote Sensing}, vol.~14, pp.~4205--4230, 2021.

\bibitem{ben}
G.~Sumbul, M.~Charfuelan, B.~Demir, and V.~Markl, ``Bigearthnet: A large-scale benchmark archive for remote sensing image understanding,'' in {\em {IGARSS} 2019 - 2019 {IEEE} International Geoscience and Remote Sensing Symposium}, IEEE, jul 2019.

\bibitem{fmow}
G.~Christie, N.~Fendley, J.~Wilson, and R.~Mukherjee, ``Functional map of the world,'' 2018.

\bibitem{bigearthnet}
G.~Sumbul, M.~Charfuelan, B.~Demir, and V.~Markl, ``Bigearthnet: A large-scale benchmark archive for remote sensing image understanding,'' {\em CoRR}, vol.~abs/1902.06148, 2019.

\bibitem{du2022survey}
Y.~Du, Z.~Liu, J.~Li, and W.~X. Zhao, ``A survey of vision-language pre-trained models,'' 2022.

\bibitem{filip}
L.~Yao, R.~Huang, L.~Hou, G.~Lu, M.~Niu, H.~Xu, X.~Liang, Z.~Li, X.~Jiang, and C.~Xu, ``Filip: Fine-grained interactive language-image pre-training,'' 2021.

\bibitem{flip}
Y.~Li, H.~Fan, R.~Hu, C.~Feichtenhofer, and K.~He, ``Scaling language-image pre-training via masking,'' 2022.

\bibitem{mae}
K.~He, X.~Chen, S.~Xie, Y.~Li, P.~Dollár, and R.~Girshick, ``Masked autoencoders are scalable vision learners,'' 2021.

\bibitem{airs}
L.~Zhang and L.~Zhang, ``Artificial intelligence for remote sensing data analysis: A review of challenges and opportunities,'' {\em IEEE Geoscience and Remote Sensing Magazine}, vol.~10, no.~2, pp.~270--294, 2022.

\bibitem{wen2023visionlanguage}
C.~Wen, Y.~Hu, X.~Li, Z.~Yuan, and X.~X. Zhu, ``Vision-language models in remote sensing: Current progress and future trends,'' 2023.

\bibitem{rsvqa}
S.~Lobry, D.~Marcos, J.~Murray, and D.~Tuia, ``{RSVQA}: Visual question answering for remote sensing data,'' {\em {IEEE} Transactions on Geoscience and Remote Sensing}, vol.~58, pp.~8555--8566, dec 2020.

\bibitem{hu2023rsgpt}
Y.~Hu, J.~Yuan, C.~Wen, X.~Lu, and X.~Li, ``Rsgpt: A remote sensing vision language model and benchmark,'' 2023.

\bibitem{selo}
Z.~Yuan, W.~Zhang, C.~Li, Z.~Pan, Y.~Mao, J.~Chen, S.~Li, H.~Wang, and X.~Sun, ``Learning to evaluate performance of multimodal semantic localization,'' {\em {IEEE} Transactions on Geoscience and Remote Sensing}, vol.~60, pp.~1--18, 2022.

\bibitem{galr}
Z.~Yuan, W.~Zhang, C.~Tian, X.~Rong, Z.~Zhang, H.~Wang, K.~Fu, and X.~Sun, ``Remote sensing cross-modal text-image retrieval based on global and local information,'' {\em IEEE Transactions on Geoscience and Remote Sensing}, vol.~60, pp.~1--16, 2022.

\bibitem{clip-rs}
L.~D. Basso, ``Clip-rs: A cross-modal remote sensing image retrieval based on clip, a northern virginia case study,'' 2022.

\bibitem{rsp}
D.~Wang, J.~Zhang, B.~Du, G.-S. Xia, and D.~Tao, ``An empirical study of remote sensing pretraining,'' 2022.

\bibitem{bert}
J.~Devlin, M.-W. Chang, K.~Lee, and K.~Toutanova, ``Bert: Pre-training of deep bidirectional transformers for language understanding,'' 2019.

\bibitem{gpt}
A.~Radford, K.~Narasimhan, T.~Salimans, and I.~Sutskever, ``Improving language understanding by generative pre-training,'' 2018.

\bibitem{houlsby2019parameterefficient}
N.~Houlsby, A.~Giurgiu, S.~Jastrzebski, B.~Morrone, Q.~de~Laroussilhe, A.~Gesmundo, M.~Attariyan, and S.~Gelly, ``Parameter-efficient transfer learning for nlp,'' 2019.

\bibitem{pfeiffer2021adapterfusion}
J.~Pfeiffer, A.~Kamath, A.~Rücklé, K.~Cho, and I.~Gurevych, ``Adapterfusion: Non-destructive task composition for transfer learning,'' 2021.

\bibitem{li2021prefixtuning}
X.~L. Li and P.~Liang, ``Prefix-tuning: Optimizing continuous prompts for generation,'' 2021.

\bibitem{hu2021lora}
E.~J. Hu, Y.~Shen, P.~Wallis, Z.~Allen-Zhu, Y.~Li, S.~Wang, L.~Wang, and W.~Chen, ``Lora: Low-rank adaptation of large language models,'' 2021.

\bibitem{unipelt}
Y.~Mao, L.~Mathias, R.~Hou, A.~Almahairi, H.~Ma, J.~Han, W.~tau Yih, and M.~Khabsa, ``Unipelt: A unified framework for parameter-efficient language model tuning,'' 2022.

\bibitem{clipadapter}
P.~Gao, S.~Geng, R.~Zhang, T.~Ma, R.~Fang, Y.~Zhang, H.~Li, and Y.~Qiao, ``Clip-adapter: Better vision-language models with feature adapters,'' 2021.

\bibitem{tipadapter}
R.~Zhang, R.~Fang, W.~Zhang, P.~Gao, K.~Li, J.~Dai, Y.~Qiao, and H.~Li, ``Tip-adapter: Training-free clip-adapter for better vision-language modeling,'' 2021.

\bibitem{vladapter}
Y.-L. Sung, J.~Cho, and M.~Bansal, ``Vl-adapter: Parameter-efficient transfer learning for vision-and-language tasks,'' 2022.

\bibitem{coop}
K.~Zhou, J.~Yang, C.~C. Loy, and Z.~Liu, ``Learning to prompt for vision-language models,'' {\em International Journal of Computer Vision}, vol.~130, pp.~2337--2348, jul 2022.

\bibitem{cocoop}
K.~Zhou, J.~Yang, C.~C. Loy, and Z.~Liu, ``Conditional prompt learning for vision-language models,'' 2022.

\bibitem{laion5b}
C.~Schuhmann, R.~Beaumont, R.~Vencu, C.~Gordon, R.~Wightman, M.~Cherti, T.~Coombes, A.~Katta, C.~Mullis, M.~Wortsman, P.~Schramowski, S.~Kundurthy, K.~Crowson, L.~Schmidt, R.~Kaczmarczyk, and J.~Jitsev, ``Laion-5b: An open large-scale dataset for training next generation image-text models,'' 2022.

\bibitem{laion400m}
C.~Schuhmann, R.~Vencu, R.~Beaumont, R.~Kaczmarczyk, C.~Mullis, A.~Katta, T.~Coombes, J.~Jitsev, and A.~Komatsuzaki, ``Laion-400m: Open dataset of clip-filtered 400 million image-text pairs,'' 2021.

\bibitem{coyo700m}
M.~Byeon, B.~Park, H.~Kim, S.~Lee, W.~Baek, and S.~Kim, ``Coyo-700m: Image-text pair dataset.'' \url{https://github.com/kakaobrain/coyo-dataset}, 2022.

\bibitem{cc3m}
P.~Sharma, N.~Ding, S.~Goodman, and R.~Soricut, ``Conceptual captions: A cleaned, hypernymed, image alt-text dataset for automatic image captioning,'' in {\em Proceedings of ACL}, 2018.

\bibitem{cc12m}
S.~Changpinyo, P.~Sharma, N.~Ding, and R.~Soricut, ``{Conceptual 12M}: Pushing web-scale image-text pre-training to recognize long-tail visual concepts,'' in {\em CVPR}, 2021.

\bibitem{yfcc100m}
B.~Thomee, D.~A. Shamma, G.~Friedland, B.~Elizalde, K.~Ni, D.~Poland, D.~Borth, and L.-J. Li, ``{YFCC}100m,'' {\em Communications of the {ACM}}, vol.~59, pp.~64--73, jan 2016.

\bibitem{wit}
K.~Srinivasan, K.~Raman, J.~Chen, M.~Bendersky, and M.~Najork, ``Wit: Wikipedia-based image text dataset for multimodal multilingual machine learning,'' {\em arXiv preprint arXiv:2103.01913}, 2021.

\bibitem{redcaps}
K.~Desai, G.~Kaul, Z.~Aysola, and J.~Johnson, ``{RedCaps: Web-curated image-text data created by the people, for the people},'' in {\em NeurIPS Datasets and Benchmarks}, 2021.

\bibitem{sbu}
V.~Ordonez, G.~Kulkarni, and T.~L. Berg, ``Im2text: Describing images using 1 million captioned photographs,'' in {\em Neural Information Processing Systems ({NIPS})}, 2011.

\bibitem{vg}
R.~Krishna, Y.~Zhu, O.~Groth, J.~Johnson, K.~Hata, J.~Kravitz, S.~Chen, Y.~Kalantidis, L.-J. Li, D.~A. Shamma, M.~Bernstein, and L.~Fei-Fei, ``Visual genome: Connecting language and vision using crowdsourced dense image annotations,'' 2016.

\bibitem{openclip}
G.~Ilharco, M.~Wortsman, R.~Wightman, C.~Gordon, N.~Carlini, R.~Taori, A.~Dave, V.~Shankar, H.~Namkoong, J.~Miller, H.~Hajishirzi, A.~Farhadi, and L.~Schmidt, ``Openclip,'' July 2021.
\newblock If you use this software, please cite it as below.

\bibitem{wang2022advancing}
D.~Wang, Q.~Zhang, Y.~Xu, J.~Zhang, B.~Du, D.~Tao, and L.~Zhang, ``Advancing plain vision transformer towards remote sensing foundation model,'' 2022.

\bibitem{aid}
G.-S. Xia, J.~Hu, F.~Hu, B.~Shi, X.~Bai, Y.~Zhong, L.~Zhang, and X.~Lu, ``{AID}: A benchmark data set for performance evaluation of aerial scene classification,'' {\em {IEEE} Transactions on Geoscience and Remote Sensing}, vol.~55, pp.~3965--3981, jul 2017.

\bibitem{RESISC45}
G.~Cheng, J.~Han, and X.~Lu, ``Remote sensing image scene classification: Benchmark and state of the art,'' {\em Proceedings of the {IEEE}}, vol.~105, pp.~1865--1883, oct 2017.

\bibitem{eurosat}
P.~Helber, B.~Bischke, A.~Dengel, and D.~Borth, ``Eurosat: A novel dataset and deep learning benchmark for land use and land cover classification,'' 08 2017.

\bibitem{adamw}
I.~Loshchilov and F.~Hutter, ``Decoupled weight decay regularization,'' 2019.

\bibitem{infonce}
A.~van~den Oord, Y.~Li, and O.~Vinyals, ``Representation learning with contrastive predictive coding,'' 2019.

\bibitem{selov2}
M.~Yu, H.~Yuan, J.~Chen, C.~Hao, Z.~Wang, Z.~Yuan, and B.~Lu, ``Selo v2: Toward for higher and faster semantic localization,'' {\em IEEE Geoscience and Remote Sensing Letters}, vol.~20, pp.~1--5, 2023.

\bibitem{vse++}
F.~Faghri, D.~J. Fleet, J.~R. Kiros, and S.~Fidler, ``Vse++: Improving visual-semantic embeddings with hard negatives,'' 2018.

\bibitem{kcr}
L.~Mi, S.~Li, C.~Chappuis, and D.~Tuia, ``Knowledge-aware cross-modal text-image retrieval for remote sensing images,'' 2022.

\bibitem{LW-MCR}
Z.~Yuan, W.~Zhang, X.~Rong, X.~Li, J.~Chen, H.~Wang, K.~Fu, and X.~Sun, ``A lightweight multi-scale crossmodal text-image retrieval method in remote sensing,'' {\em IEEE Transactions on Geoscience and Remote Sensing}, vol.~60, pp.~1--19, 2022.

\bibitem{hvsa}
W.~Zhang, J.~Li, S.~Li, J.~Chen, W.~Zhang, X.~Gao, and X.~Sun, ``Hypersphere-based remote sensing cross-modal text–image retrieval via curriculum learning,'' {\em IEEE Transactions on Geoscience and Remote Sensing}, vol.~61, pp.~1--15, 2023.

\bibitem{Zheng_2023}
F.~Zheng, X.~Wang, L.~Wang, X.~Zhang, H.~Zhu, L.~Wang, and H.~Zhang, ``A fine-grained semantic alignment method specific to aggregate multi-scale information for cross-modal remote sensing image retrieval,'' {\em Sensors}, vol.~23, p.~8437, Oct. 2023.

\bibitem{pir}
J.~Pan, Q.~Ma, and C.~Bai, ``A prior instruction representation framework for remote sensing image-text retrieval,'' pp.~611--620, 10 2023.

\bibitem{liu2023remoteclip}
F.~Liu, D.~Chen, Z.~Guan, X.~Zhou, J.~Zhu, and J.~Zhou, ``Remoteclip: A vision language foundation model for remote sensing,'' 2023.

\bibitem{multilingual}
M.~M.~A. Rahhal, Y.~Bazi, N.~A. Alsharif, L.~Bashmal, N.~Alajlan, and F.~Melgani, ``Multilanguage transformer for improved text to remote sensing image retrieval,'' {\em IEEE Journal of Selected Topics in Applied Earth Observations and Remote Sensing}, vol.~15, pp.~9115--9126, 2022.

\bibitem{yuan2023parameterefficient}
Y.~Yuan, Y.~Zhan, and Z.~Xiong, ``Parameter-efficient transfer learning for remote sensing image-text retrieval,'' 2023.

\bibitem{MTGFE}
X.~Zhang, W.~Li, X.~Wang, L.~Wang, F.~Zheng, L.~Wang, and H.~Zhang, ``A fusion encoder with multi-task guidance for cross-modal text–image retrieval in remote sensing,'' {\em Remote Sensing}, vol.~15, p.~4637, Sept. 2023.

\bibitem{ucm}
Y.~Yang and S.~Newsam, ``Bag-of-visual-words and spatial extensions for land-use classification,'' pp.~270--279, 11 2010.

\bibitem{6910306}
F.~Zhang, B.~Du, and L.~Zhang, ``Saliency-guided unsupervised feature learning for scene classification,'' {\em IEEE Transactions on Geoscience and Remote Sensing}, vol.~53, no.~4, pp.~2175--2184, 2015.

\bibitem{slip}
N.~Mu, A.~Kirillov, D.~Wagner, and S.~Xie, ``Slip: Self-supervision meets language-image pre-training,'' 2021.

\bibitem{soho}
Z.~Huang, Z.~Zeng, Y.~Huang, B.~Liu, D.~Fu, and J.~Fu, ``Seeing out of the box: End-to-end pre-training for vision-language representation learning,'' 2021.

\bibitem{vilbert}
J.~Lu, D.~Batra, D.~Parikh, and S.~Lee, ``Vilbert: Pretraining task-agnostic visiolinguistic representations for vision-and-language tasks,'' 2019.

\bibitem{vlbert}
W.~Su, X.~Zhu, Y.~Cao, B.~Li, L.~Lu, F.~Wei, and J.~Dai, ``Vl-bert: Pre-training of generic visual-linguistic representations,'' 2020.

\bibitem{vlmo}
H.~Bao, W.~Wang, L.~Dong, Q.~Liu, O.~K. Mohammed, K.~Aggarwal, S.~Som, and F.~Wei, ``Vlmo: Unified vision-language pre-training with mixture-of-modality-experts,'' 2022.

\bibitem{glip}
L.~H. Li*, P.~Zhang*, H.~Zhang*, J.~Yang, C.~Li, Y.~Zhong, L.~Wang, L.~Yuan, L.~Zhang, J.-N. Hwang, K.-W. Chang, and J.~Gao, ``Grounded language-image pre-training,'' in {\em CVPR}, 2022.

\bibitem{mdetr}
A.~Kamath, M.~Singh, Y.~LeCun, G.~Synnaeve, I.~Misra, and N.~Carion, ``Mdetr -- modulated detection for end-to-end multi-modal understanding,'' 2021.

\bibitem{xdetr}
Z.~Cai, G.~Kwon, A.~Ravichandran, E.~Bas, Z.~Tu, R.~Bhotika, and S.~Soatto, ``X-detr: A versatile architecture for instance-wise vision-language tasks,'' 2022.

\bibitem{regionclip}
Y.~Zhong, J.~Yang, P.~Zhang, C.~Li, N.~Codella, L.~H. Li, L.~Zhou, X.~Dai, L.~Yuan, Y.~Li, {\em et~al.}, ``Regionclip: Region-based language-image pretraining,'' in {\em Proceedings of the IEEE/CVF Conference on Computer Vision and Pattern Recognition}, pp.~16793--16803, 2022.

\bibitem{groupvit}
J.~Xu, S.~D. Mello, S.~Liu, W.~Byeon, T.~Breuel, J.~Kautz, and X.~Wang, ``Groupvit: Semantic segmentation emerges from text supervision,'' 2022.

\bibitem{dalle2}
A.~R. et~al, ``Hierarchical text-conditional image generation with clip latents,'' 2022.

\bibitem{imagine}
P.~Wang, Y.~Li, K.~K. Singh, J.~Lu, and N.~Vasconcelos, ``Imagine: Image synthesis by image-guided model inversion,'' 2021.

\bibitem{faiss}
J.~Johnson, M.~Douze, and H.~J{\'e}gou, ``Billion-scale similarity search with {GPUs},'' {\em IEEE Transactions on Big Data}, vol.~7, no.~3, pp.~535--547, 2019.

\bibitem{compacter}
R.~K. Mahabadi, J.~Henderson, and S.~Ruder, ``Compacter: Efficient low-rank hypercomplex adapter layers,'' 2021.

\bibitem{hyperformer}
R.~K. Mahabadi, S.~Ruder, M.~Dehghani, and J.~Henderson, ``Parameter-efficient multi-task fine-tuning for transformers via shared hypernetworks,'' 2021.

\bibitem{colorfulprompt}
Y.~Yao, A.~Zhang, Z.~Zhang, Z.~Liu, T.-S. Chua, and M.~Sun, ``Cpt: Colorful prompt tuning for pre-trained vision-language models,'' 2022.

\bibitem{dota}
G.-S. Xia, X.~Bai, J.~Ding, Z.~Zhu, S.~Belongie, J.~Luo, M.~Datcu, M.~Pelillo, and L.~Zhang, ``Dota: A large-scale dataset for object detection in aerial images,'' 2019.

\bibitem{mañas2021seasonal}
O.~Mañas, A.~Lacoste, X.~G. i~Nieto, D.~Vazquez, and P.~Rodriguez, ``Seasonal contrast: Unsupervised pre-training from uncurated remote sensing data,'' 2021.

\bibitem{ayush2022geographyaware}
K.~Ayush, B.~Uzkent, C.~Meng, K.~Tanmay, M.~Burke, D.~Lobell, and S.~Ermon, ``Geography-aware self-supervised learning,'' 2022.

\bibitem{mocov2}
X.~Chen, H.~Fan, R.~Girshick, and K.~He, ``Improved baselines with momentum contrastive learning,'' 2020.

\bibitem{vincenzi2020color}
S.~Vincenzi, A.~Porrello, P.~Buzzega, M.~Cipriano, P.~Fronte, R.~Cuccu, C.~Ippoliti, A.~Conte, and S.~Calderara, ``The color out of space: learning self-supervised representations for earth observation imagery,'' 2020.

\bibitem{mai2021geographic}
G.~Mai, K.~Janowicz, R.~Zhu, L.~Cai, and N.~Lao, ``Geographic question answering: Challenges, uniqueness, classification, and future directions,'' 2021.

\bibitem{chen2022geoqa}
J.~Chen, J.~Tang, J.~Qin, X.~Liang, L.~Liu, E.~P. Xing, and L.~Lin, ``Geoqa: A geometric question answering benchmark towards multimodal numerical reasoning,'' 2022.

\bibitem{huang2019geosqa}
Z.~Huang, Y.~Shen, X.~Li, Y.~Wei, G.~Cheng, L.~Zhou, X.~Dai, and Y.~Qu, ``Geosqa: A benchmark for scenario-based question answering in the geography domain at high school level,'' 2019.

\bibitem{Contractor_2021}
D.~Contractor, S.~Goel, Mausam, and P.~Singla, ``Joint spatio-textual reasoning for answering tourism questions,'' in {\em Proceedings of the Web Conference 2021}, {ACM}, apr 2021.

\bibitem{punjani2021templatebased}
D.~Punjani, M.~Iliakis, T.~Stefou, K.~Singh, A.~Both, M.~Koubarakis, I.~Angelidis, K.~Bereta, T.~Beris, D.~Bilidas, T.~Ioannidis, N.~Karalis, C.~Lange, D.-A. Pantazi, C.~Papaloukas, and G.~Stamoulis, ``Template-based question answering over linked geospatial data,'' 2021.

\bibitem{dreambooth}
N.~Ruiz, Y.~Li, V.~Jampani, Y.~Pritch, M.~Rubinstein, and K.~Aberman, ``Dreambooth: Fine tuning text-to-image diffusion models for subject-driven generation,'' 2022.

\bibitem{diffusers}
P.~von Platen, S.~Patil, A.~Lozhkov, P.~Cuenca, N.~Lambert, K.~Rasul, M.~Davaadorj, and T.~Wolf, ``Diffusers: State-of-the-art diffusion models.'' \url{https://github.com/huggingface/diffusers}, 2022.

\end{thebibliography}


\appendix
\section{Appendix}
\subsection{Related Work}
\label{appendix:related_work}
\subsubsection{Image-Text Paired Datasets for Remote Sensing}

\textbf{UCM Captions} dataset \cite{UCMSydeney}, derived from the UC Merced Land Use Dataset \cite{ucm} by Qu et al. The image data is extracted from the USGS National Map Urban Area Imagery collection and consists of 2,100 RGB aerial images from 21 classes. Each image includes 5 captions, with 2032 unique captions in total. The image resolution is 256 $\times$ 256, and the spatial resolution is 1ft.

\textbf{Sydney Captions} dataset \cite{UCMSydeney}, a version of the Sydney scene classification dataset proposed in \cite{6910306}, contains 613 RGB images of Sydney, Australia, acquired using Google Earth. Qu et al. provided 3,065 captions, 1109 of which are non-duplicate captions. The image size is 500 $\times$ 500, with 1 ft spatial resolution.

\textbf{RSICD} \cite{RSICD} is a dataset contributed by Lu et al., containing 10,921 remote sensing RGB images from Google Earth, Baidu Map, MapABC and Tianditu. Each image is annotated with 5 natural language captions, with 18,190 unique ones. The image resolution is 224 $\times$ 224 pixels. This dataset, along with UCM Captions and Sydney Captions dataset, contains very repetitive language with little detail.

\textbf{RSITMD} \cite{RSITMD} (Remote Sensing Image-Text Match dataset) is a fine-grained and challenging RS dataset for image-text matching, proposed by Yuan et al. It is originally designed for RS multimodal retrieval tasks and features detailed captions describing object relations compared to other RS image-text paired datasets. Additionally, it contains keyword attributes (1–5 keywords for each image) that can be utilized for RS text retrieval tasks based on keywords. The dataset has a total of 23,715 captions for 4,743 images across 32 scenes, with 21,829 of these being non-duplicate.

\textbf{RSVGD} \cite{rsvg} is a comprehensive benchmark dataset for Remote Sensing Visual Grounding (RSVG) tasks, introduced by Zhan et al. in 2022. The RSVG task focuses on localizing objects of interest referenced in queries within RS images. The dataset is built upon the DIOR RS image dataset, originally designed for object detection. RSVGD comprises 38,320 RS image-text pairs and 17,402 RS images, with an average expression length of 7.47 and a vocabulary size of 100. The image resolution is 800 × 800 pixels, and the spatial resolution ranges from 0.5m to 30m. The text description is synthesized from templates and pre-defined rules.

\subsubsection{Large-Scale Image Dataset for Remote Sensing} \label{large_scale_rs_dataset}

\textbf{BigEarthNet} \cite{bigearthnet} is a large-scale RS dataset comprising 590,326 pairs of Sentinel-1 and Sentinel-2 image patches. The BigEarthNet archive project is supported by the European Research Council. Each image is accompanied by multi-class labels. The data was collected from June 2017 to May 2018 across 10 European countries and has been atmospherically corrected. BigEarthNet with Sentinel-1 image patches has 2 channels (VV and VH), while BigEarthNet with Sentinel-2 image patches includes 12 channels \footnote{https://www.tensorflow.org/datasets/catalog/bigearthnet}.

\textbf{Functional Map of the World}, a.k.a. FMoW \cite{fmow}, is a RS dataset consists of 1,047,691 images covering 207 countries, collected by Christie1 et al. in 2018 from the DigitalGlobe constellation \footnote{https://www.digitalglobe.com/resources/satellite-information}. They provide extra meta information such as location, time, sun angles, physical sizes, etc. For each image, at least one bounding box annotation for 1 of 63 categories is offered. There are two versions of the dataset, fMoW-full includes 4-band and 8-band multi-spectral information and is in tif format, and fMoW-rgb is in JPEG format with RGB channels only.

\textbf{Million-AID} \cite{millionaid} is another large-scale RS benchmark dataset containing 1 million RGB images for remote sensing image scene classification tasks. Proposed by Long et al., the dataset extracts aerial images from Google Earth and features a three-level class taxonomy tree with 51 third-level (leaf) nodes, 28 second-level nodes, and 8 first-level nodes. The authors also devised several strategies for manual, automatic, and interactive annotation of RS images.

\subsubsection{Vision-Language Model Overview and Application} 

Large-scale pre-trained VLMs can be categorized based on their pre-training task objectives, such as contrastive vision-text alignment, image-text matching, masked language modeling, etc. \cite{du2022survey}.

CLIP \cite{clip}, which uses 400 million image-text pairs, demonstrates remarkable generalizability even when faced with distribution shifts. ALIGN \cite{align} further illustrates that increasing dataset size, even with noisy data, can lead to performance improvements. Variants of CLIP either mine fine-grained alignment between image and text tokens \cite{filip} or aim to learn better representations through self-supervision \cite{slip} and cross-modality supervision \cite{declip}. These models align textual and visual information in a shared semantic space using contrastive learning tasks, and their success is closely linked to the vast amount of data. UNITER \cite{uniter}, SOHO \cite{soho}, ViLBert \cite{vilbert}, ALBEF \cite{albef}, and BLIP \cite{blip} employ image-text matching task objectives, allowing them to learn fine-grained alignment between image and text representations. Models such as Oscar \cite{oscar}, VL-bert \cite{vlbert}, VisualBert \cite{visualbert}, FLIP \cite{flip}, and BEIT3 \cite{beit3} utilize Masked Language Modeling objectives, a strategy proven to be not only effective but also efficient \cite{flip}. Predictions for masked tokens in these models are based on both unmasked visual and language tokens, leveraging and aligning tokens from both modalities.

Various innovative approaches have been introduced to enhance the performance of pre-trained VLMs. These include in-context learning by Flamingo \cite{flamingo}, captioning loss by CoCa \cite{coca}, Language-Image Bootstrapping by BLIP \cite{blip}, and the Mixture of Experts Framework from VLMo \cite{vlmo}. It is important to note that most pre-trained VLMs combine multiple pre-training task objectives. For instance, ALBEF \cite{albef} employs contrastive loss and image-text matching loss, CoCa \cite{coca} utilizes contrastive loss and captioning loss, and FLIP \cite{flip} uses contrastive loss and loss from MAE \cite{mae}.

Pre-trained VLMs have demonstrated their ability to tackle not only general tasks like image-text retrieval, zero-shot classification, image captioning, and VQA, but also more complex vision tasks. Examples include GLIP \cite{glip} for visual grounding, MDETR \cite{mdetr}, XDETR \cite{xdetr}, and RegionClip \cite{regionclip} for cross-modal object detection, and GroupViT \cite{groupvit} for text-supervised image segmentation. Additionally, generative VLMs such as DALLE \cite{dalle}, DALLE2 \cite{dalle2}, IMAGINE \cite{imagine}, and stable-diffusion \cite{stable-diffusion} have gained significant attention in recent years.

\subsubsection{Vision-Language Model for Remote Sensing}

Wen et al. \cite{wen2023visionlanguage} presented a survey on Vision-Language Models in Remote Sensing, and concluded several promising research directions. These include making Large-scale datasets and Vision-language Vision-Language models, creating text-based image generation using diffusion models, doing few-/zero-shot learning with LLMs and VLMs, efficient finetuning on RS data, integrating RS expert knowledge into LLMs, and linking text-based information with RS via geolocation. Zhang et al.\cite{airs} provided a comprehensive overview of recent advancements in applying artificial intelligence techniques to remote sensing data analysis. It covers major AI aspects including machine learning, deep learning, computational intelligence, AI explicability, data mining, natural language processing, and AI security. Key topics include CNNs for tasks like classification, detection, and fusion; generative models like GANs; evolutionary algorithms and neural architecture search for optimization; efforts towards interpretable models; mining multimodal data; generating image descriptions; and adversarial threats. Hu et al. \cite{hu2023rsgpt} made a new high-quality remote sensing image captioning dataset called RSICap, consisting of 2,585 human-annotated image captions with rich scene and object details. The authors also introduce an evaluation benchmark called RSIEval with image captions and visual question-answering pairs. Based on these datasets, the authors develop a remote sensing vision-language model called RSGPT by finetuning InstructBLIP on RSICap.  

Yuan et al. introduced AMFMN \cite{RSITMD}, an asymmetric multimodal feature matching network using triplet loss with a dynamic variable margin, designed for cross-modal RS text-image retrieval tasks. They later developed LW-MCR \cite{LW-MCR} and GALR \cite{galr}, lightweight cross-modal text-image retrieval methods that outperform AMFMN in speed and performance. Additionally, they proposed Semantic Localization tasks \cite{selo}, a weak visual grounding task enabling semantic-level retrieval with caption-level annotation. Zhan et al. presented MLCM, a transformer-based multi-Level cross-modal feature learning module that adaptively filters irrelevant noise and enhances salient features for the RSVG task \cite{rsvg}. Basso introduced CLIP-RS \cite{clip-rs}, a cross-modal remote sensing image retrieval platform that combines CLIP with the FAISS library \cite{faiss} (a library for similarity search), using RS data from Northern Virginia. Arutiunian et al. fine-tuned CLIP with RSICD, achieving significant improvements in top-1 accuracy for zero-shot classification \footnote{https://huggingface.co/blog/fine-tune-clip-rsicd}.

\subsubsection{Parameter-Efficient Tuning for Large Language Models}

Large Language Models (LLMs) like BERT \cite{bert} and GPT \cite{gpt}, trained on vast text corpora, have achieved state-of-the-art results across numerous NLP benchmarks. However, their millions or billions of parameters make full fine-tuning for each downstream task unrealistic. Adapters offer an alternative solution for LLM fine-tuning, as they freeze the pre-trained LLM's weights while training only the adapter's parameters, which have significantly less number of parameters. This approach speeds up adaptation while maintaining comparable performance to full fine-tuning. 

The adapter was originally proposed by Houlsby et al. \cite{houlsby2019parameterefficient}, adding two MLPs with bottleneck structures and residual connections after the feed-forward layers in every transformer layer. Pfeiffer et al. introduced AdapterFusion \cite{pfeiffer2021adapterfusion}, which designs different adapters for various tasks and learns a parameterized mixer to encode information in the end. Inspired by prompt learning, Li et al. proposed Prefix-Tuning \cite{li2021prefixtuning}, which adds a small, continuous, task-specific vector (a.k.a, prefix) before the text for adaptation. Hu et al. developed the well-known LoRA \cite{hu2021lora}, Low-Rank Adaptation, which injects trainable rank decomposition matrices ($W_0 x+\Delta W x=W_0 x+B A x$) to approximate adaptation in each transformer layer. Mao et al. proposed UNIPELT \cite{unipelt}, a unified framework for parameter-efficient language model tuning (PELT), which combines different PELT methods as submodules (e.g., bottleneck adapter \cite{houlsby2019parameterefficient}, LoRA \cite{hu2021lora}, prefix-tuning\cite{li2021prefixtuning}) and learns to activate the best-suited ones for the current task through a gating mechanism.

\subsubsection{Parameter-Efficient Tuning for Vision-Language Models}
Gao et al. proposed CLIP-Adapter \cite{clipadapter}, which adds a two-layer MLP with a bottleneck structure and residual connection to the text and vision encoder for visual classification tasks. Zhang et al. introduced Tip-Adapter \cite{tipadapter}, a training-free adapter for CLIP that constructs a key-value cache model from few-shot examples in the training set, allowing cached examples to vote for their labels. In VL-Adapter \cite{vladapter}, Sung et al. applied adapter-based parameter-efficient transfer tuning methods (Bottleneck adapter \cite{houlsby2019parameterefficient}, compacter \cite{compacter}, hyperformer \cite{hyperformer}) to VLMs, addressing VQA and image captioning tasks.

Another approach to parameter-efficient tuning for VLMs is prompt-based learning. CoOp \cite{coop} learns prompt tokens for input in the text encoder to assist zero-shot classification. CoCoop \cite{cocoop} extends CoOp by using a lightweight neural network to generate an input-conditional token for each image. Colorful Prompt \cite{colorfulprompt} presents cross-modal prompt tuning, constructing fill-in-the-blank problems in a color-based co-referential manner.

\subsubsection{Pre-trained Models in Remote Sensing}

Wang et al. trained CNN (ResNet) and ViT-based backbones (Swin Transformer and ViTAEv2)\cite{rsp} in MillionAID \cite{millionaid}, examining the impact of RSP on various downstream tasks such as scene recognition, semantic segmentation, object detection, and change detection. Their findings indicate that RSP can improve performance in scene recognition tasks but has limitations in others. Wang et al. also proposed a 100M parameter model later \cite{wang2022advancing}, an advanced plain ViT with rotated varied-size window attention, pre-trained with unsupervised MAE method \cite{mae}, achieving sota performance on the DOTA-V1.0 \cite{dota} dataset with an 81.24 \% mAP and competitive results for downstream classification and segmentation tasks. However, their dataset and models are single-modality and therefore cannot utilize the supervision from text labels, suggesting potential improvements with VLMs using image-text paired datasets. Their models are the largest model in the field of RS so far. 

Moreover, numerous self-supervised pre-trained models have emerged in the remote sensing field since labeled data are rare in the RS domain. SeCo \cite{mañas2021seasonal} utilizes unlabeled data from multiple Earth locations and different times. By leveraging position invariance, the model learns transferable representations for remote sensing in a self-supervised manner. Ayush et al. \cite{ayush2022geographyaware} using spatially aligned images to construct temporal positive pairs for contrastive learning by injecting temporal and geographical information into MoCo-V2 \cite{mocov2}.  Vincenzi et al. \cite{vincenzi2020color} propose to leverage the high-dimensionality of spectral bands to reconstruct visible colors, thus learning the good RS feature with the self-supervised learning approach.

\subsubsection{GeoQA and GeoAI}

Mai et al. provided an amazing survey in GeoQA \cite{mai2021geographic} (Geographic Question Answering). GeoQA focuses on solving problems that require a comprehensive understanding of textual descriptions, visual diagrams, and theorem knowledge in the context of geography. Datasets in this domain such as GeoQA dataset \cite{chen2022geoqa}, GeoSQA \cite{huang2019geosqa}, Tourism \cite{Contractor_2021}, GADM \cite{punjani2021templatebased} which contains geometric problems with corresponding annotated programs that illustrate the problem-solving process. These datasets are designed to facilitate research on explicit and explainable reasoning in a multimodal context in the geographic domain.

GeoAI, geospatial artificial intelligence, is a rapidly emerging field that integrates AI techniques with geographic and geospatial data. GeoAI is particularly effective in harnessing vast amounts of spatial and non-spatial data, offering advantages such as large-scale analytics, automation, high accuracy, sensitivity in detecting subtle changes, noise tolerance, and rapid technological advancement. The field encompasses a wide range of applications, including large-scale image analysis using various types of data like satellite and drone images, street views, and geo-scientific data. GeoAI research aims to provide solutions that are more efficient, accurate, and capable of detecting new patterns and processes in geospatial data. One of the challenges in GeoAI is ensuring that models are interpretable, explainable, and generalizable. GeoQA focuses on the aspect of question answering in the geographic domain, and GeoAI is a broader field that integrates AI techniques with geospatial data for various applications, including analysis, modeling, and prediction.

\subsection{RS5M}
\subsubsection{PUB11}
\label{appendix:pub11}

\begin{figure}[ht]
    \centering
    \includegraphics[width=0.49\textwidth]{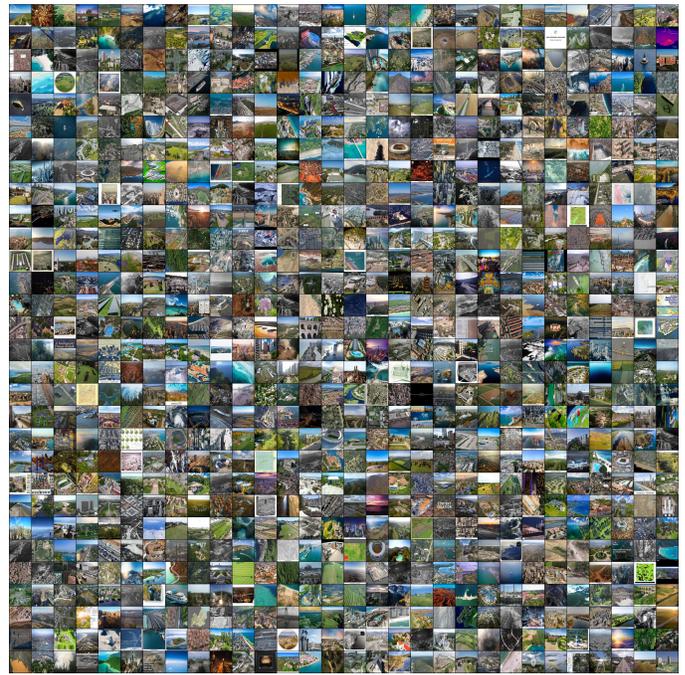}
    \caption{Visualization of images sampled from PUB11. Both aerial images and satellite images can be found, but the former is dominant in PUB11.}
    \label{fig:pub11_pure_image}
\end{figure}

The LAION2B-en dataset is an English subset of the well-known LAION5B \cite{laion5b}, collected by laion.ai and filtered by CLIP. It has 2.3 billion image-text pairs and is the largest publicly available image-text dataset so far. Similarly, LAION400M contains 400 million image-text pairs. LAIONCOCO\footnote{https://laion.ai/blog/laion-coco/} selects 600 million images from Laion2B-en, re-captioned using BLIP and CLIP to generate more descriptive captions. COYO700M collects 700 million informative image-alt-text pairs from HTML documents. Both CC3M and CC12M follow a similar collection process to COYO700M. YFCC15M is an English subset of YFCC100M, cleaned by Redford et al. in \cite{clip}. WIT is a multilingual Wikipedia-based image-text dataset, from which we only select English data for our dataset. Redcaps is a web-curated image-text dataset, primarily sourced from Reddit. SBU is collected from Flickr using a vast number of queries and then filters the noisy results. Visual Genome is an image dataset containing structured image concepts such as region descriptions, object instances, relationships, and more. Although both CC3M and CC12M datasets are publicly accessible and designed to be disjoint, some duplicate images may still exist, as is the case for LAION2B and LAION400M. Visualization of images sampled from PUB11 can be found in Figure \ref{fig:pub11_pure_image}.

\begin{table}[htbp]
\caption{The statistics of 11 public image-text paired datasets, "KF" represents "Keywords Filter", and "RVLMFD" means "Removed by VLM filter and RS Image Detector"}
\label{appendix:pubdataset_stat}
\centering 
\small
\setlength{\tabcolsep}{1.0pt}
\rowcolors{2}{white}{lightgray} 
\begin{tabular}{|c|c|c|c|c|c|}\hline
Name & Size & \# AfterKF  & \# Download & \# RVLMFD & \# Remain\\ \hline
\textit{LAION2B} & 2.3B &1,980,978 &1,737,584 & 333,686 & 1,060,779 \\
\textit{COYO700M} & 746M &680,089&566,076 & 94,329 & 226,069 \\
\textit{LAIONCOCO} & 662M &3,014,283&2,549,738 & 527,941 & 1,604,028\\
\textit{LAION400M} & 413M &286,102&241,324 & 23,860 & 75,781\\
\textit{WIT} & 37 M &98,540&93,754 & 9,299 & 10,374 \\
\textit{YFCC15M} & 15M &27,166&25,020 & 15,126 & 9,629\\
\textit{CC12M} & 12M &18,892&16,234 & 4,330 & 10,034\\
\textit{Redcaps} & 12M &2,842&2,686 & 972 & 1,486\\
\textit{CC3M} & 3.3M &12,563&11,718 & 1,817 & 9,572\\
\textit{SBU} & 1M &102&91 & 36 &51\\
\textit{VG} & 0.1M &26&26 & 20 & 6\\
\textit{Total} & 4.2B &6,121,583&5,244,251 & 1,011,416 & 3,007,809\\ \hline
\end{tabular}
\end{table}

\subsubsection{Keywords for Keyword Filtering} \label{keywords}
\textbf{Group 1}:    "remote sensing", "earth observ", "aerial imag", "aerial photo", "aerial map", "aerial pic", "aerial view", "aerial scan", "aerial satellite", "satellite imag", "satellite photo", "satellite map", "satellite pic", "satellite view", "satellite scan", "satellite data", "satellite surveillance", "space photo", "spaceborne photo", "space-borne photo", "space imag", "spaceborne imag", "space-borne imag", "space view", "spaceborne view", "space-borne view", "space surveillance"

\textbf{Group 2}:    "Google Earth", "Freesound", "Sentinel-1", "Sentinel-2", "Gaofen", "USGS", "NAIP", "MODIS", "EOSDIS", "WorldView", "Planet Dove", "ArcGIS", "Maxar", "Landsat", "Geographic Information System",

\subsubsection{Remote Sensing Prompt Template} \label{rs_pt}

\textbf{rs\_templates} = [
        'a remote sensing image.',
        'a low resolution remote sensing image.',
        'a bad remote sensing image.',
        'a cropped remote sensing image.',
        'a bright remote sensing image.',
        'a dark remote sensing image.',
        'a close-up remote sensing image.',
        'a black and white remote sensing image.',
        'a jpeg corrupted remote sensing image.',
        'a blurry remote sensing image.',
        'a good remote sensing image.',
        'an aerial image.',
        'a low resolution aerial image.',
        'a bad aerial image.',
        'a cropped aerial image.',
        'a bright aerial image.',
        'a dark aerial image.',
        'a close-up aerial image.',
        'a black and white aerial image.',
        'a jpeg corrupted aerial image.',
        'a blurry aerial image.',
        'a good aerial image.',
        'a satellite image.',
        'a low resolution satellite image.',
        'a bad satellite image.',
        'a cropped satellite image.',
        'a bright satellite image.',
        'a dark satellite image.',
        'a close-up satellite image.',
        'a black and white satellite image.',
        'a jpeg corrupted satellite image.',
        'a blurry satellite image.',
        'a good satellite image.',
    ]

\subsubsection{Remote Sensing Binary Classification Dataset} \label{bi_cls_dataset} 
We select 2500 satellite images from MillionAID, 2500 aerial images from LAION2B as positive data, and 5000 non-RS images from ImageNet-1k as negative data. We split the dataset with the ratio of 7:1:2 for the train, validation, and test set. Classes are balanced in each split. The trained classifier achieves 99.20\% in the validation set, and 97.55\% in the test set.

\subsubsection{Detail on Filtering Large-Scale Image-Text Paird Datasets}
\label{appendix:detail_filter_pub}
When downloading the images from URLs, some images may be missing due to broken links. For invalid image checking and deduplication, we filter out images that cannot be opened or have a zero file size. 

When utilizing fastdup to detect and cluster duplicate images, we use the cosine similarity distance function and set the number of nearest neighbors to 5. Additionally, we set the $"min\_distance"$ parameter in the $connected\_components()$ API to 1. For each cluster of duplicate images, to select the image we keep, we establish a set of priority rules. Initially, we discard images which are from the "laioncoco" dataset, as it contains long and redundant captions (The caption for images from laioncoco dataset are generated by concatenating many captions generated by BLIP \cite{blip}, and most of them are not informative). Then, we verify if there is an image from the "laion2b", "laion400m," or "coyo700m" datasets, which will constitute the training set of our RS5M dataset. If neither of the previous conditions is met, we randomly select an image.

\begin{figure}[H]
    \centering
    \includegraphics[width=0.49\textwidth]{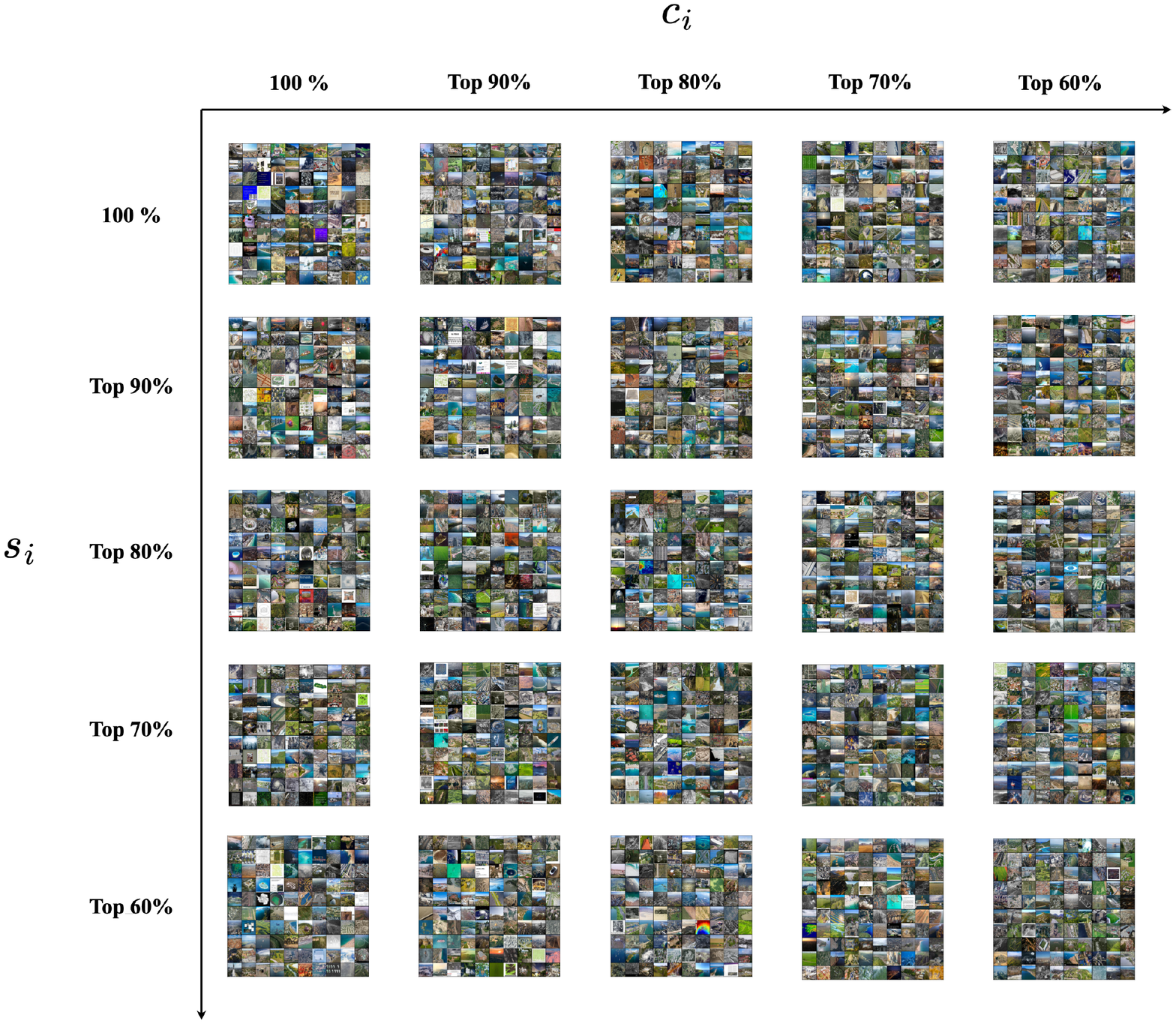}
    \caption{A image sample of thresholding with different quantiles of $s_i$ and $c_i$, top to bottom are the values of $s_i$ in descending order, and left to right are the values of $c_i$ in ascending order. Filtering with top 90\% $s_i$ and 80\% $c_i$ can already give us a good remote sensing dataset (center block).}
    \label{fig:grid_random}
\end{figure}

A demonstration of the VLM filtering results can be seen in Figure \ref{fig:grid_random}. In each of the nine blocks, we filter the dataset using fixed $m$ and $n$ and randomly sample 100 images to show. We apply thresholds to $s_i$ (from top to bottom) and $c_i$ (from left to right) to keep images that have top 100\% (no threshold), top 90\%, and top 80\% $s_i$ and $c_i$. 

\begin{figure}[htbp]
    \centering
    \includegraphics[height=0.49\textwidth]{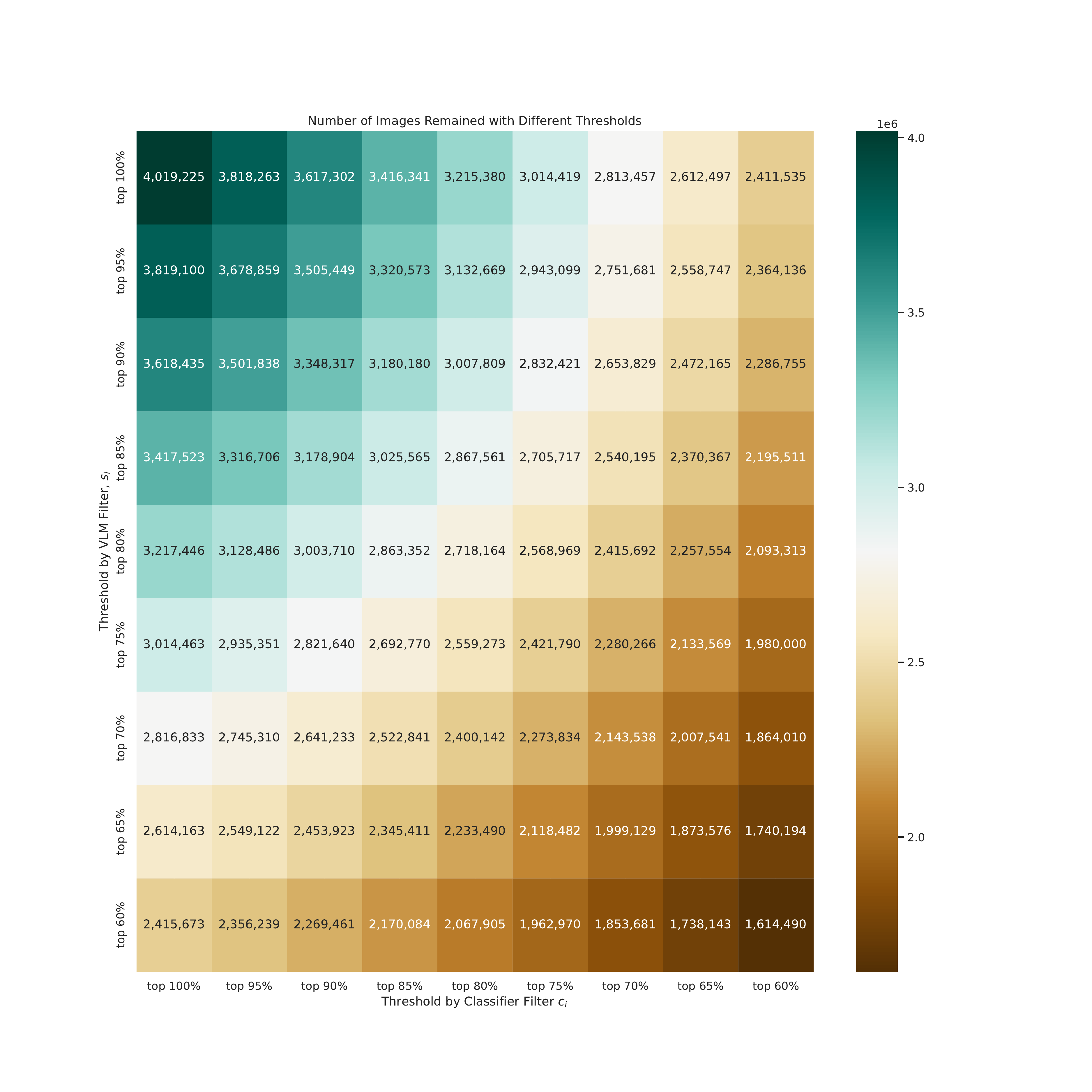}
    \caption{A heatmap indicates the number of remaining images for different combinations of thresholds. From top 100\% (no threshold) to top 60\%, with an decrement of 5\%}
    \label{fig:num_image_heatmap}
\end{figure}

\begin{figure}[htbp]
    \centering
    \includegraphics[width=0.49\textwidth]{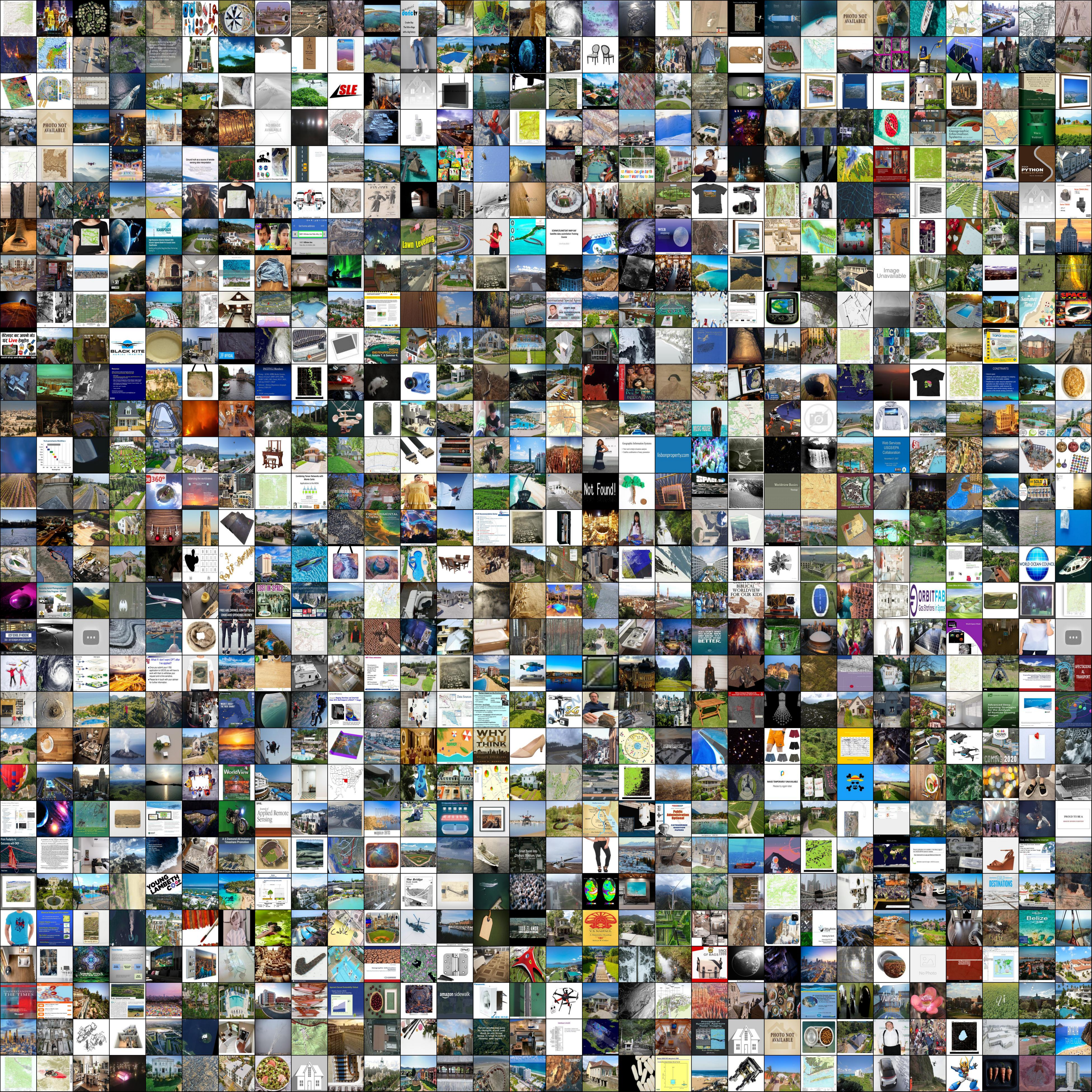}
    \caption{Images filtered by VLMFRSD having bottom 10\% $s_i$ and bottom 20\% $c_i$.}
    \label{fig:pub11_vlmcf_removed_900}
\end{figure}

A heatmap shows the number of remaining images for different thresholds is shown in Figure \ref{fig:num_image_heatmap}. It illustrates the trade-off between dataset noise and the number of images. We choose a group of loose thresholds (top 90\% $s_i$ and top 80\% $c_i$) to retain more images while addressing the outliers in the next section. After this step, 3,007,809 images remain in the dataset. Samples of filtered images (outliers) processed by Vision-Language Model Filtering and Classifier Filtering can be found in Figure \ref{fig:pub11_vlmcf_removed_900}.

Finally, the images removed by all kinds of processing methods are listed in Table \ref{table:pubdataset_step_stat}.

\begin{table}[H]
\caption{The statistics of public image-text paired datasets after each processing step, "RIIC" means "Removed by Invalid Image Checking", "RDIF" denotes "Removed by Duplicate Image Filtering (URL+Fastdup)", "RVLMFRSD" means "Removed by VLM filter and RS Image Detector".}
\label{table:pubdataset_step_stat}
\centering 
\rowcolors{2}{white}{lightgray} 
\begin{tabular}{|c|c|c|c|c|c|}\hline
Name & \# RIIC & \# RDIF  & \# RVLMFRSD & \# Remain \\\hline
\textit{LAION2B} & 102 & 343,017 & 333,686 & 1,060,779\\
\textit{COYO700M} & 28 & 245,650& 94,329  & 226,069 \\
\textit{LAIONCOCO} & 80 & 417,689& 527,941 & 1,604,028  \\
\textit{LAION400M} & 25 & 141,658 & 23,860 & 75,781 \\
\textit{WIT} & 0 & 74,081 & 9,299 & 10,374  \\
\textit{YFCC15M} & 0 & 265 & 15,126 & 9,629\\
\textit{CC12M} & 0 & 1,870 & 4,330 & 10,034\\
\textit{Redcaps} & 0 & 228 & 972 & 1,486\\
\textit{CC3M} & 1 & 328 & 1,817 & 9,572\\
\textit{SBU} & 0 & 4 & 36 & 51 \\
\textit{VG} & 0 & 0 & 20 & 6\\
\textit{Total} & 236 & 1,224,790 & 1,011,416 & 3,007,809 \\ \hline
\end{tabular}
\end{table}

\subsubsection{RS3}
\label{appendix:RS3}

\begin{figure}[H]
    \centering
    \includegraphics[width=0.49\textwidth]{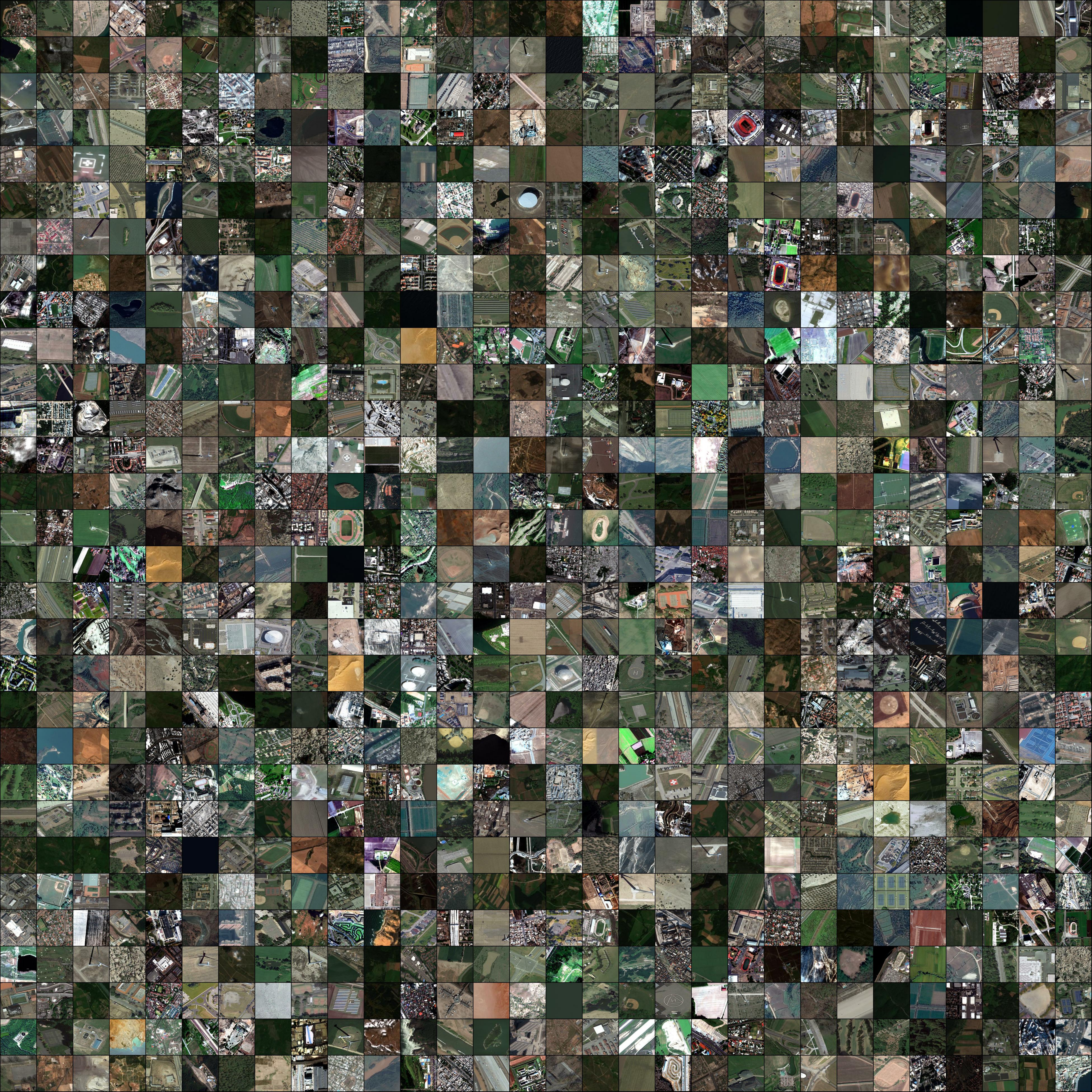}
    \caption{Visualization of images sampled from RS3. Almost all of them are satellite images.}
    \label{fig:rs3_pure_image}
\end{figure}

Visualization of images sampled from RS3 can be found in Figure \ref{fig:rs3_pure_image}.

\begin{figure}[H]
    \centering
    \includegraphics[width=0.49\textwidth]{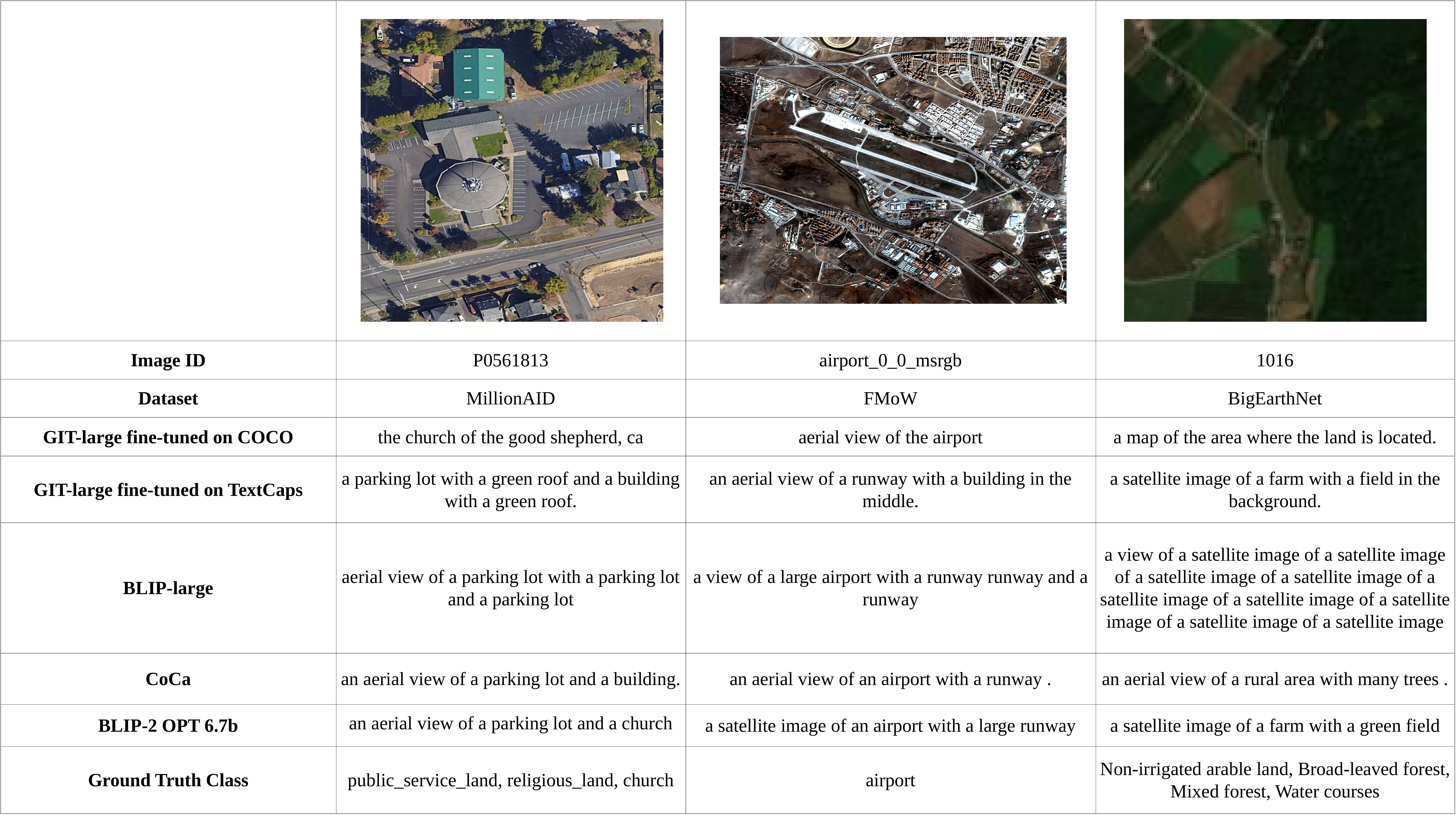}
    \caption{Captioning results in comparison for images from 3 large-scale remote sensing datasets using various captioning models from huggingface.} 
    \label{fig:caption_compare}
\end{figure}

Captioning RS images with VLMs pre-trained on images with common objects has proven to be effective, as demonstrated in \cite{blip} and Figure \ref{fig:caption_compare}. Although some captioning models, such as BLIP-Large and GIT-Large, may generate repetitive or nonsensical captions, other models like CoCa and BLIP2 can produce impressive and meaningful captions for RS images. For instance, these models are capable of recognizing concepts like "church," "airport," and "farm" in satellite views. Image-text pairs collected through this process are primarily satellite images. 

\footnotetext{https://huggingface.co/spaces/nielsr/comparing-captioning-models}

\begin{figure}[H]
    \centering
    \includegraphics[width=0.49\textwidth]{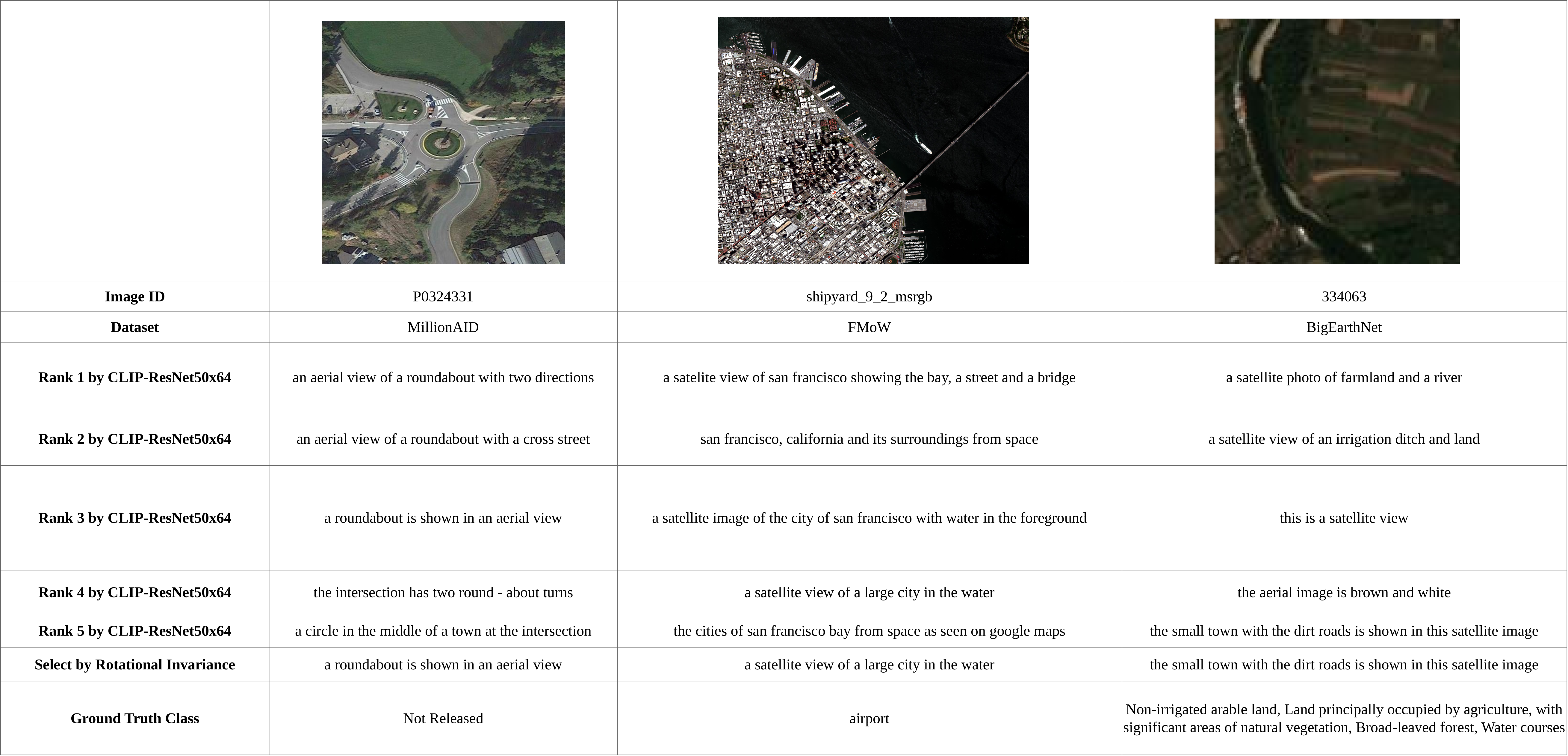}
    \caption{Top 5 captioning results using BLIP2 model for images from MillionAID, FMoW, and BigEarthNet.} 
    \label{fig:blip2_captioning_result}
\end{figure}

Examples of captioning MillionAID, BigEarthNet, and FMoW datasets with different Vision-Language Models are provided in Figure \ref{fig:blip2_captioning_result}.

\subsubsection{Captioning Result with Different Sampling Method}
\label{sampleing_method}

\textbf{Image Sample}: 

\begin{figure}[H]
    \centering
    \includegraphics[width=0.49\textwidth]{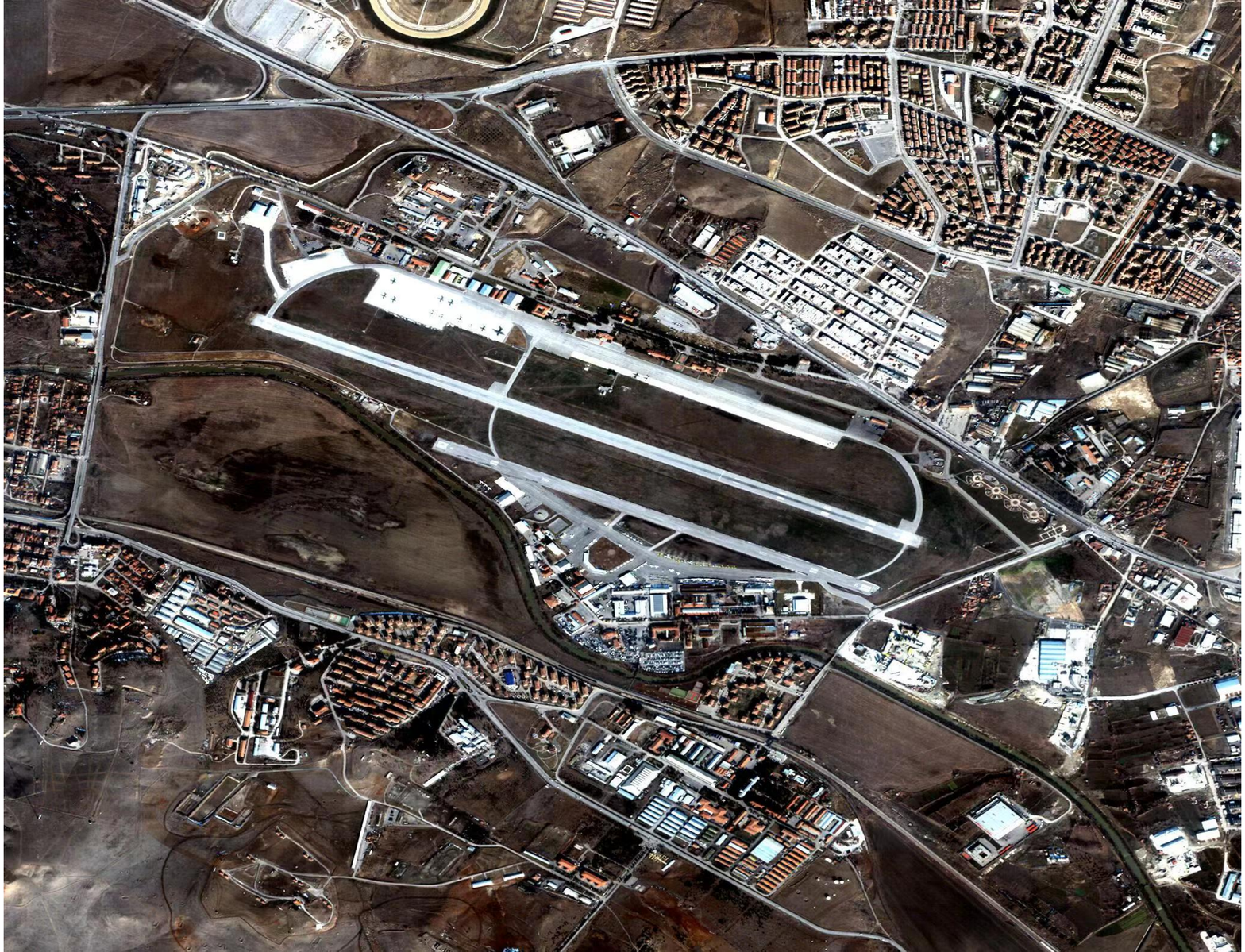}
    \caption{An Airport in Turkey in the satellite view, select from the FMoW dataset.}
    \label{fig:airport_turkey}
\end{figure}

\textbf{beam search caption result}: ['an aerial view of an airport in the middle of a city', 'a satellite image of an airport in the middle of a city', 'a satellite view of an airport in the middle of a city', 'an aerial view of a large airport in the middle of a city', 'a satellite image of a large airport in the middle of a city', 'an aerial view of a small airport in the middle of a city', 'an aerial view of an airport in the middle of the city', 'a satellite image of an airport in the middle of a town', 'a satellite image of an airport in the middle of the city', 'an aerial view of an airport in the middle of a town', 'an aerial view of an airport in the middle of a large city', 'a satellite view of an airport in the middle of the city', 'a satellite view of an airport in the middle of a town', 'a satellite image of an airport in the middle of a large city', 'an aerial view of an airport and surrounding area', 'an aerial view of a small airport in the middle of a large city', 'a satellite image of an airport and surrounding area', 'an aerial view of an airport in the middle of a field', 'a satellite image of an airport in the middle of a field', 'a satellite view of an airport in the middle of a large city']

\textbf{nucleus sampling caption result}: ['a satellite image shows a small airport', 'a satellite image of an airport with several runways and buildings', 'a satellite image of an airport', '\textbf{aerial views of airfields in turkey}', 'nigeria, zaria - airbase and airforce base, zaria', 'a satellite image of an airport and city in the middle', 'a satellite image of an airport in a city', 'a satellite view showing a runway and airport terminal', 'a satellite image shows an aerial view of airport', 'aerial photograph of small airplane on the runway', 'map aeropuerto de grecia zante aegean', 'a city with a small airport in the background', 'aerial view of an airport in a satellite image', 'this is an aerial image of the airport near town', 'aerial view of a city airport and its airport parking lots', 'a satellite image shows the airport and its surroundings', 'a google satellite image of an airport', 'gis map of fethiye international airport by fethiye international airport, istanbul, turkey', 'satellite view of the airport area with some airplanes on it', 'airport atatürk airport, turkey, satellite image'], ['cairo international airport from the air', 'a satellite image shows an airport near a city', 'view of the airport from above', 'a map is displayed as it shows an aerial view of a city', 'the area where planes land in a small town', 'google satellite map image of a airport on a grass field', 'a satellite image of the airport with its runway', 'this is a satellite image of a large airport', 'a picture of an airport with a plane sitting on the ground', 'aerial view of the airport that looks like it has a lot of aircraft on the ground', 'a satellite photo of an airport in the middle of a large city', 'aerial view of an airport next to a city with a couple of large buildings', 'the aerial photo shows an airport from above', 'a satellite view shows many aircraft and airplanes', 'a satellite photo of an airport in a rural area', 'a satellite map shows the location of an airport in iraq', 'a aerial photo of the airport in the middle of an arid landscape', 'an aerial view of an airport near a village', 'this is an aerial view of an airport in the middle of a field', 'a satellite image of an air base and a road']

\subsubsection{Rotational Invariance}
\begin{figure}
    \centering
    \includegraphics[height=4.5cm]{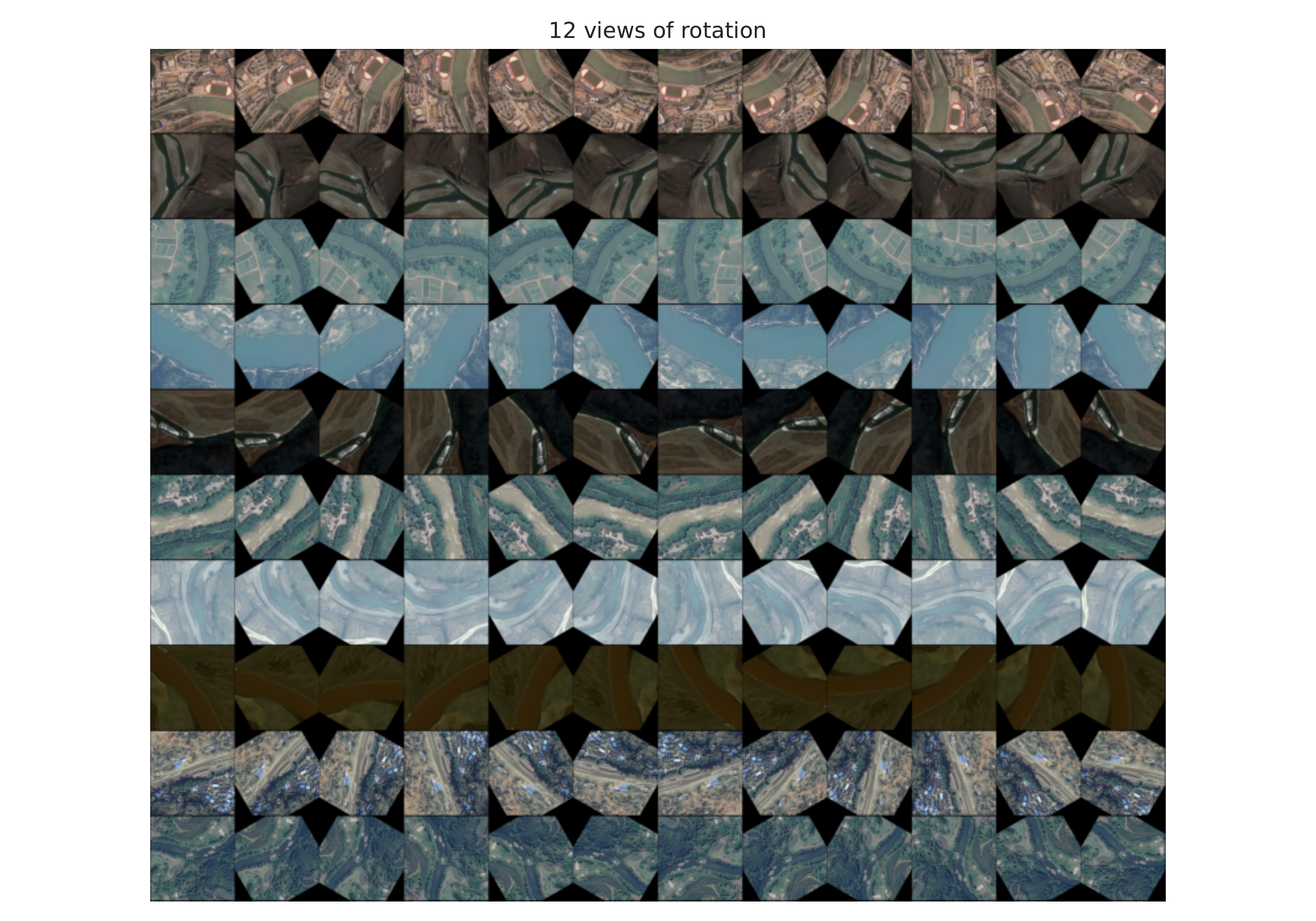}
    \caption{Visualization for image rotation in 12 different angles}
    \label{fig:rotation}
\end{figure}

See Figure \ref{fig:rotation} for examples.
\label{appendix:rotational_invariance}

\subsubsection{Combine Machine Generated Caption with Image Meta Information}
\label{appendix:metacap}

We enhanced the caption from RS5M by integrating meta information (geometa, class labels, UTC, etc.) into readable sentences as a part of image caption. This structured meta-caption, combined with the model-generated caption, offers a more comprehensive view. We believe this incorporation of image meta information not only augments our dataset's richness but also aids in drawing more precise insights from it. The datasets and their utilized meta info are listed below:

\begin{itemize}
    \item \textbf{FMoW}
    
    Included Meta Info: Longitude, latitude, class label, bounding box coordinates, ground sample distance, UTM zone, timestamp, cloud cover rate, scan direction, target azimuth, and off-nadir.

    Geographic Details: city, country.
    
    Temporal Details: season (color of tree/leaves, snow-covered or not, etc.), timestamp.
    
    Image Specifics (for objects in the image): class labels, relative location (Top/Centre/Bottom \& Left/Centre/Right).
    
    Additional Details: ground sample distance, UTM zone, cloud cover rate, scan direction, target azimuth, off-nadir.

    \item \textbf{BigEarthNet}
    
    Included Meta Info: Class labels, timestamp, UTM zone.
    
    Temporal Details: season, timestamp.
    
    Image Specifics (for objects in the image): class labels.
    
    Additional Details: UTM zone.

    \item \textbf{YFCC14M}
    
    Included Meta Info: Date taken, longitude, latitude.

    Geographic Details: city, country.
    
    Temporal Details: season, timestamp.

    \item \textbf{CC3M}

    Included Meta Info: Machine tags (aligned with caption).
    
    Image Specifics: class labels.

    \item \textbf{Redcaps}

    Included Meta Info: Created\_utc (UNIX-timestamp).
    
    Note: As the timestamp denotes blog creation and not image capture, we've decided to exclude this meta info.

\end{itemize}

\begin{table}[H]
\caption{Statistics of meta caption per dataset}
\centering 
\begin{tabular}{|c|c|c|c|c|c|}\hline
  Dataset & FMoW  & BigEarthNet & YFCC14M & CC3M & RedCaps \\ \hline
  Count & 727,144 & 344,385 & 9,629 & 2,487 & 1,486 \\ \hline
\end{tabular}
\end{table}

\subsubsection{Tuned BLIP2 Details}
\label{appendix:tunedblip2}
The BLIP2 model is not good at RS captioning task compared with the common objects captioning task (MSCOCO obtains METEOR score of 0.1506 using BLIP2-vanilla, but RSICD and RSITMD only have 0.0687 and 0.0625). We enhanced the BLIP2 opt6.7B model by tuning it (vision encoder) using LoRA with the RS-specific data from the RSITMD dataset(training set). 

To evaluate the improvement in the quality of captions with respect to Remote Sensing (RS), we assessed the METEOR score across three test sets: RSVG test set, RSCID test set, and RSITMD test set. By comparing the performance of the original BLIP2 ("BLIP2-vanilla") and the refined BLIP2 ("BLIP2-RS"), we observed a marked enhancement in the latter's capability to generate RS-related captions.

\begin{table}[H]
\caption{The results of METEOR score for RSVG, RSICD, RSITMD and MSCOCO(test sets)}
\centering 
\begin{tabular}{|c|c|c|c|c|c|c|}\hline
    \textbf{Model/Dataset} &	MSCOCO	& RSICD	& RSITMD	& RSVG \\ \hline
  BLIP2-vanilla &	0.1506	& 0.0687	& 0.0625&	0.0949\\ \hline
  BLIP2-RS &	- &	0.1528 &	0.1420 &	0.1301 \\ \hline
\end{tabular}
\end{table}

\subsubsection{Rating System for Model Generated Captions}
\label{appendix:ratingexp}
We designed a 5-level rating system from 3 major perspectives for evaluating the sampled caption (~2000 samples for now, we will continue adding the sample size) \footnote{https://github.com/om-ai-lab/RS5M/tree/main/rating\_app}.

\begin{figure}[H]
    \centering
    \includegraphics[width=0.49\textwidth]{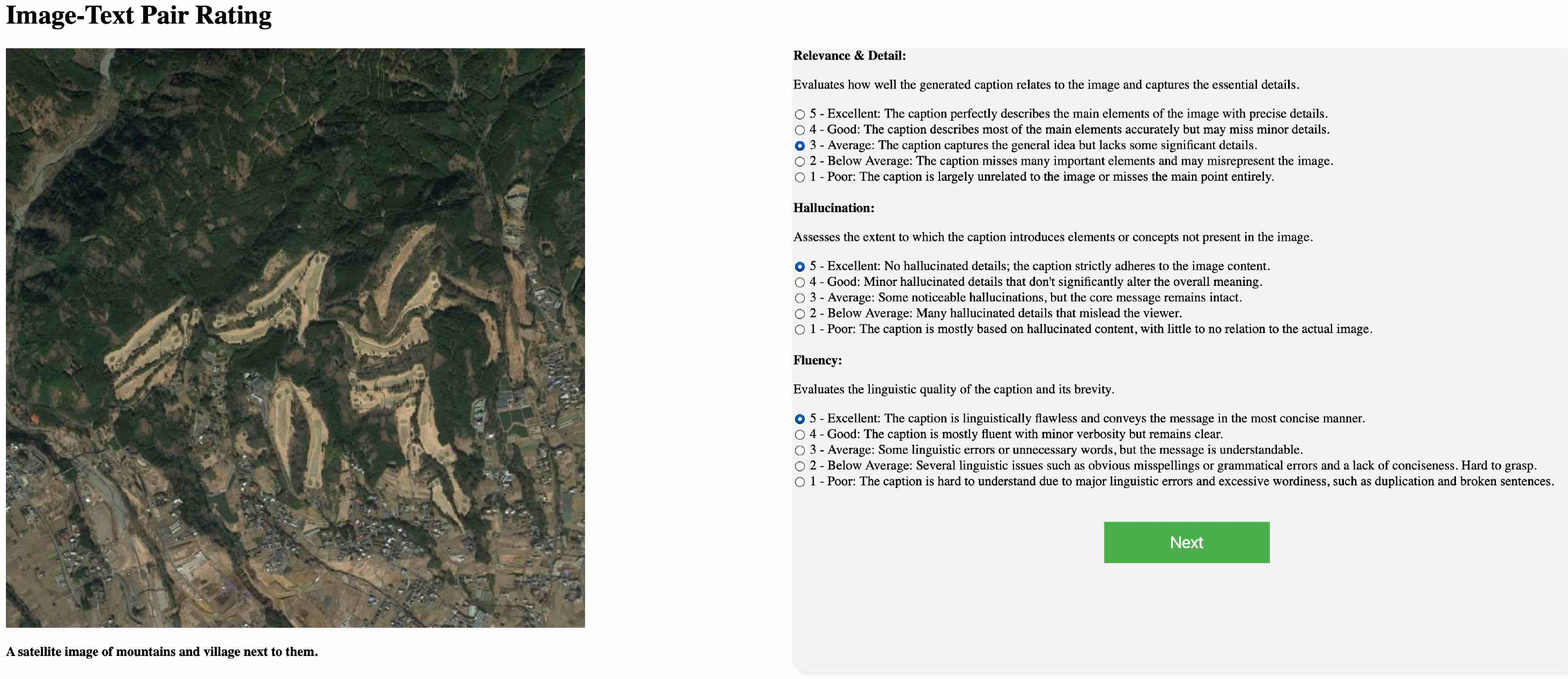}
    \caption{A snapshot of our rating system.}
\end{figure}

\textbf{Relevance \& Detail}
Evaluates how well the generated caption relates to the image and capture the essential details.

5 - Excellent: The caption perfectly describes the main elements of the image with precise details.

4 - Good: The caption describes most of the main elements accurately but may miss minor details.

3 - Average: The caption captures the general idea but lacks some significant details.

2 - Below Average: The caption misses many important elements and may misrepresent the image.

1 - Poor: The caption is largely unrelated to the image or misses the main point entirely.

\textbf{Hallucination}
Assesses the extent to which the caption introduces elements or concepts not present in the image.

5 - Excellent: No hallucinated details; the caption strictly adheres to the image content.

4 - Good: Minor hallucinated details that don't significantly alter the overall meaning.

3 - Average: Some noticeable hallucinations, but the core message remains intact.

2 - Below Average: Many hallucinated details that mislead the viewer.

1 - Poor: The caption is mostly based on hallucinated content, with little to no relation to the actual image.

\textbf{Fluency \& Conciseness}:
Evaluates the linguistic quality of the caption and its brevity.

5 - Excellent: The caption is linguistically flawless and conveys the message in the most concise manner.

4 - Good: The caption is mostly fluent with minor verbosity but remains clear.

3 - Average: Some linguistic errors or unnecessary words, but the message is understandable.

2 - Below Average: Several linguistic issues such as obvious misspellings or grammatical errors and a lack of conciseness, make it harder to grasp.

1 - Poor: The caption is hard to understand due to major linguistic errors and excessive wordiness, such as duplication and broken sentences.

The results are shown below:

\begin{table}[H]
\caption{Rating Scores: Overall}
\centering
\begin{tabular}{|c|c|c|c|c|c|c|}\hline
   & relevance\_detail &	hallucination &	fluency\\ \hline
   mean &	4.53 &	4.73 & 4.94\\ \hline
   std &	0.69 &	0.66 & 0.32\\ \hline
\end{tabular}
\end{table}

\begin{table}[H]
\caption{Rating Scores: Relevance \& Detail}
\centering
\begin{tabular}{|c|c|c|c|}\hline
   subset & count &	mean &	std\\ \hline
   ben	 &310	 &4.41 &	0.76\\ \hline
   fmow	 & 617 &	4.85 &	0.36\\ \hline
   millionaid &	878 &	4.34 &	0.75 \\ \hline
\end{tabular}
\end{table}

\begin{table}[H]
\caption{Rating Scores: Hallucination}
\centering
\begin{tabular}{|c|c|c|c|}\hline
   subset & count &	mean &	std\\ \hline
   ben	 &310	 & 4.54 &	0.79\\ \hline
   fmow	 & 617 & 4.87 &	0.38\\ \hline
   millionaid &	878 & 4.70 & 0.73 \\ \hline
\end{tabular}
\end{table}

\begin{table}[H]
\caption{Rating Scores: Fluency}
\centering
\begin{tabular}{|c|c|c|c|}\hline
   subset & count &	mean &	std\\ \hline
   ben	 &310	 & 4.80 &	0.58\\ \hline
   fmow	 & 617 &	4.98 &	0.16\\ \hline
   millionaid &	878 &	4.97 &	0.24 \\ \hline
\end{tabular}
\end{table}

The overall caption quality is good after adding meta info to the caption. ("Relevance \& Detail" score is $4.53 / 5.00$ and "Hallucination" score is $4.73 / 5.00$)
FMoW subset has the best "Relevance \& Detail" score (4.85) and "Hallucination" score (4.87), with the lowest standard deviations (0.36 and 0.38). This result is greatly helped by adding Geographic details (city, country), Temporal details (season, timestamp), Image specifics ( class labels, relative location), and Additional details (ground sample distance, UTM zone, cloud cover rate, scan direction, target azimuth, off-nadir).
"Hallucination" scores are relatively low for BigEarthNet and MillionAID. This can be attributed to the image quality is not good in BigEarthNet (120x120 pixels, too blurry) and the MillionAID does not contain meta information (only main targets are described in most cases).
The overall caption fluency is good for all sub-datasets. The BLIP2 model is able to generate linguistically flawless captions and conveys the message in a concise manner in most cases.
For high-quality images, the BLIP2 model is able to generate more accurate captions that contain targets of interest.

\subsubsection{Visualization}
\label{appendix:visualization}
\begin{figure}[H]
    \centering
    \includegraphics[width=0.15\textwidth]{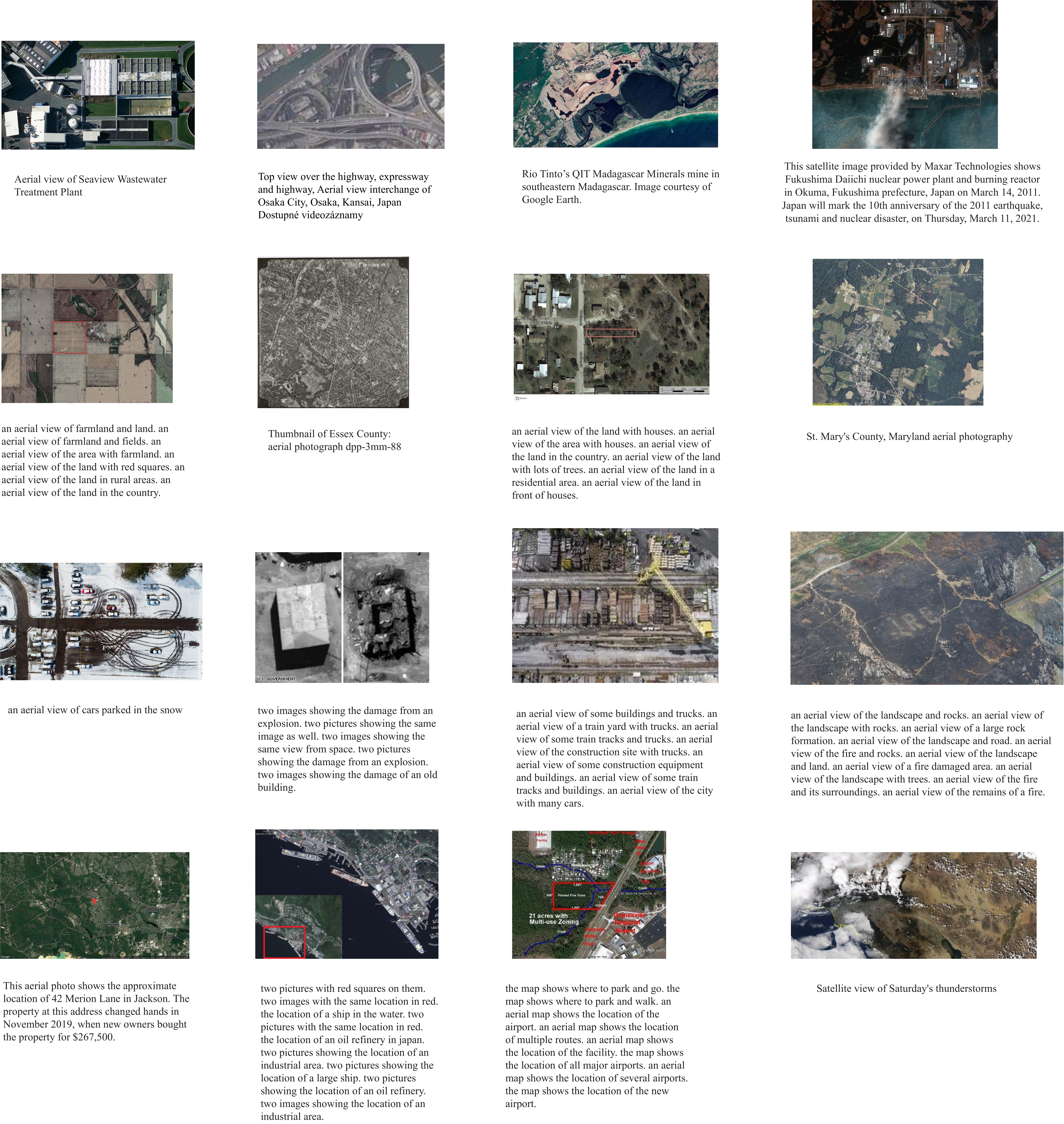}
    \includegraphics[width=0.15\textwidth]{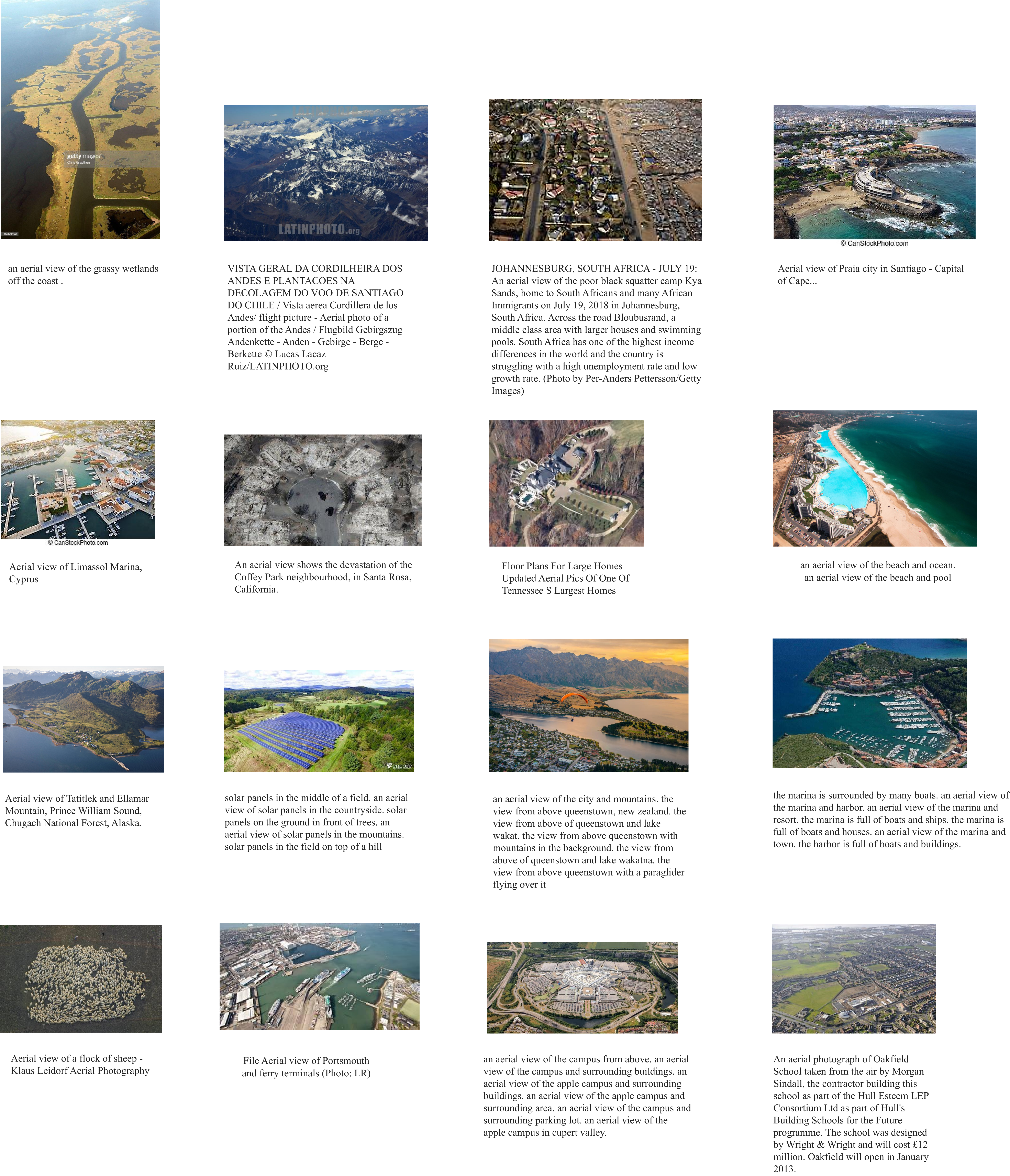}
    \includegraphics[width=0.15\textwidth]{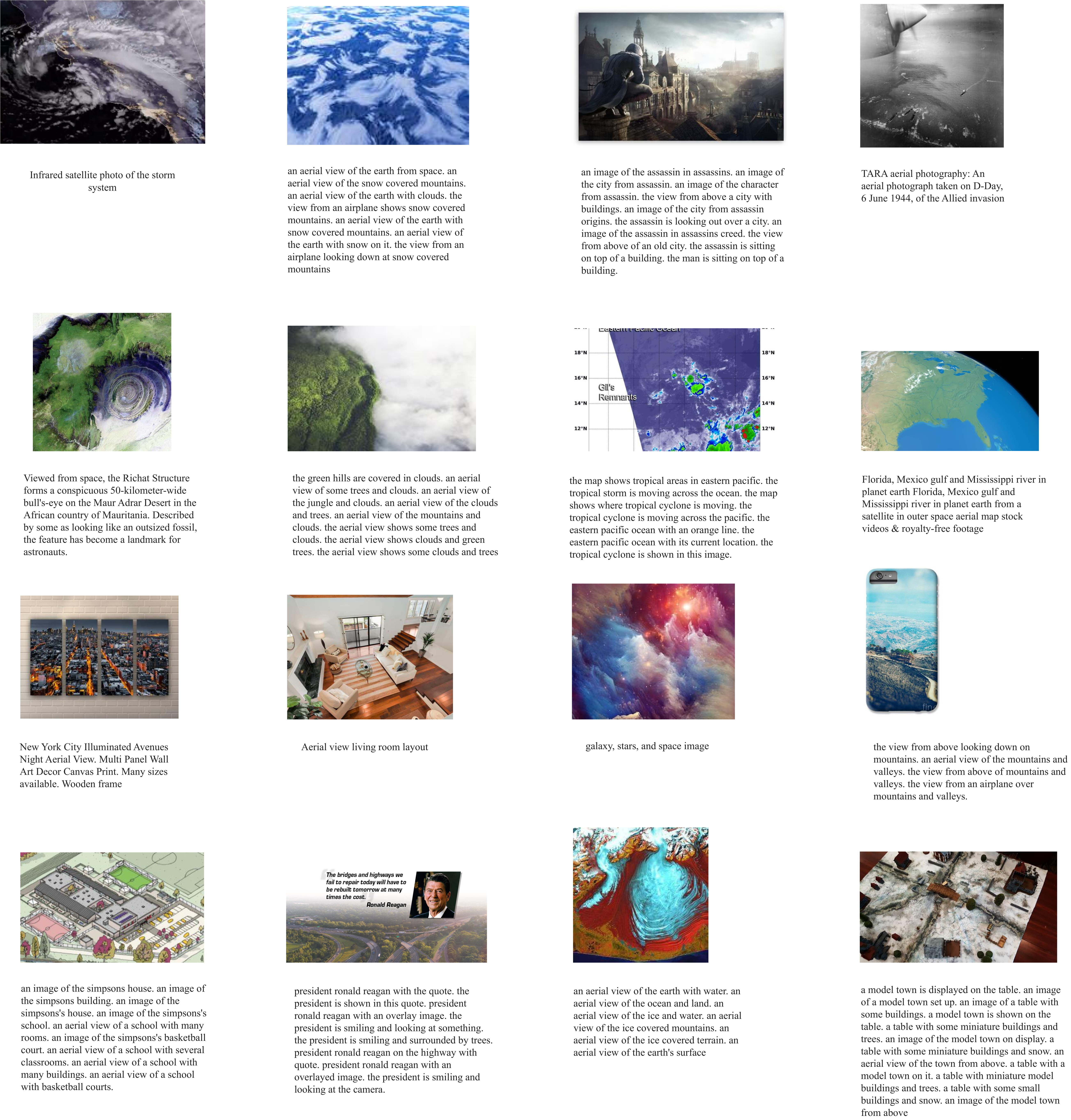}
    \caption{Pub11 images in satellite view (left 4 columns), aerial view (middle 4 columns), and outliers (right 4 columns).}
    \label{fig:pub11_presentation}
\end{figure}

We showcase a selection of image-text pairs from PUB11, dividing the dataset into three categories: aerial view, satellite view, and outlier groups. Left 4 columns in Figure \ref{fig:pub11_presentation} displays randomly sampled satellite images from PUB11, accompanied by their captions. For improved visual presentation, we have truncated longer captions. While not all captions are informative, they do relate to their corresponding images. Middle 4 columns in Figure \ref{fig:pub11_presentation} present images taken from aerial views within PUB11. In contrast to satellite images, aerial images are captured at lower altitudes, offering more detailed views of the ground. Additionally, the shooting angles differ significantly from those of satellite images. Right 4 columns in Figure \ref{fig:pub11_presentation} right features a selection of representative outliers and intriguing images. Most outliers consist of meteorological satellite images, illustrative figures, and space images, which are distinct from conventional aerial and satellite images. The dataset also includes some interesting outliers, such as an artwork depicting a city's nighttime scene and a model of town photograph. Although these images do not strictly fall within the realm of remote sensing, they do have delicate connections with the RS.

\begin{figure}[H]
    \centering
    \includegraphics[width=0.49\textwidth]{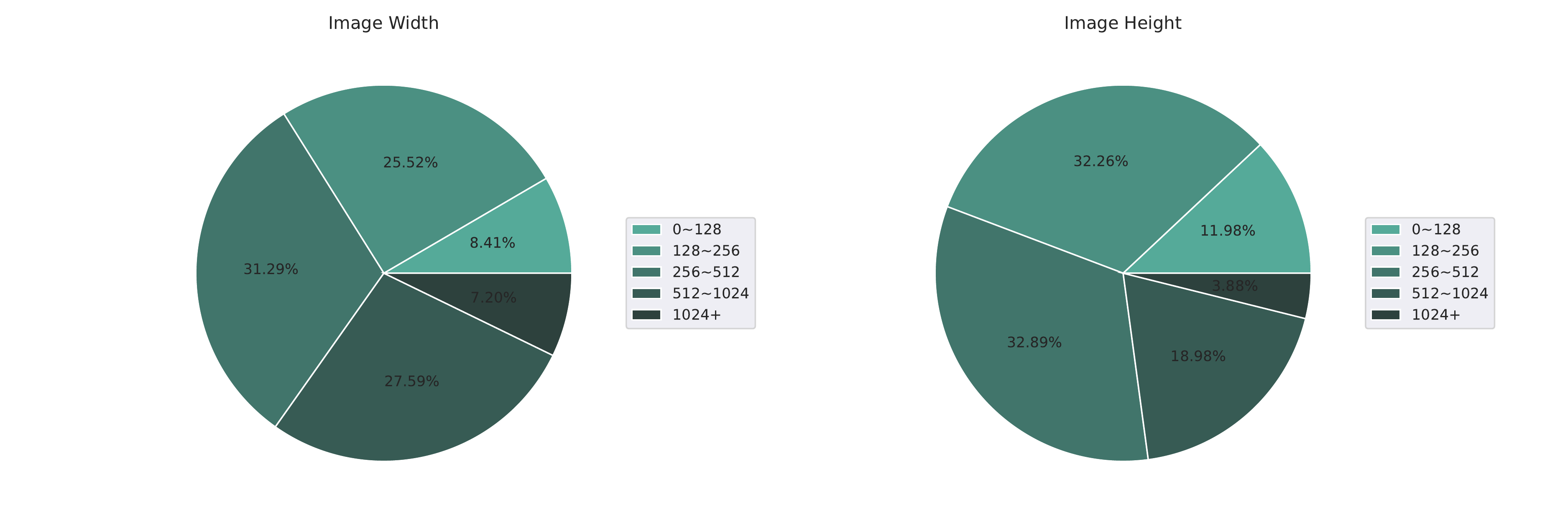}
    \caption{PUB11 image width and height statistics.}
    \label{fig:wh}
\end{figure}

We present the statistics for the width and height of images from PUB11 in Figure \ref{fig:wh}, the average height and width are 402.87 pixels and 522.53 pixels respectively.

\subsubsection{Outlier \& Misfiltered Image Analysis}
\label{appendix:outlier_analysis}
\begin{figure}[H]
    \centering
    \includegraphics[width=0.23\textwidth]{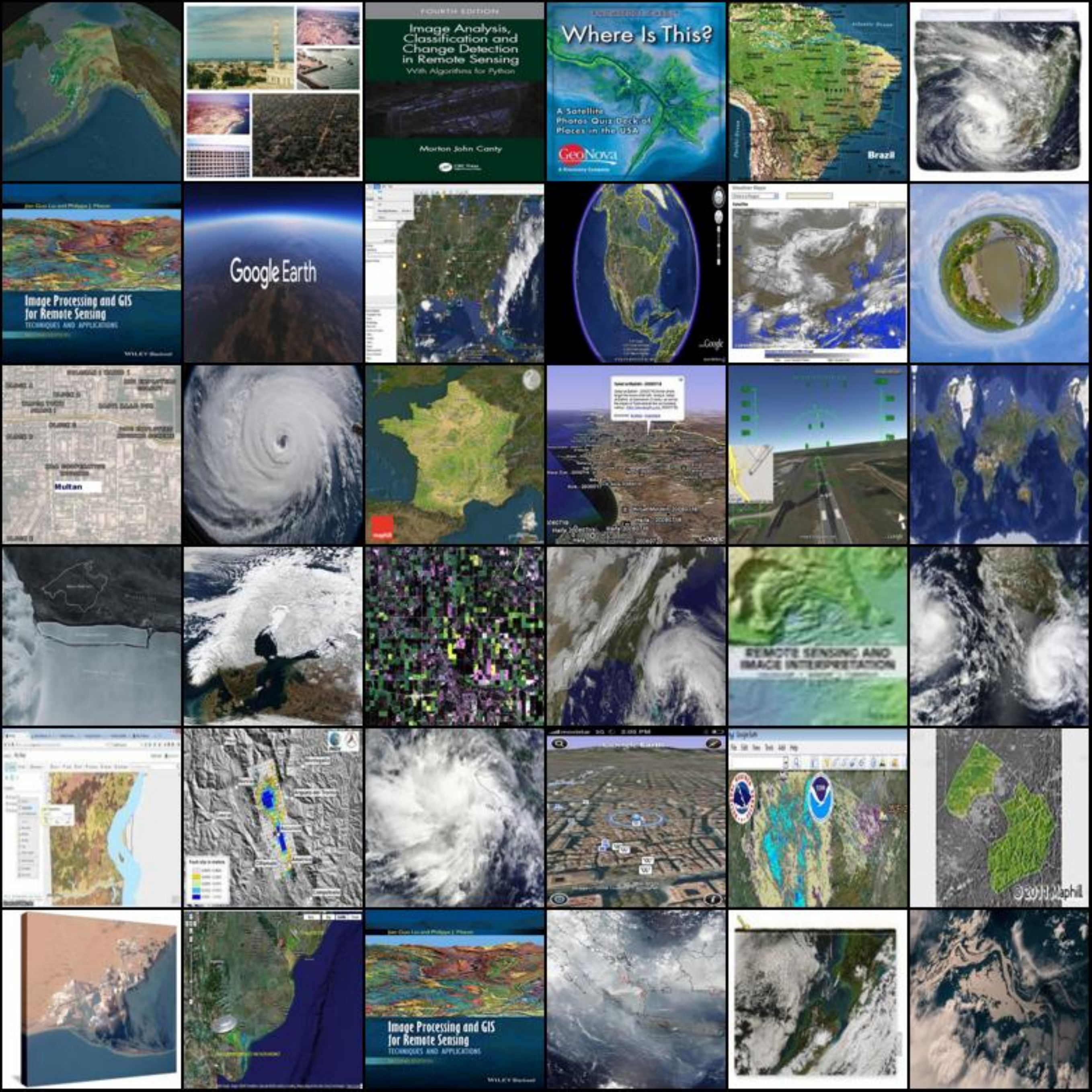}
    \includegraphics[width=0.23\textwidth]{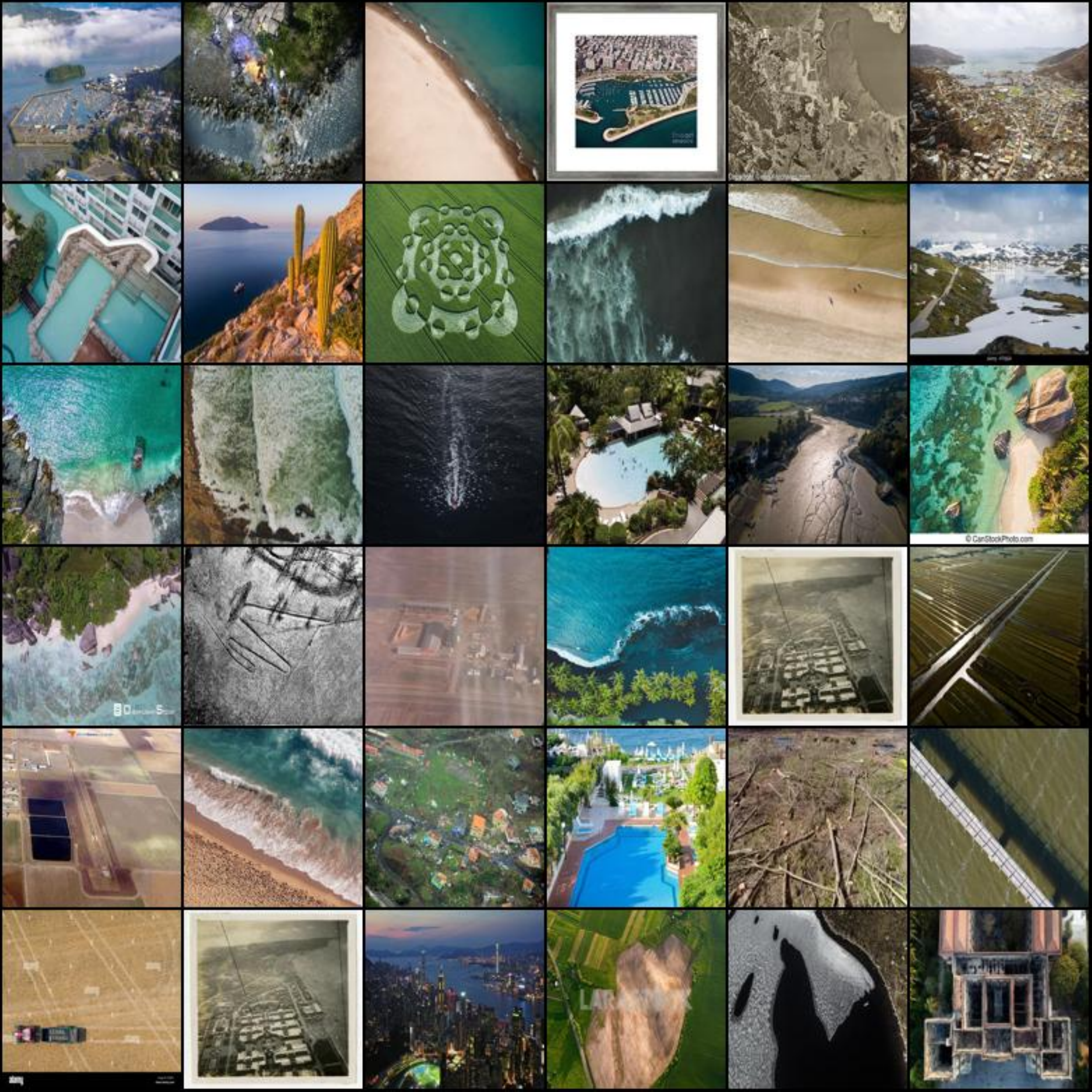}
    \caption{Visualization for outliers in RS5M (left) and misfiltered images from the removed image collection (right).}
    \label{fig:sample_outlier_misfiltered}
\end{figure}

Although we used multiple procedures to filter images that are not RS images. There are still some outliers remaining in the PUB11. In contrast, some RS images are over-filtered by the VLMFRSD process. To analyze this quantitatively, we sampled 5,000 images from PUB11 and 5,000 images from the removed image collection (from the VLMFRSD step) and assessed them one by one, which are 0.1\% from RS5M and 0.5\% from the removed collection. In Table \ref{table:outlier_misfiltered}, we list the confusion matrix for further discussion. Samples for outliers in RS5M and misfiltered images from the removed image collection are presented in Figure \ref{fig:sample_outlier_misfiltered}. Around 0.8\% of images in the sampled RS5M images are outliers, and 3.4\% of images from the removed collection are RS5M images. Most of the images in the former case are maps, illustrations, and weather imagery, and the images in the latter case are RS images shot from a low altitude.

\begin{table}[H]
\caption{The confusion matrix for RS images removed by VLMFRSD, and outliers in RS5M dataset.}
\label{table:outlier_misfiltered}
\centering 
\begin{tabular}{|c|c|c|}\hline
  & Recognized as RS image & Recognized as Outlier \\ \hline
\textit{Is RS image} & 4831 & 169 \\ \hline
\textit{Is Outlier} & 40 & 5960 \\ \hline
\end{tabular}
\end{table}

\subsubsection{Hardware \& Safety Check}
Our dataset was processed on a desktop computer equipped with a single NVIDIA RTX 4090 24GB GPU, a 16-core Ryzen 3950x processor, 64 GB of RAM, and 4TB SSD. The experiments are done with 1 NVIDIA A100 40GB GPUs (main experiments and ablations) for 4 weeks, and 1 RTX A100 80GB GPU (tuning stable-diffusion) for 1 week.

We utilized publicly available remote sensing data, which should not contain any NSFW content. To ensure safety, we converted the TIFF files to standard JPG format while removing sensitive geographical coordinates.

\subsection{License}
\label{appendix:license}

\begin{table*}[htbp]
\caption{License for data source of RS5M}
\label{table:license}
\centering 
\rowcolors{2}{white}{lightgray} 
\begin{tabular}{|c|c|c|}\hline
Dataset & License & Allow Redistribution \\ \hline
\textit{LAION2B} &  CC-BY-4.0 License & Yes \\
\textit{COYO700M} & CC-BY-4.0 License & Yes\\
\textit{LAIONCOCO} &  CC-BY-4.0 License & Yes \\
\textit{LAION400M} &  CC-BY-4.0 License & Yes \\
\textit{WIT} & CC-AS-3.0 Unported license & Yes \\
\textit{YFCC15M} & relevant Webscope License Agreement & Yes \\
\textit{CC12M} & The dataset may be freely used for any purpose & Yes\\
\textit{Redcaps} & Only be used for non-commercial research & Yes\\
\textit{CC3M}  & The dataset may be freely used for any purpose & Yes  \\
\textit{SBU}  & Unknown & Unknown\\
\textit{VG}  & CC-BY-4.0 License & Yes\\ 
\textit{BigEarthNet} & The Community Data License Agreement – Permissive & Yes\\
\textit{FMoW}  & Functional Map of the World Challenge Public License & Yes\\
\textit{Million-AID}  & Unknown & Unknown\\ 
\textit{RS5M}  & Only be used for non-commercial research  & Yes\\ \hline

\end{tabular}
\end{table*}

As shown in Table \ref{table:license}, almost all involved datasets allow redistributing the metadata. We have sent several emails to the authors of the MillionAID dataset, but got no response. For the PUB11 subset, we plan to release the metadata of PUB11 first. For RS3, since BigEarthNet and FMoW allow the redistribution of image data, we plan to release the meta file and image-text tar file in the webdataset format. We will claim that our RS5M dataset is only allowed to be used for academic purposes, and we bear all responsibility in case of violation of rights. We will take appropriate action when needed, e.g. to remove data with such issues. We will host our RS5M in Baidu Disk for at least 1 year. A copy has been uploaded to Google Drive as well.

\subsection{Stable Diffusion Tuned with RS5M}
\label{appendix:sd}

Given the impracticality of training the Stable Diffusion model from scratch with only 5M data, we present a Stable Diffusion model tuned by 1\% data of RS5M, which we refer to as RS-SD. Specifically, we use Dreambooth \cite{dreambooth} from a modified Diffuser repository \cite{diffusers}\footnote{https://github.com/ShivamShrirao/diffusers}. The image resolution is set to 512, with a batch size of 50 for 50,000 steps. The text encoder was trained as well. 

\begin{figure}[htbp]
    \centering
    \includegraphics[width=0.49\textwidth]{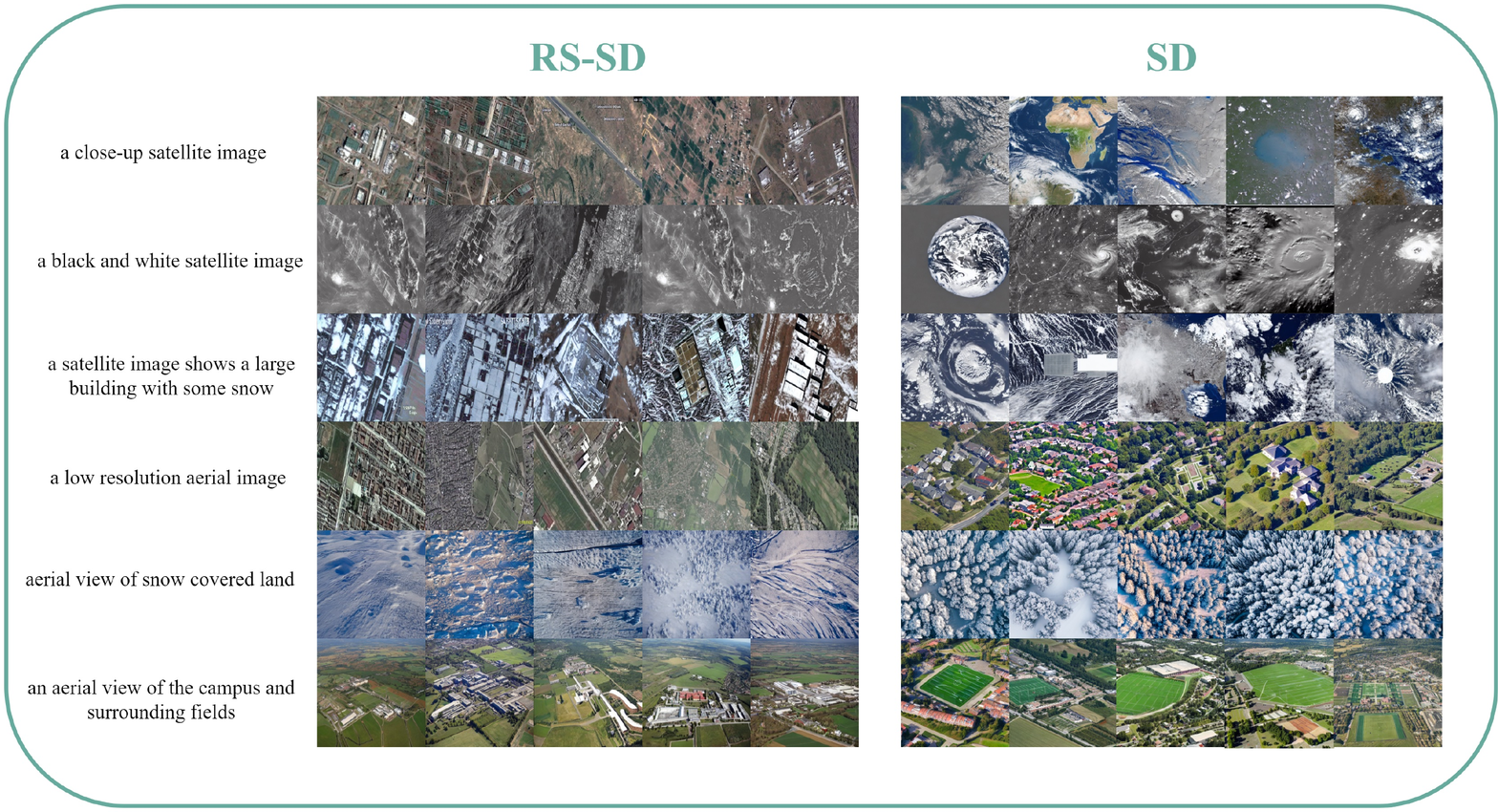}
    \caption{Comparison between images generated by SD and RS-SD with the same text prompts.}
    \label{fig:stable_diffusion_demo}
\end{figure}

\begin{figure}[htbp]
    \centering
    \includegraphics[width=0.49\textwidth]{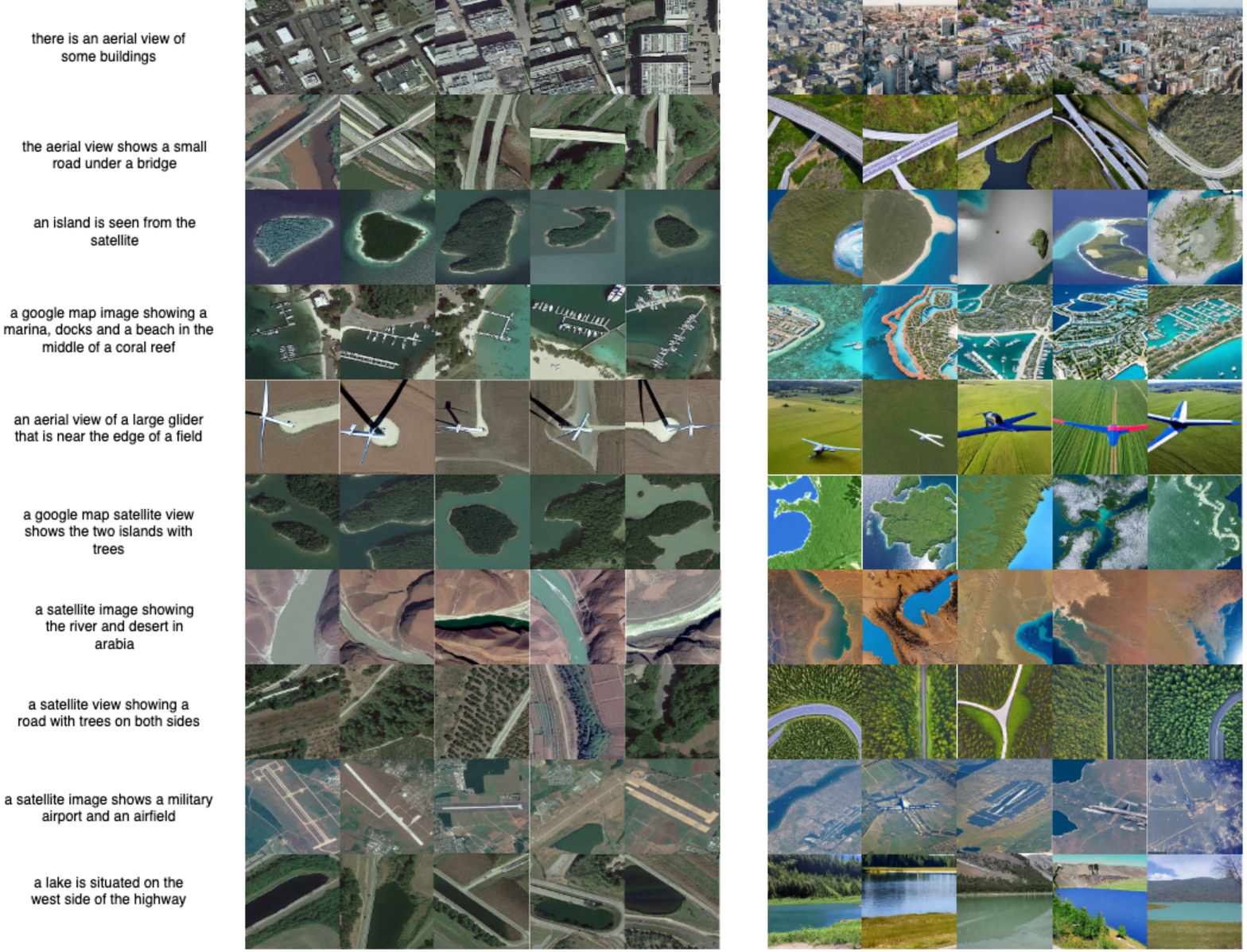}
    \caption{More Comparison between images generated by SD and RS-SD with the same text prompts. Left: RS-SD, Right: SD.}
    \label{fig:stable_diffusion_demo_2}
\end{figure}

We generate 40,000 samples using different queries to calculate the fid of vanilla Stable Diffusion and Tuned Stable Diffusion (RS-SD, tuned with RS5M). The vanilla Stable Diffusion model yields the FID score of \textbf{36.86} for the RS domain generation task. However, the RS-SD model achieves significantly improved FID scores of \textbf{28.32}. In overall, RS-SD outperforms vanilla SD in generating RS images qualitatively and quantitatively. The RS-SD model is capable of generating more realistic RS images that better match corresponding captions, regardless of whether the images are in satellite or aerial view.  As demonstrated by Figure \ref{fig:stable_diffusion_demo} and Figure \ref{fig:stable_diffusion_demo_2}, for prompts containing "satellite", the vanilla SD tends to generate unrealistic or meteorological images, but RS-SD can generate the RS images that are more realistic and in accord with RS images for common RS downstream tasks. Besides, the understandings of "snow-covered land", "building with some snow" and "surrounding fields" of RS-SD are significantly better than the SD.

\end{document}